\documentclass[11pt]{article}

\usepackage{hyperref}
\hypersetup{colorlinks,linkcolor={blue},citecolor={blue},urlcolor={red}}

\usepackage[verbose=true,letterpaper,margin=0.9in]{geometry}

\usepackage[mathscr]{eucal}
\usepackage{amsbsy}

\DeclareMathAlphabet{\mathpzc}{OT1}{pzc}{m}{it}

\usepackage{amsmath}     % core math
\usepackage{amssymb}     % symbols
\usepackage{amsfonts}    % fonts
\usepackage{mathtools}   % extensions; must come after amsmath
\usepackage{amsthm}      % proof env + qedhere

\usepackage{fancyhdr}
\usepackage{booktabs}

\usepackage{enumitem}
\usepackage{float}

\usepackage{color}

\usepackage{algorithm}
\usepackage{algpseudocode}

\usepackage{mathabx}
\usepackage{bbm}

\usepackage{caption}

\usepackage{subcaption}
\usepackage{scalerel}

\usepackage{tikz}

\usepackage[giveninits=true,  % use initials only
  maxnames = 2,  % won't do et al.
  backref=true,
  backend=bibtex,% or biber
  style=alphabetic,
  natbib=true,
  doi=false,
  isbn=false,
  url=false]{biblatex}

\usepackage{macros}

\newlist{cdesc}{description}{1}
\setlist[cdesc]{font=\mdseries,itemsep=0.5pt}

\usepackage{macros}

\usepackage[noabbrev,nameinlink]{cleveref}

\usepackage{multirow}

\begin{document}

%%%%%%% TITLE PAGE %%%%%%%%%%%%%%%%%%%%%%%%%%%%%%%%%%%%%%%%%%%%%%%%%%%

\begin{center}

  {\bf{\LARGE{Classification Imbalance \\
  \vspace{0.15cm}
  as Transfer Learning}}}
\vspace*{.2in}

{\large{
\begin{tabular}{ccc}
Eric Xia & Jason M.
Klusowski 
\end{tabular}
}}

\vspace*{.2in}

\begin{tabular}{c}
  Department of Operations Research \& Financial Engineering\\
  Princeton University, Princeton NJ
\end{tabular}

\medskip

\today

\vspace*{.2in}

\begin{abstract}
Classification imbalance arises when one class is much rarer than the other. We frame this setting as transfer learning under label (prior) shift between an imbalanced source distribution induced by the observed data and a balanced target distribution under which performance is evaluated. Within this framework, we study a family of  oversampling procedures that augment the training data by generating synthetic samples from an estimated minority-class distribution to roughly balance the classes, among which the celebrated SMOTE algorithm is a canonical example. We show that the excess risk decomposes into the rate achievable under balanced training (as if the data had been drawn from the balanced target distribution) and an additional term, the cost of transfer, which quantifies the discrepancy between the estimated and true minority-class distributions. In particular, we show that the cost of transfer for SMOTE dominates that of bootstrapping (random oversampling) in moderately high dimensions, suggesting that we should expect bootstrapping to have better performance than SMOTE in general. We corroborate these findings with experimental evidence. More broadly, our results provide guidance for choosing among augmentation strategies for imbalanced classification. 

\end{abstract}

\end{center}

\section{Introduction}

Classification imbalance, a setting where we have far more samples of one class than the other, is a common challenge that applied machine learning (ML) practitioners face. Detecting fraudulent credit card transactions is difficult because most transactions are legitimate~\cite{puh2019detecting}. The technology industry faces challenges in designing personalized advertisements because most users do not interact with ads~\cite{he2014practical}. Uber struggles to identify which drivers are most likely to commit sexual assault because of this imbalance, even though it ``[receives] a report of sexual assault or sexual misconduct ... almost every eight minutes''~\cite{NYTuber}. Using machine learning to identify those at risk of developing diseases is challenging because most patients in datasets are healthy, even for relatively common diseases like type-II diabetes~\cite{alghamdi2017predicting}. These challenges span a multitude of domains, often in settings with great consequences, driving interest in designing better methods.

As an overview of our results, we cast classification imbalance as a transfer learning problem under label (prior) shift: we observe data from a distribution that has imbalanced classes and seek to obtain a classifier that performs well under a target distribution with balanced class priors. This position is supported by the fact that the most common approaches used to address classification imbalance adopt a rebalancing strategy in which one trains a classifier on a dataset that follows a more balanced distribution via synthetic data. 

%%%%%%%%%%%%%%%%%%%%%%%%%%%%%%%%%%%%%%%%%%%%%%%%%%%%%%%%%%%%%%%%%%%%%

\section{Background}
\label{sec:background}

\subsection{Problem setting}

The formulation of binary classification is as follows. We observe samples \mbox{$(\Covariate_i, \Response_i) \stackrel{i.i.d.}{\sim} \PP = \PP_{\Covariate, \Response}$}, where $\Covariate \in \RR^d$ and $\Response \in \{0, 1\}$. We use $\cprob_0 \coloneqq \PP(\Response = 0)$ and $\cprob_1 \coloneqq \PP(\Response = 1) = 1 - \cprob_0$ to denote the (unconditional) probability that we observe each class. Adopting the convention that positive responses are rare, i.e., $\cprob_0 > \cprob_1$, practitioners often report that na\"{\i}vely applying an off-the-shelf classification algorithm to $\{(\Covariate_i, \Response_i)\}_{i=1}^\numobs$ will fail because it returns a classifier $\phat$ that nearly always predicts $\Response = 0$, i.e., $\phat(\Covariate) = 0$ too often when the true class is $\Response = 1$. More precisely, the type-II error $\PP(\phat(\Covariate) = 0 \mid \Response = 1)$ is too high. We claim that applied machine learning researchers facing such challenges are implicitly trying to solve a \emph{transfer learning} problem, albeit one with a lot of structure. 

Observe that we can write the distribution $\PP_{\Covariate, \Response}$ as 
\begin{align}
\label{eqn:observed}
\Response \sim \text{Ber}(\cprob_1), \qquad \Covariate \mid \Response \sim (1 - \Response) \cdot \PP_{\Covariate \mid \Response = 0} + \Response \cdot \PP_{\Covariate \mid \Response = 1}.
\end{align} 
In this case, we can write the marginal distribution as $\Covariate \sim \PP_\Covariate = \cprob_0 \cdot \PP_{\Covariate \mid \Response = 0} + \cprob_1 \cdot \PP_{\Covariate \mid \Response = 1}$, and the conditional distribution as 
\begin{align*}
\PP(\Response = 1 \mid \Covariate) = \frac{\cprob_1 \cdot d\PP_{\Covariate \mid \Response = 1}}{\cprob_0 \cdot d\PP_{\Covariate \mid \Response = 0} + \cprob_1 \cdot d\PP_{\Covariate \mid \Response = 1}}.
\end{align*}
Even with prior knowledge of the conditional distribution, this setting can still be challenging as large values of $\cprob_0$ will cause the type-II error to be high. As a special case, consider the Gaussian mixture model $\PP_{\Covariate \mid \Response = l} = \normal(\mu_l, 1)$ for $l = 0, 1$ where $\mu_l \in \RR$ is known, and, without loss of generality, assume $\mu_1 > \mu_0$. Consider the Bayes optimal classifier given by thresholding the conditional distribution, i.e., $\pstar(\Covariate) = \mathbf{1}\{\PP(\Response = 1 \mid \Covariate) \geq 1/2\}$. We can compute the type-II error as
\begin{align}
\label{eqn:type-II-error}
\PP(\pstar(\Covariate) = 0 \mid \Response = 1) = \Phi\left( - \tfrac{\mu_1 - \mu_0}{2} + \tfrac{1}{\mu_1 - \mu_0} \log(\tfrac{\cprob_0}{\cprob_1})  \right),
\end{align}
where $\Phi(\cdot)$ is the CDF of a standard Gaussian. Thus, large values of $\cprob_0/\cprob_1$ will cause the Bayes optimal classifier to have large type-II error, indicating that the difficulty of imbalance is fundamental to the distribution.

%Furthermore, the lower bound of $\Phi\left( \sqrt{2 \log(\tfrac{\cprob_0}{\cprob_1})} \right)$ indicates that no matter how large the signal $\frac{\mu_1 - \mu_0}{2}$ is, we will still pay a fundamental price regarding the classification imbalance. 

In contrast, we argue that practitioners are interested in obtaining a classifier that performs well when the data is generated according to the (hypothetical) balanced distribution $\QQ = \QQ_{\Covariate, \Response}$, where
\begin{align}
\label{eqn:target}
\Response \sim \text{Ber}(\tfrac{1}{2}), \qquad \Covariate \mid \Response \sim (1 - \Response) \cdot \PP_{\Covariate \mid \Response = 0} + \Response \cdot \PP_{\Covariate \mid \Response = 1}.
\end{align}
This results in a marginal distribution $\Covariate \sim \QQ_\Covariate \coloneqq \frac{1}{2} \cdot \PP_{\Covariate \mid \Response = 0} + \frac{1}{2} \cdot \PP_{\Covariate \mid \Response = 1}$ and conditional distribution
\begin{align*}
\QQ(\Response = 1 \mid \Covariate) = \frac{d\PP_{\Covariate \mid \Response = 1}}{d\PP_{\Covariate \mid \Response = 0} + d\PP_{\Covariate \mid \Response = 1}}.
\end{align*}
Furthermore, for the analogous version of the Gaussian mixture model described previously, we can compute the type-II error of $\qstar(\Covariate) = \mathbf{1}\{\QQ(\Response = 1 \mid \Covariate) \geq 1/2\}$ as
\begin{align}
\label{eqn:type-II-error-Q}
\PP(\qstar(\Covariate) = 0 \mid \Response = 1) = \Phi\left(-\tfrac{\mu_1 - \mu_0}{2}\right).
\end{align}
We use $\PP$ here instead of $\QQ$ because the distribution $\Covariate \mid \Response = 1$ is the same under $\PP$ or $\QQ$. As expected, the error does not exhibit the same imbalance issues that it did under $\PP$. Thus, obtaining a classifier based on data observed from $\PP$ that classifies well when test data are observed from $\QQ$, an instance of transfer learning, will not suffer from the aforementioned large type-II errors.

Furthermore, we can interpret many methods that practitioners use to address classification imbalance as natural approaches to training a classifier under $\QQ$. Suppose we have a dataset $\dataset \defn \{(\Covariate_i, \Response_i)\}_{i=1}^\numobs$ generated according to $\PP_{\Covariate, \Response}$. We define $\dataset_l \coloneqq \{(\Covariate_i, \Response_i) \in \dataset \mid \Response_i = l\}$ and $\numobs_l = |\dataset_l|$ for $l = 0, 1$; that is, $\dataset_l$  contains all the ($\numobs_l$) points for which we have $\Response_i = l$. The simplest method is to subsample from the majority class $\Response = 0$ to construct a dataset $\dataset'$ where there is little classification imbalance and then train a classifier on $\dataset'$. With this approach, the resulting distribution of $\dataset'$ will be close to a sample drawn i.i.d.~from $\QQ$, and the obtained classifier (ideally) will have good performance on this distribution. Alternatively, a popular class of methods involves generating synthetic samples $\SCovariate \sim \SynDist$ for some $\SynDist \approx \PP_{\Covariate \mid \Response = 1}$ and training a classifier on the augmented dataset 
$
\dataset' = \dataset \cup \{(\SCovariate_j, 1)\}_{j=1}^J
$
where $J$ is chosen so that the classes are approximately balanced. We call this class of methods \emph{rebalancing methods}. If $\SynDist = \PP_{\Covariate \mid \Response = 1}$, then $\dataset'$ (with an appropriate $J$) will have distribution that resembles $\QQ$. These include resampling techniques; the simplest such example uses bootstrapping to resample, where $\SynDist$ is the empirical distribution over $\dataset_1$, which the literature often refers to as \emph{random oversampling}. The celebrated SMOTE method perturbs samples in $\dataset_1$ to generate new samples whose distribution imitates $\PP_{\Covariate \mid \Response = 1}$.

Thus, the problem of classification imbalance can be precisely stated as transfer learning: given data drawn from $\PP = \PP_{\Covariate, \Response}$ (i.e.,~\Cref{eqn:observed}), how do we obtain a classifier that performs well for data drawn from $\QQ = \QQ_{\Covariate, \Response}$ (i.e.,~\Cref{eqn:target})? Formally, given observations $(\Covariate, \Response) \sim \PP_{\Covariate, \Response}$ we seek to estimate the conditional distribution
\begin{align*}
\fstar(\covariate) = \QQ(\Response = 1 \mid \Covariate = \covariate).
\end{align*}

\subsection{SMOTE and bootstrapping}

As a preview of our results, we illustrate one (of many) conclusions that our theory provides; namely, a theoretical comparison of the performance of SMOTE and bootstrapping (random oversampling), henceforth denoted by $\fhat^{\text{SMOTE}}$ and $\fhat^{\text{BOOT}}$, respectively. We establish that the excess risks of $\fhat^{\text{SMOTE}}$ and $\fhat^{\text{BOOT}}$ are bounded above by
\begin{align*}
\EE_\QQ\left[\loss(\Response, \fhat^{\text{BOOT}}(\Covariate)) - \loss(\Response, \fstar(\Covariate)) \right] &\lesssim \text{Oracle}_{\numobs}(\QQ) + \text{Oracle}_{\numobs_1}(\CondDist{1}), \quad \text{and} \\
\EE_\QQ\left[\loss(\Response, \fhat^{\text{SMOTE}}(\Covariate)) - \loss(\Response, \fstar(\Covariate)) \right] &\lesssim \text{Oracle}_{\numobs}(\QQ) + \text{Oracle}_{\numobs_1}(\CondDist{1}) + \left(\frac{1}{\numobs_1}\right)^{1/d}.
\end{align*}
Here $\text{Oracle}_{n}(\PP')$, to be defined precisely later, represents the oracle error achieved by performing empirical risk minimization with respect to a loss $\loss$ on $n$ samples drawn from $\PP'$. As we can see, the SMOTE guarantee has an additional $\numobs_1^{-1/d}$ term, exhibiting a very strong curse-of-dimensionality error term. This term matches the convergence rate typical of nearest-neighbor methods, which are often discouraged in high dimensions due to their inferior predictive performance relative to modern machine learning algorithms. Accordingly, our theory cautions against using SMOTE, unless dimensionality-driven costs are negligible in the regime of interest.

\subsection{Related work}

Since classification imbalance is a problem of great practical interest, there have been several recent theoretical works studying this setting. Our transfer learning formulation bears some similarity to the idea of \textit{balanced risk} approaches~\cite{lyu2025statistical, lyu2025bias, nakada2024synthetic, ahmad2025concentration}. Often the minimizer of these balanced risk loss functions is given by $\fstar$.~\citet{nakada2024synthetic} provided some theory on rebalancing approaches for parametric settings. The paper~\cite{lyu2025statistical} considered the proportional asymptotics regime under a Gaussian mixture model with $\Covariate$ having i.i.d.~entries. Concurrently,~\citet{ahmad2025concentration} study a variant of the SMOTE procedure under a set of structural and distributional assumptions, obtaining nonasymptotic bounds. When we specialize our framework to SMOTE, we obtain bounds that improve certain aspects of the rates in~\citet{ahmad2025concentration} under a different, and in some directions weaker, set of assumptions. The approach in ~\citet{lyu2025bias} can be interpreted in our framework as incurring a cost of transfer that is not asymptotically negligible and relies on assumptions that are stronger than those used here. Compared to these balanced-risk approaches, our formulation differs in several qualitative ways. See~\Cref{sec:comparison} for an in-depth comparison.

As mentioned earlier, arguably the most popular approach is the SMOTE algorithm~\cite{chawla2002smote}, which synthetically generates samples from the minority class, and then trains a machine learning algorithm on the dataset augmented with these synthetic samples. Due to its simplicity and popularity, a large number of variants have been proposed.~\citet{gao2012probability} consider a variant that involves using kernel density estimation to obtain $\SynDist$ and sample from it. ADASYN and borderline-SMOTE~\cite{he2008adasyn, han2005borderline} are variants of SMOTE that adaptively sample based on the number of majority class samples that are the nearest neighbors of a given minority class sample. Beyond these, there are a plethora of high-impact SMOTE variants~\cite{douzas2018improving, blagus2013smote, dablain2022deepsmote} that continue to be published. However, despite the popularity of SMOTE in the literature, it seems that the empirical literature has not reached a consensus on its benefits. There is active debate in the applied literature~\cite{elor2022smotesmote} and online forums on machine learning~\cite{reddit1SMOTE, reddit2SMOTE} about the empirical benefits of SMOTE, with some reports finding limited gains relative to simpler baselines. A technical report from Meta~\cite{he2014practical} writes that in dealing with classification imbalance, undersampling the majority class (which we study in~\Cref{sec:undersampling-methods}) performs well, with no mention of SMOTE. Our analysis provides a theoretical perspective on when and why such behavior can occur.

Our approach to studying classification imbalance is motivated by the transfer learning literature, which is fundamentally concerned with methods for estimating a desired quantity according to the target distribution $\QQ$ when accessing data according to the source distribution $\PP$. We provide a summary of the literature here. One well-known setting in transfer learning is covariate shift~\cite{shimodaira2000improving, ma2023optimally}, where the relationship between $\Response$ and $\Covariate$ remains the same between $\PP$ and $\QQ$ but the input distribution of $\Covariate$ differs. These papers often study methods that involve adjusting for the differences in covariate distributions, such as likelihood reweighting. Our approach is an example of label shift~\cite{lipton2018detecting, garg2020unified}, where the label distribution $\PP_\Response$ changes, but the conditional distribution $\PP_{\Covariate \mid \Response}$ does not. Another line of work~\cite{kpotufe2021marginal, fan2025robust, cai2021transfer} focuses on transfer learning in binary classification where the relationship between $\Response$ and $\Covariate$ may change between the source and target but is constrained in some way to allow for statistical benefits of incorporating source data.  An alternative transfer learning approach for classification imbalance is to incorporate modeling of some intermediate outcome that is less rare than the desired one~\cite{bastani2021predicting, faletto2023predicting}. One limitation of these various approaches is that the assumptions about the relationship between source and target are made for theoretical reasons and not necessarily justified by the applications. The formulation here, by contrast, follows from first principles.

\subsection{Our contributions}

This paper provides a unified theoretical framework for studying classification imbalance through the lens of transfer learning. Our main contributions can be summarized as follows.

\begin{itemize}

\item \textbf{A general excess-risk decomposition for rebalancing.}
We state a general rebalancing procedure (Algorithm~\ref{AlgRebal}) that augments the minority class using an arbitrary synthetic generator $\SynDist$. Our first result, \Cref{thm:slow-rates}, shows that for uniformly bounded losses
\[
\EE_\QQ\big[\loss(\Response,\fhat(\Covariate)) - \loss(\Response,\fstar(\Covariate))\big]
\lesssim \text{Oracle}_{\numobs}(\QQ) + \text{Cost}(\SynDist,\CondDist{1}),
\]
where $\text{Cost}(\SynDist,\CondDist{1})$ is the \emph{cost of transfer}, controlled by the total-variation distance between $\SynDist$ and the minority-class distribution $\CondDist{1}$. This result is fully agnostic to the choice of synthetic generator, loss, and hypothesis class. We remark these results do not follow directly from applying standard empirical process analysis because those would lead to oracle terms dependent on $\SynDist$. To match the transfer learning setting, we desire bounds that are expressed solely in terms of the target distribution $\QQ$, so they match the oracle rate without transfer. We address this by constructing an explicit coupling between $\CondDist{1}$ and $\SynDist$ that allows us to transfer empirical process guarantees from $\SynDist$ back to $\QQ$.

\item \textbf{Fast rates via localization and $\chi^2$-type cost of transfer.}
Under additional Lipschitz and strong convexity assumptions on the loss, we sharpen the bounds above using localized empirical process techniques. In \Cref{thm:fast-rates}, we show that the estimation error between $\fhat$ and $\fstar$ under $\QQ$ can be controlled by localized complexity measures under $\QQ$ and under $\SynDist$, together with an explicit cost of transfer term that depends on the $\chi^2$-divergence between $\SynDist$ and $\CondDist{1}$. For parametric model classes, this yields (in some cases, optimal) $O(\numobs^{-1/2})$ rates for the estimation error under $\QQ$. Obtaining these fast rates requires adapting the earlier arguments to reflect the localized nature of the analysis.

\item \textbf{Concrete guarantees and guidance for synthetic generators.}
Specializing our meta-results to several concrete choices of $\SynDist$ (including kernel density estimators, diffusion-based generators, bootstrapping, and SMOTE) we obtain nonasymptotic bounds that make the dependence on the ambient dimension $d$, the minority sample size $\numobs_1$, and algorithmic parameters (e.g., the number of neighbors $k$ in SMOTE) explicit; see~\Cref{sec:instantiations}. In particular, we show that bootstrapping behaves, up to logarithmic factors, like having access to $2\numobs_1$ minority samples from $\CondDist{1}$, whereas SMOTE incurs an additional term of order $\numobs_1^{-1/d}$, exhibiting a curse-of-dimensionality phenomenon. These results yield concrete, theory-driven recommendations on when one should expect bootstrapping, SMOTE, density estimation, or diffusion-based methods to be preferable. Standard approaches to analyzing SMOTE often require a lower bound on the density of $\CondDist{1}$~\cite{ahmad2025concentration}. We relax this requirement by invoking results from the high-dimensional geometry literature, which allows us to control the geometry of $k$-nearest neighbor graphs and some dimension-dependent terms (for example, improving an exponential-in-$d$ factor to a base $(3/2)^d$ in the special case $ k =1$), providing a more nuanced characterization of the curse-of-dimensionality in SMOTE. See~\Cref{sec:discrete-synthetic} for this comparison between SMOTE and bootstrapping as well as further details.

\item \textbf{Undersampling and plug-in methods as complementary baselines.} 
Within our framework, we study a plug-in estimator of the balanced conditional probability
$\fstar(\Covariate)=\QQ(\Response=1\mid\Covariate)$, obtained by reweighting a standard
estimate of $\PP(\Response=1\mid\Covariate)$; see~\Cref{prop:plug-in}.
We also analyze undersampling methods and show that, up to logarithmic factors, their guarantees
parallel those of oversampling-based rebalancing, with an effective sample size governed by the
downsampled majority class; see~\Cref{prop:undersampling} in the supplement~\cite{proofs}.

\item \textbf{General target mixtures and data-dependent $J$.}
Although we focus on considering a balanced objective with input distribution $\QQ_\Covariate = \frac{1}{2} \CondDist{0} + \frac{1}{2} \CondDist{1}$, our results can be extended to arbitrary mixtures $\QQ^*_\Covariate = \cprob_0' \CondDist{0} + \cprob_1' \CondDist{1}$, allowing one to represent alternative target regimes such as cost-sensitive risk, type-I/type-II error control, regulatory/compliance requirements, resource/budget limits, and safety/fairness constraints. In addition, some of the guarantees rely on the number of synthetic samples $J$ to be deterministic; the results in this paper can be extended to handle random $J$ dependent on the data. These results, in~\Cref{sec:more-results}, as well as additional proofs, can be found in the supplement~\cite{proofs}.

\end{itemize}

\subsection{Paper organization and notation}

\vspace{3pt}
\noindent\textbf{Paper organization:} The remainder of the paper is organized as follows.~\Cref{sec:rebalancing} provides guarantees on the rebalancing estimator, and in~\Cref{sec:further-results} we instantiate our general theory for several concrete choices of $\SynDist$, along with various other extensions.~\Cref{sec:plugin} is dedicated to analyzing the plug-in estimator. The results of numerical simulations are in~\Cref{sec:sims}. Finally, we conclude with some discussion in~\Cref{sec:discussion}. An overview of our proofs can be found in~\Cref{sec:proofs}.

\vspace{3pt}
\noindent\textbf{Notation:} Here we introduce notation used throughout our paper. We adopt the shorthand $[n] \defn \{1, 2, \ldots, n\}$. We write $(\Covariate_1, \Covariate_2) \sim \PP_1 \otimes \PP_2$ to indicate that we draw $\Covariate_1 \sim \PP_1$, $\Covariate_2 \sim \PP_2$ such that $\Covariate_1 \indep \Covariate_2$.  Given a random variable $\Covariate$, we use $\text{supp}(\Covariate)$ to denote the support of $\Covariate$. For a distribution $\PP'_{\Covariate}$ on the covariates, we write $\|f\|_{\PP'_{\Covariate}} \coloneqq \EE_{\Covariate \sim\PP'_{\Covariate}}[f(\Covariate)^2]^{1/2}$. Furthermore, we use $\Delta$ to indicate the Dirac delta distribution, i.e., $\Delta_{\Response = 1}(\Response = 1) = 1.$ We also use $\lesssim$ and $\gtrsim$ to denote relations that hold up to constants. Furthermore, we use $\BB_d(0, r)$ to denote the ball of radius $r$ centered at $0$ in $\RR^d$. For two sequences $\{a_n\}$ and $\{b_n\}$ we write $a_n \asymp b_n$ if $a_n = O(b_n)$ and $b_n = O(a_n)$ where $O$ and $O_\PP$ retain the standard definitions. For convenience, we write $I_0 = \{i \in [\numobs] : \Response_i = 0\}$, $I_1 = \{i \in [\numobs] : \Response_i = 1\}$, and $\totobs \defn \numobs + J$.

%%%%%%%%%%%%%%%%%%%%%%%%%%%%%%%%%%%%%%%%%%%%%%%%%%%%%%%%%%%%%%%%%%%%%

\section{Rebalancing approaches}
\label{sec:rebalancing}

Given the formulation as described in~\Cref{sec:background}, there are a few approaches to estimating $\fstar$. Our focus here is based on generating synthetic samples from some distribution $\SynDist$ to construct a dataset that is more balanced, which we call the \emph{rebalancing} method. We provide a meta-result for rebalancing procedures agnostic to the choice of $\SynDist$. The procedure is formally defined in Algorithm~\ref{AlgRebal}. We focus on the setting where we augment $\dataset$.

Note that classification methods typically rely on minimizing an empirical loss $\loss$ over a class of functions $\fclass$.
Throughout the paper, we assume the realizable setting, $\fstar \in \fclass$, and we use a (strictly) proper loss so that
\begin{align*}
\fstar \in \argmin_{f \in \fclass} \EE_{\QQ}[\loss(\Response, f(\Covariate))].
\end{align*}
Thus, given data observed from $\QQ$, we expect that performing empirical risk minimization using the loss function $\loss$ will converge to the desired $\fstar$. The challenges arise because we lack access to data generated from $\QQ$ and instead use an approximation of it. The most popular choice of loss function in modern ML is the cross-entropy loss
\begin{align*}
\loss(\Response, f(\Covariate)) = -\Response \log(f(\Covariate)) - (1 - \Response) \log(1 - f(\Covariate)).
\end{align*}
We remark that since we are trying to estimate probabilities, we assume that $f(x) \in [0, 1]$ for all $x$ and $f \in \fclass$.

\begin{algorithm}[t]
\caption{Rebalancing method}
\begin{algorithmic}[1]
\State \texttt{Inputs}: (i) Dataset $\dataset = \{(\Covariate_i, \Response_i)\}_{i=1}^\numobs$ and number of synthetic samples $J$. (ii) Procedure $\mathcal{P}$ for constructing $\SynDist$. (iii) Loss function $\loss$ and function class $\fclass$ for estimating $\fstar$. 
\vspace{6pt}

\State Construct the distribution $\SynDist$ by applying procedure $\mathcal{P}$ to $\dataset$ (or $\dataset_1$).

\State Generate $J$ independent synthetic samples $\SCovariate_j \sim \SynDist$ for $j =1, \ldots, J$.

\State Compute the estimate
\begin{align*}
\fhat \in \argmin_{f \in \fclass} \left( \frac{1}{\numobs + J} \left\{ \sum_{i=1}^\numobs \loss(\Response_i, f(\Covariate_i)) + \sum_{j=1}^J \loss(1, f(\SCovariate_j))  \right\} \right).
\end{align*}
\end{algorithmic}
\label{AlgRebal}
\end{algorithm}

This section is organized as follows. In~\Cref{sec:intuition}, we present some basic calculations to illustrate what should be expected from these guarantees. Then in~\Cref{sec:slow-rates}, we focus on controlling the \emph{excess risk}, i.e., upper bounding the quantity $\EE_\QQ\left[\loss(\Response, \fhat(\Covariate)) - \loss(\Response, \fstar(\Covariate))\right]$. Finally, in~\Cref{sec:fast-rates} we focus on controlling the estimation error $\| \fhat - \fstar \|_{\QQ_\Covariate}$, obtaining faster rates but under stronger assumptions.

\subsection{Some intuition}
\label{sec:intuition}

We begin by providing some intuition and motivation for our results. The first part is dedicated to relating the gap in type-II error with the excess risk and the estimation error, which means that controlling these quantities also controls type-II error. The second part describes the population minimizer of the empirical loss described in Algorithm~\ref{AlgRebal}, giving some insight as to the influence of the synthetic distribution $\SynDist$. See~\Cref{app:intuition-calc} for the proof of these results.

\vspace{3pt}
\noindent\textbf{Type-II error comparison:} Define $\qhat(\Covariate) \coloneqq \mathbf{1}\{\fhat(\Covariate) \geq \frac{1}{2}\}$ and recall $\qstar(\Covariate) = \mathbf{1}\{\fstar(\Covariate) \geq \frac{1}{2}\}$. We can bound the type-II error under some structural assumptions on the loss $\loss$ and the margin of $\fstar$ via
\begin{align}
\begin{split}
\label{eqn:typeII-Qnorm}
&\left| \PP(\qhat(\Covariate) = 0 \mid \Response = 1) - \PP(\qstar(\Covariate) = 0 \mid \Response = 1) \right| \\
&\qquad \qquad \lesssim \min\left\{\|\fhat - \fstar\|_{\QQ_\Covariate},\; \psi^{-1}_{\loss}\left(\EE_\QQ\left[\loss(\Response,  \fhat(\Covariate)) - \loss(\Response, \fstar(\Covariate)) \right] \right) \right\},
\end{split}
\end{align}
where $\psi_\loss$ is an increasing convex function dependent on $\loss$ with $\psi_\loss(0) = 0$. Thus, under these conditions, minimizing the excess risk or the estimation error under $\QQ$ will result in type-II error control.

\vspace{3pt}
\noindent\textbf{Computing the population minimizer:} The minimizer of the following population risk
\begin{align*}
\ftil \in \argmin_{f \in \fclass} \, \EE\left[  \frac{1}{\numobs + J} \left\{ \sum_{i=1}^\numobs \loss(\Response_i, f(\Covariate_i)) + \sum_{j=1}^J \loss(1, f(\SCovariate_j))  \right\}  \right]
\end{align*}
for the cross-entropy loss is given by (taking the ideal choice $J = (2\cprob_0 - 1) \numobs$, assuming it is a positive integer)
\begin{align}
\label{eqn:pop-min}
\ftil(\Covariate) = \frac{\tfrac{1}{2} d\CondDist{1} + (1 - \frac{1}{2\cprob_0})(d\SynDist - d\CondDist{1})}{\tfrac{1}{2} d\CondDist{0} + \tfrac{1}{2} d\CondDist{1} + (1 - \frac{1}{2\cprob_0}) (d\SynDist - d\CondDist{1})}.
\end{align}
Note that the target is $\fstar(\Covariate) = \frac{\frac{1}{2} d\CondDist{1}}{\frac{1}{2} d\CondDist{0} + \frac{1}{2} d\CondDist{1}}$.

\subsection{Excess risk bounds}
\label{sec:slow-rates}

Here we state guarantees for the procedure described in Algorithm~\ref{AlgRebal}. The main error in our guarantees can be split into two terms: one is the \emph{oracle accuracy}, which reflects the rate of estimation if we had access to data generated from the target distribution $\QQ$, and the other is the \emph{cost of transfer}, which characterizes the price of generating synthetic samples from the distribution $\SynDist$ instead of the true $\CondDist{1}$.

The guarantees here are sometimes referred to as \emph{slow rates} in the literature, which impose weaker requirements on the loss function. For our guarantees we impose the condition that the loss is $\bound$-uniformly bounded, i.e.,
\begin{align}
\label{eqn:bound-loss}
\sup_{f \in \fclass} \left|\loss(\Response, f(\Covariate)) \right| \leq \bound \quad \text{a.s.}
\end{align}
 For the cross-entropy loss, this requires the functions in $\fclass$ to be bounded away from $0$ and $1$. However, this assumption is more for technical convenience and can be relaxed via truncation arguments or results on controlling unbounded empirical processes~\cite{adamczak2008unbounded}. Furthermore, let $\rad_i \stackrel{i.i.d.}{\sim} \text{Rad} = \text{Unif}\{-1, +1\}$ be a collection of independent Rademacher variables; for a positive integer $n$, we define the \emph{Rademacher complexity} with respect to a distribution $\PP$  as 
\begin{align}
\label{eqn:global-radcomp}
\RadComp_{n}(\PP) \coloneqq \EE_{\rad_i \stackrel{i.i.d.}{\sim} \text{Rad}, \; (\Covariate_i, \Response_i) \stackrel{i.i.d.}{\sim} \PP}\left[ \sup_{f \in \fclass} \left|\frac{1}{n}\sum_{i=1}^n \rad_i \cdot \loss(\Response_i, f(\Covariate_i)) \right|  \right].
\end{align}
A standard argument~\cite{dinosaur2019} establishes that given observations $\dataset = \{(\Covariate_i, \Response_i)\}_{i=1}^\numobs$ distributed according to $\PP$,  performing empirical risk minimization over $\dataset$ using the loss $\loss$ would result in excess risk $\RadComp_{\numobs}(\PP)$ along with some higher-order fluctuations. We remark that we will occasionally abuse notation and write $\RadComp_{\numobs}(\CondDist{1})$ as shorthand for $\RadComp_{\numobs}(\CondDist{1} \otimes \Delta_{\Response = 1})$, and likewise for $\RadComp_{\numobs}(\SynDist)$. The final component of our error guarantee is the discrepancy between the true distribution $\CondDist{1}$ and the synthetic sampling distribution $\SynDist$. We measure the accuracy of sampling from $\SynDist$ when the desired distribution is $\CondDist{1}$ via the total-variation distance
\begin{align*}
\dTV{\SynDist}{\CondDist{1}} \coloneqq \sup_{A} \big| \SynDist(A) - \CondDist{1}(A) \big|,
\end{align*}
where the supremum is over all measurable subsets.

We now state a guarantee for Algorithm~\ref{AlgRebal}. For convenience, we use the shorthand 
\begin{align*}
t_{n}(\parprob) \coloneqq (14\sqrt{2\log(4/\parprob)} + 36) \cdot \tfrac{\bound}{\sqrt{n}}
\end{align*}
to express the higher-order probabilistic fluctuations. Recall the notation $\totobs \defn \numobs + J$.

\begin{theorem}
\label{thm:slow-rates}
Given observations $\dataset = \{(\Covariate_i, \Response_i)\}_{i=1}^\numobs \stackrel{i.i.d.}{\sim}\PP_{\Covariate, \Response}$, consider Algorithm~\ref{AlgRebal} implemented with an arbitrary synthetic sampling distribution $\SynDist$ and a $\bound$-uniformly bounded loss~\eqref{eqn:bound-loss}. Then for any $\parprob \in (0, 1)$, we have, with probability exceeding $1 - \parprob$,
\begin{align*}
\EE_{\QQ}\left[\loss(\Response, \fhat(\Covariate)) - \loss(\Response, \fstar(\Covariate))\right] & \leq 12\RadComp_{\totobs}(\QQ)  +  4\bound \cdot  \dTV{\SynDist}{\CondDist{1}}  \\
&\qquad \qquad  + 28\bound \cdot \left| \frac{\numobs\cprob_0}{\numobs + J} - \frac{1}{2} \right| + t_{\totobs}(\parprob).
\end{align*}
\end{theorem}

\noindent See~\Cref{sec:proof-slow-rates} for a proof of this result. We remark that while we provide explicit constants, they are by no means sharp. Several comments are in order to help interpret the various terms.

\vspace{3pt}
\noindent\textbf{Oracle excess risk:} In a setting without transfer learning, i.e.,~when we have $\numobs$ samples drawn i.i.d. from $\QQ$, the resulting empirical risk minimizer $\widebar{f}$ will have excess risk given by
\begin{align*}
\EE_\QQ\left[\loss(\Response, \widebar{f}(\Covariate)) - \loss(\Response, \fstar(\Covariate))\right] \lesssim \RadComp_{\numobs}(\QQ) + t_{\numobs}(\parprob)
\end{align*}
with high probability.
Thus, the quantity $\RadComp_{\totobs}(\QQ)$ is an oracle excess risk described by $\text{Oracle}_{\totobs}(\QQ)$ previously; this is an error term we would expect to see if there were no transfer learning involved. Often this term is sharp without further assumptions on the loss; testing between two different distributions via Le Cam's two-point argument (cf. Chapter 15 in~\citet{dinosaur2019}) has a lower bound of $\Omega(\numobs^{-1/2})$, matching the Rademacher complexity.

\vspace{3pt}
\noindent\textbf{TV-distance:} If the previous quantity represented the oracle excess risk, the TV-distance $\dTV{\SynDist}{\CondDist{1}}$ captures the \emph{cost} of transfer learning. It represents the additional error we pay  to get a guarantee in terms of the distribution $\QQ$ when we only observe data from $\PP$. The term arises from a coupling argument involving the synthetic samples $\{\SCovariate_j\}_{j=1}^J$.

\vspace{3pt}
\noindent\textbf{Mixture bias:} The final quantity $|\frac{\numobs\cprob_0}{\numobs + J} - \frac{1}{2}| $ captures the deviation in expectation of the mixture distribution $\dataset \cup \{(\SCovariate_j, 1)\}_{j=1}^J$ from $\QQ$; if $J = (2\cprob_0 - 1) \numobs$ then this quantity is zero. Since $J$ must be a positive integer, if we choose $J = (2\cprob_0 - 1) \numobs + \Delta$ for some $\Delta$ (e.g., $J = \ceil{(2\cprob_0 - 1) \numobs}$), we can compute $\left| \frac{\numobs\cprob_0}{\numobs + J} - \frac{1}{2} \right| \asymp \frac{|\Delta|}{\numobs}$.
Thus, the deviation of the choice of $J$ shows up in the overall error as being rescaled by $\numobs$. However, this choice requires that $J$ is fixed; see~\Cref{sec:random-J} for random $J$.

\subsection{Localization bounds}
\label{sec:fast-rates}

Now we provide guarantees that are often referred to as \emph{fast rates}, with stronger requirements imposed on the loss function $\loss$, namely Lipschitz and strong convexity. We require that the loss is $L$-Lipschitz in its second argument
\begin{subequations}
\begin{align}
\label{eqn:lipschitz}
\big|\loss(y, u) - \loss(y, v) \big| \leq L\big|u - v\big|,
\end{align}
for all $y \in \{0, 1\}$ and $u,v \in [0, 1]$.
Furthermore, we require that $\EE_\QQ[\loss(\Response, f(\Covariate))]$ is $\gamma$-strongly convex at $\fstar$:
\begin{align}
\label{eqn:strong-convex}
\EE_{\QQ}\left[\loss(\Response, f(\Covariate)) - \loss(\Response, \fstar(\Covariate)) \right] \geq \frac{\gamma}{2} \| f - \fstar \|_{\QQ_\Covariate}^2, \quad \text{for all} \; f \in \fclass.
\end{align}
\end{subequations}
We remark that these assumptions hold for a broad class of loss functions, including the cross-entropy loss with some additional restrictions on $\fclass$. 

% We now verify that the cross-entropy loss function
% %
% \begin{align*}
% \loss(\Response, f(\Covariate)) = - \Response \log(f(\Covariate)) - (1 - \Response) \log(1 - f(\Covariate))
% \end{align*}
% %
% satisfies these conditions. Strong convexity follows from the fact that
% %
% \begin{align*}
% \EE_\QQ\left[ \loss(\Response, f(\Covariate)) \right] - \EE_\QQ\left[ \loss(\Response, \fstar(\Covariate)) \right] &= \EE_\QQ[\kull{\fstar(\Covariate)}{f(\Covariate)}] \\
% &\geq \EE_\QQ\left[\frac{1}{2}\big(\fstar(\Covariate) - f(\Covariate)\big)^2\right] = \frac{1}{2} \|f - \fstar \|_{\QQ_\Covariate}^2.
% \end{align*}
% %
% If we impose the requirement that $\fstar$ and $\fclass$ are clipped, i.e.,~there exists $\eta$ such that$f(x) \in [\eta, 1- \eta]$ for all $x$, then~\Cref{eqn:lipschitz} follows from the mean value theorem. 

Key to obtaining fast rates are the notions of localized complexities and critical radii~\cite{bartlett2002localized}, in contrast to the global complexities defined in~\Cref{eqn:global-radcomp}. For a given $t > 0$ and positive integer $n$, we define the \emph{localized Rademacher complexity} as
\begin{align}
\label{eqn:local-radcomp}
\RadComp_{n}(t; \PP_\Covariate) \coloneqq \EE_{\rad_i \stackrel{i.i.d.}{\sim} \text{Rad},\; \Covariate_i \stackrel{i.i.d.}{\sim} \PP_\Covariate}\left[ \sup_{f \in \fclass:\|f - \fstar \|_{\PP_\Covariate} \leq t} \left|\frac{1}{n}\sum_{i=1}^n \rad_i \big( f(\Covariate_i) - \fstar(\Covariate_i) \big) \right| \right].
\end{align}
Crucial to our guarantees are the following \emph{critical radii} with respect to $\PP_\Covariate$, defined as the solution to the fixed point relation
\begin{align*}
\omega_n(\PP_\Covariate) \coloneqq \inf\left\{ t > 0: t \geq \frac{\RadComp_n(t; \PP_\Covariate)}{t}  \right\}.
\end{align*}
Of particular interest are the critical radii $\ocrit_{\numobs}(\QQ_\Covariate)$ and $\ocrit_{\numobs}(\SynDist)$. Note that such critical radii often determine the optimal rate of convergence in many scenarios; see Chapters 14 and 15 in~\citet{dinosaur2019} for further details. 

The dependence on the discrepancy between $\SynDist$ and $\CondDist{1}$ differs from before. We assume that $\SynDist$ is absolutely continuous with respect to $\QQ_{\Covariate}$, i.e., $\SynDist \ll \QQ_{\Covariate}$ (note that we always have $\CondDist{1} \ll \QQ_{\Covariate}$, since $\QQ_{\Covariate}$ is a mixture distribution with component $\CondDist{1}$).
%define the densities
%
%\begin{align*}
%\phat(x) \coloneqq \frac{d\SynDist}{d\QQ}(x) \qquad \text{and} \qquad \pstar(x) \coloneqq \frac{d\CondDist{1}}{d\QQ}(x).
%\end{align*}
%
Our upper bound depends on the $\chi^2$-divergence between $\SynDist$ and $\CondDist{1}$, defined as
\begin{align*}
\chi^2(\SynDist;\, \CondDist{1}) \coloneqq \int \frac{\left(\frac{d\SynDist}{d\QQ_\Covariate} - \frac{d\CondDist{1}}{d\QQ_\Covariate}\right)^2}{\frac{d\CondDist{1}}{d\QQ_\Covariate}} \, d\QQ_\Covariate.
\end{align*}

 Furthermore, we denote $\mcrit_n \coloneqq \min\{\ocrit_n(\QQ_\Covariate), \ocrit_n(\SynDist)\}$, and use the shorthand
\begin{align*}
s_n(\parprob) \coloneqq \frac{28L}{\gamma} \cdot \sqrt{\frac{\log(4\logfun(\mcrit_n)/\parprob)}{n}} + \frac{320L}{\gamma} \cdot \frac{\log(4\logfun(\mcrit_n)/\parprob)}{n\mcrit_n },
\end{align*}
where $\logfun(\mcrit_n) \coloneqq \log_2(\frac{4}{\mcrit_n})$. The following result is conditional on the outcomes $\{\Response_i\}_{i=1}^{\numobs}$.
\begin{theorem}
\label{thm:fast-rates}
Under the conditions of~\Cref{thm:slow-rates}, for any $L$-Lipschitz and $\gamma$-strongly convex loss, Algorithm~\ref{AlgRebal} with $J = \numobs_0 - \numobs_1$ returns an estimate $\fhat$ such that, for any $\parprob \in (0, 1)$,
\begin{align*}
\| \fhat - \fstar \|_{\QQ_\Covariate} &\leq \big( \tfrac{32L}{\gamma} + 1 \big) \cdot \big(8\ocrit_\totobs(\QQ_\Covariate) + \ocrit_\totobs(\SynDist)\big)  \\ & \qquad\qquad + \big( \tfrac{2L}{\gamma} + 1 \big) \sqrt{2\chi^2(\SynDist;\, \CondDist{1})}  + s_\totobs(\parprob),
\end{align*}
with probability exceeding $1 - \parprob$.
\end{theorem}

\noindent See~\Cref{sec:proof-fast-rates} for a proof of this result. Several comments are to help interpret this theorem. Note that in most cases of interest, $\mcrit_\totobs \gtrsim \frac{1}{\sqrt{\totobs}}$ which implies that $s_\totobs(\parprob) \in O\Big(\frac{\log(\log(\totobs))}{\sqrt{\totobs}}\Big)$, making it higher order in most cases of interest.

\vspace{3pt}
\noindent\textbf{Critical radii:} In a setting without transfer learning, i.e.,~when we have $\numobs$ samples drawn i.i.d. from $\QQ$, the resulting empirical risk minimizer $\widebar{f}$ will have excess risk given by
\begin{align*}
\|\widebar{f} - \fstar \|_{\QQ_\Covariate} \lesssim \ocrit_\totobs(\QQ_\Covariate) + s_\totobs(\parprob),
\end{align*}
with high probability. Comparing this with our guarantee above we have established that the estimation guarantee matches that of the standard empirical risk minimizer without transfer learning with further additive terms.

\vspace{3pt}
\noindent\textbf{Cost of transfer:} As a consequence of the above, the cost of transfer learning is governed by the $\chi^2$ divergence between $\SynDist$ and $\CondDist{1}$, rather than the TV-distance. In some ways this divergence is nicer to work with than the TV-distance as its behavior is smoother when $\SynDist$ is close to $\CondDist{1}$, which is important when considering localized empirical processes. Note the relationship $ \dTV{\SynDist}{\CondDist{1}} \leq \frac{1}{2}\sqrt{\chi^2(\SynDist;\, \CondDist{1})} $.

\section{Concrete instantiations}
\label{sec:further-results}

In this section, we present various results that extend~\Cref{thm:slow-rates}. Here we provide guarantees for several different choices of $\SynDist$, illustrating the breadth of our results.

\subsection{Guarantees for specific $\SynDist$}
\label{sec:instantiations}

\Cref{thm:slow-rates} was presented as a meta-result, agnostic to the choice $\SynDist$ and presented a general guarantee that involved the accuracy of the estimation $\dTV{\SynDist}{\CondDist{1}}$. In this section, we instantiate our general guarantee for several different approaches for constructing $\SynDist$. The first two approaches that we present, density estimation and diffusion sampling, already have a body of literature dedicated to establishing TV-distance guarantees. For the remainder of this section, we assume that we are able to choose $J = \ceil{(2\cprob_0 -1)\numobs}$. Consequently, ignoring higher order terms,~\Cref{thm:slow-rates} implies
\begin{align*}
\EE_\QQ\left[ \loss(\Response, \fhat(\Covariate)) - \loss(\Response, \fstar(\Covariate)) \right] \lesssim \RadComp_{\numobs + J}(\QQ) + \dTV{\SynDist}{\CondDist{1}},
\end{align*}
with high probability.
Thus we focus on directly providing upper bounds on the TV-distance $\dTV{\SynDist}{\CondDist{1}}$. The next two approaches, bootstrapping and SMOTE, involve sampling from discrete $\SynDist$; directly applying~\Cref{thm:slow-rates} will result in vacuous bounds if $\CondDist{1}$ is continuous, because the TV-distance between discrete and continuous distributions is always $1$. Thus, we will need to slightly rework our proofs to accommodate these approaches.

\subsubsection{Continuous $\SynDist$}

\vspace{3pt}
\noindent\textbf{Density estimation:} A classical approach would be to estimate the density of $\CondDist{1}$ by using $\dataset_1$, i.e.,~the density of $\Covariate$ for the samples which have $\Response = 1$. This can be achieved through many methods. One such example is kernel density estimation; this particular instantiation was proposed in~\citet{gao2012probability}. If the density $d\CondDist{1}$ is supported on the unit ball $\BB_d(0,1) \subset \RR^d$ and has finite $(\beta, 2)$-Sobolev norm, then we have~\cite{Tsybakov2008IntroductionTN}
\begin{align*}
\EE_{\dataset_1}\left[\dTV{\SynDist}{\CondDist{1}}\right] \lesssim \numobs_1^{-\frac{\beta}{2\beta + d}}.
\end{align*}
Given this estimate $d\SynDist$ of the density, we can draw synthetic samples $\SCovariate \sim \SynDist$.

\vspace{5pt}
\noindent\textbf{Diffusion sampling:} There has been much interest in diffusion-based approaches for sampling in a generative context. It is especially powerful in settings where the features are highly complex, such as images. See~\cite{yang2025diffusionmodelscomprehensivesurvey} for a summary of this literature, but here we provide a (very) brief overview of the denoising diffusion probabilistic model (DDPM) sampler. The objective is to sample from some distribution $\CondDist{1}$ on $\RR^d$. Suppose we draw some initial $\Covariate_0 \sim \CondDist{1}$, and define the recursive process (known as the \emph{forward process})
\begin{align*}
\Covariate_t = \sqrt{1 - \beta_t} \cdot \Covariate_{t-1} + \sqrt{\beta_t} \cdot Z_t, \quad \text{where} \quad Z_t \stackrel{i.i.d.}{\sim} \normal(0, \mathbf{I}_d),\;  \beta_t \in (0, 1), \quad \text{for}\;\; t = 1, \ldots, T.
\end{align*}
Eventually, for large enough $T$, we have $\Covariate_T \approx \normal(0, \mathbf{I}_d)$. Given access to the score functions 
$
s_t^*(x) \coloneqq \nabla \log d\PP_{\Covariate_t}(x)
$
we can construct the \emph{reverse process} by drawing $W_T \sim \normal(0, \mathbf{I}_d)$ and recursively defining
\begin{align}
\label{eqn:reverse-process}
W_{t-1} = \frac{1}{\sqrt{1 - \beta_t}} \left( W_t + \eta_t s_t^*(W_t) + \sigma_t Z_t' \right) \quad \text{where} \quad Z_t' \stackrel{i.i.d.}{\sim} \normal(0, \mathbf{I}_d).
\end{align}
The end of this process will result in $W_0 \sim \SynDist$ with $\SynDist \approx \CondDist{1}$. 

Of course, we typically will not have access to the true score functions and instead have estimates $\widehat{s}_t$ of $s^*_t$. In this case, our error will depend on the estimation error defined as
\begin{align*}
\Delta^2_{\text{score}} \coloneqq \frac{1}{T} \sum_{t=1}^T \EE\left[\| \widehat{s}_t(\Covariate_t) - s^*_t(\Covariate_t) \|_2^2 \right].
\end{align*}
\citet{li2025odt} state that this process (under regularity conditions) will have error  
\begin{align*}
\dTV{\SynDist}{\CondDist{1}} \lesssim  \frac{d\log^3 T}{T} + \Delta_{\text{score}} \cdot \sqrt{\log T},
\end{align*}
where $\SynDist$ is the distribution of $W_0$ from~\Cref{eqn:reverse-process} using the estimated score functions. 

\subsubsection{Discrete $\SynDist$}
\label{sec:discrete-synthetic}

As mentioned previously, since both SMOTE and bootstrapping result in a discrete $\SynDist$, the resulting TV-bound will be vacuous. Here we state results that follow from modifications of the proofs to avoid introducing TV-distances.

\vspace{3pt}
\noindent\textbf{Bootstrapping:} The idea behind bootstrapping (also known as random oversampling in the computer science literature) is to use
\begin{align*}
\SynDist = \widehat{\PP}_{\dataset_1}\coloneqq \frac{1}{\numobs_1} \sum_{\Covariate_j \in \dataset_1} \Delta_{\Covariate = \Covariate_j}.
\end{align*}
To draw from $\SynDist$, we would sample $\SCovariate$ uniformly with replacement from the covariates in $\dataset_1$. Ignoring higher order terms, we have, with high probability,
\begin{align}
\begin{split}
\label{eqn:bootstrapping}
\EE_{\QQ}\left[\loss(\Response, \fhat(\Covariate)) - \loss(\Response, \fstar(\Covariate)) \right] &\lesssim \RadComp_{\numobs + J}(\QQ) + \left(1 + \sqrt{\tfrac{\numobs_1}{J} \log(2\numobs_1)}\right)^2 \cdot \RadComp_{\numobs_1}(\CondDist{1}) 
\end{split}
\end{align}
See~\Cref{sec:proof-instantiations} for a formal statement and proof. Thus, the cost of transfer is given by
\begin{align}
\label{eqn:cost-bootstrapping}
\left(1 + \sqrt{\tfrac{\numobs_1}{J} \log(2\numobs_1)}\right)^2 \cdot \RadComp_{\numobs_1}(\CondDist{1}).
\end{align}
Furthermore, since
\begin{align}
\label{eqn:rad-relations}
\RadComp_{\numobs_1}(\CondDist{1}) \lesssim \RadComp_{2\numobs_1}(\QQ),\quad \text{and} \quad \RadComp_{\numobs+J}(\QQ) \lesssim \RadComp_{2\numobs_1}(\QQ),
\end{align}
we can show
\begin{align*}
\EE_{\QQ}\left[\loss(\Response, \fhat(\Covariate)) - \loss(\Response, \fstar(\Covariate)) \right] &\lesssim \left(1 + \sqrt{\tfrac{\numobs_1}{\numobs_0 - \numobs_1} \log(2\numobs_1)}\right)^2 \cdot \RadComp_{2\numobs_1}(\QQ),
\end{align*}
which implies that the overall error of bootstrapping is, up to logarithmic factors, no worse than the error of performing empirical risk minimization on the target distribution $\QQ$ with an effective sample size of $2\numobs_1$ observations. See~\Cref{sec:proof-rad-relations} for a proof of this result.

\vspace{3pt}
\noindent\textbf{SMOTE:} SMOTE generates a synthetic sample $\SCovariate \sim \widehat{\PP}_\Covariate^{\text{SMOTE}}$ by first sampling uniformly from $\dataset_1$ to obtain $\Covariate_j$, finding the $k$-nearest neighbors of $\Covariate_j$, and returning $\SCovariate$ by randomly selecting one of these nearest neighbors and choosing a random point on the line segment between $\Covariate_j$ and its selected neighbor.

For our guarantees on SMOTE, we require the following Lipschitz assumption made in the literature~\cite{ahmad2025concentration, lyu2025bias}, which is somewhat restrictive for various reasons, especially in modern machine learning algorithms like tree-based methods (e.g., random forests, boosting) and neural networks. We require that for all $\covariate, \covariate' \in \text{supp}(\Covariate) \subset \RR^d$,
\begin{align}
\label{eqn:SMOTE-lipschitz}
\sup_{f \in \fclass} \left| \loss(1, f(\covariate)) - \loss(1, f(\covariate')) \right| \leq L\, \|\covariate - \covariate'\|_2,
\end{align}
for some constant $L > 0$, where $\|\cdot\|_2$ denotes the Euclidean norm.\footnote{
Technically, neural networks with Lipschitz activations also satisfy this
assumption. For a network with $\ell$ hidden layers and weight
matrices $W_1,\dots,W_{\ell+1}$, the global Lipschitz constant is bounded by 
$
\prod_{j=1}^{\ell+1} \|W_j\|_2
$.
Without explicit spectral-norm (or similar) constraints, this product can grow
at least exponentially in $\ell$, and often also grows strongly with the width,
making any bound that scales linearly with $L$ essentially vacuous; see,
e.g., ~\cite{golowich2018size} for a discussion of Lipschitz
constants via norm products.}  See the next section for further discussion of this assumption. Further we also require that for some constant $D > 0$, we have $\norm{\Covariate}{2} \leq D$ almost surely.

Ignoring higher order terms, under~\Cref{eqn:SMOTE-lipschitz} and the conditions of~\Cref{thm:slow-rates}, we have, with high probability
\begin{align}
\begin{split}
\label{eqn:SMOTE-bound}
&\EE_{\QQ}\left[\loss(\Response, \fhat(\Covariate)) - \loss(\Response, \fstar(\Covariate)) \right] \\
&\qquad \lesssim \RadComp_{\numobs + J}(\QQ)+ \left(1 + \sqrt{\tfrac{\numobs_1}{J} \log(2\numobs_1)}\right)^2 \cdot \RadComp_{\numobs_1}(\CondDist{1}) + LD \left( 6\left( \tfrac{k}{\numobs_1} \right)^{1/d} + \tfrac{k\cdot 5^d + 1}{\sqrt{\numobs_1}} \right) \\
&\qquad \lesssim \left(1 + \sqrt{\tfrac{\numobs_1}{\numobs_0 - \numobs_1} \log(2\numobs_1)}\right)^2 \cdot \RadComp_{2\numobs_1}(\QQ) + LD \left( 6\left( \tfrac{k}{\numobs_1} \right)^{1/d} + \tfrac{k\cdot 5^d + 1}{\sqrt{\numobs_1}} \right).
\end{split}
\end{align}
See~\Cref{sec:proof-instantiations} for a proof of this result. We remark that the exponential dependence on $d$ can be refined further using modern results in high-dimensional geometry; for $k=1$ we can replace it with $(\frac{3}{2})^d + 1$. We conclude that the error in the SMOTE procedure matches that of the bootstrapping~\eqref{eqn:bootstrapping}, but with an additional 
\emph{curse of dimensionality} term $(\frac{k}{\numobs_1})^{1/d}$. Quantities of this type occur frequently in analyzing the error of nearest-neighbor procedures, among which SMOTE is an example. If $d$ is large, this is a rather slow rate, and the complexity $\RadComp_{\numobs_1}(\QQ)$ will often decay much faster than that for well-known function classes. In these settings, the error induced by SMOTE estimation will dominate.

\subsubsection{Practical takeaways for rebalancing}

Here we provide some guidelines that can be inferred from our theory for rebalancing. Our bounds indicate that, except in low dimensions, bootstrapping enjoys a more favorable dependence on $d$ than SMOTE, while relying on milder assumptions. In such regimes, our theory therefore suggests that bootstrapping will often be a more attractive choice in practice. If, despite this, the practitioner still chooses to use SMOTE, they should consider algorithms that result in Lipschitz-continuous predictors, and not discontinuous ones such as decision trees and random forests. A practitioner may consider using density estimation procedures if they have strong information on the structure of $\CondDist{1}$ such as a parametric form or smoothness assumptions, or if they can incorporate prior information in some manner. Finally, diffusion-based models can be considered if one can obtain estimates of the score function that are high quality. Our guarantees for undersampling (cf.~\Cref{sec:undersampling-methods}) match bootstrapping up to logarithmic factors; the preference between the two approaches should depend on the context. In settings where computation is the constraint one should likely prefer undersampling; when small sample sizes are the limiting factor, one should consider bootstrapping as it is more sample efficient.

\subsubsection{Localization and bootstrapping}
\label{sec:localization-boot}

The localization-based guarantees of~\Cref{thm:fast-rates} only apply to continuous $\SynDist$; to analyze discrete choices we need to rework the proof slightly. Here we focus on providing localized rates for the bootstrapping approach and do not concern ourselves with SMOTE as the results in~\Cref{sec:discrete-synthetic} suggest that even with localization it will be too slow to be interesting or useful. Ignoring higher order terms, for bootstrapping, we can establish the guarantee
\begin{align}
\begin{split}
\label{eqn:bootstrapping_fast}
\norm{\fhat - \fstar}{\QQ_\Covariate} & \lesssim \ocrit_\totobs(\QQ_\Covariate) + \left(1 + \sqrt{\tfrac{\numobs_1}{J} \log(2\numobs_1)}\right)^2 \cdot \ocrit_{\numobs_1}(\CondDist{1}),
\end{split}
\end{align}
with high probability. See~\Cref{sec:proof-localization-boot} for a formal statement and proof. As we can see from~\Cref{eqn:bootstrapping_fast}, the cost of transfer here is given by $\left(1 + \sqrt{\tfrac{\numobs_1}{J} \log(2\numobs_1)}\right)^2 \cdot \ocrit_{\numobs_1}(\CondDist{1})$, which is the localized analogue of the bootstrapping cost of transfer term in~\Cref{eqn:cost-bootstrapping}.

\subsection{Comparison with previous work}
\label{sec:comparison}

Here we provide a more detailed comparison with the papers~\cite{lyu2025bias, ahmad2025concentration}.

\vspace{3pt}
\noindent\textbf{Slow and fast rates:} Our paper provides two main guarantees,~\Cref{thm:slow-rates}, which establishes slow rates, and~\Cref{thm:fast-rates}, which establishes fast rates. As an example, consider the case where $\fclass$ is a smooth parametric function class with $p$ parameters; it is well known that $\RadComp_{\numobs}(\fclass) \lesssim \sqrt{\frac{p}{\numobs}}$~\cite{dinosaur2019}. When strong convexity~\eqref{eqn:strong-convex} holds,~\Cref{thm:slow-rates} and~\Cref{thm:fast-rates} imply
\begin{align*}
\|\fhat - \fstar \|_{\QQ_\Covariate}^2 \lesssim \EE_\QQ\left[ \loss(\Response, \fhat(\Covariate)) - \loss(\Response, \fstar(\Covariate)) \right] \lesssim \sqrt{\frac{p}{\numobs}}, \quad \text{and} \quad \|\fhat - \fstar \|_{\QQ_\Covariate}^2 \lesssim \frac{p}{\numobs},
\end{align*}
respectively, with high probability (omitting the cost of transfer and higher order terms). Thus, our guarantee in~\Cref{thm:fast-rates} complements the slow-rate bounds available in prior work~\cite{ahmad2025concentration, lyu2025bias}.

\vspace{3pt}
\noindent\textbf{Lipschitz assumption:} Both papers require the Lipschitz assumption~\eqref{eqn:SMOTE-lipschitz}
throughout their analysis, which effectively requires that the predictors $f \in \fclass$ be globally Lipschitz in the covariates. By contrast,~\Cref{thm:slow-rates} only assumes that the loss is uniformly bounded, and~\Cref{thm:fast-rates} assumes that the loss is Lipschitz in its second argument~\eqref{eqn:lipschitz}, but places no global Lipschitz constraint on the covariates. Such global Lipschitz requirements are natural for some classes, such as generalized linear models with smooth link functions or certain shape-constrained estimators, but may be less compatible with modern highly non-linear models.

\vspace{3pt}
\noindent\textbf{Further comparisons~\cite{lyu2025bias}:}  This paper approaches the setting with the same level of abstractness as we do, being agnostic to the choice of synthetic data generating mechanism, but there are several differences between their assumptions and objectives and ours. The main difference is that their final excess risk guarantee includes the bias term
\begin{align*} 
\left| \EE_{\CondDist{1}}\left[ \loss(1, f(\Covariate)) \right] - \EE_{\CondDist{0}}\left[ \loss(0, f(\Covariate)) \right] \right|.
\end{align*}
We remark that this term is a population quantity and, consequently, does not decay to zero in the limit of infinite samples. Finally, although they also consider a similar menu of synthetic distributions $\SynDist$, our analysis provides concrete excess-risk guarantees for several specific constructions as corollaries of the general theory. This allows us to compare different rebalancing schemes within a unified framework and to provide guidance on which choices of synthetic generator are expected to perform better in various regimes.

\vspace{3pt}
\noindent\textbf{Further comparisons~\cite{ahmad2025concentration}:} This paper focuses mainly on a variant of SMOTE and density estimation for choices of $\SynDist$, which can be viewed as a special case of our more general framework. In particular, they use sample-splitting where $I_1$ is not used to learn $\fhat$ and thus the minority class data enter only through the synthetic samples. Sample splitting is a common theoretical device to simplify the analysis as it abstracts away complicated dependencies between different parts of any algorithm. However, in practice it can reduce sample efficiency and, especially in nonparametric settings, may lead to worse performance. They also require the additional assumption on the distribution of $\Covariate$:
\begin{align}
\label{eqn:SMOTE-lower-bound}
\PP\left(\|\Covariate - x \|_2 \leq r  \mid \Response = 1 \right) \geq \min\{c_dr^d, 1\},
\end{align}
for some constant $c_d$ dependent on dimension. If the density $d\CondDist{1}$ was lower bounded by  $f_{\min}$, then $c_d = f_{\min} \cdot V_d(1)$ where $V_d(1)$ is the volume of the unit ball in $\RR^d$. These distributional assumptions effectively require a lower bound on the density of $\Covariate$ in a neighborhood of each point, which can be viewed as a form of truncation of the covariate distribution. With this assumption, their cost of transfer and our cost of transfer are
\begin{align*}
\left(\frac{\max\{k, d\log(\numobs_1)\}}{c_d \numobs_1} \right)^{1/d} \qquad \text{versus} \qquad LD\left(\left( \frac{k}{\numobs_1} \right)^{1/d} + \frac{k\cdot 5^d + 1}{\sqrt{\numobs_1}} \right),
\end{align*}
respectively. In particular, for densities supported on a ball, $1/c_d$ can scale exponentially in $d$, so our bound exhibits a different dependence on the dimension and does not require~\Cref{eqn:SMOTE-lower-bound}. In addition, their analysis replaces the distribution-specific Rademacher complexities by a single generic complexity term. That is, in our notation, they assume that $\RadComp_{n}(\CondDist{0})$, $\RadComp_{n}(\CondDist{1})$, and $\RadComp_{n}(\SynDist)$ are all bounded above by a common quantity that appears in their guarantees. By contrast, we explicitly relate these three complexities to $\RadComp_{n}(\QQ)$, which is necessary for expressing our results in a transfer-learning formulation.

\section{Plug-in approaches}
\label{sec:plugin}

We can also exploit the transfer-learning formulation to derive a simple estimator of $\fstar$. To see how, observe that we can write
\begin{align}
\label{eqn:transform}
\fstar(\Covariate) = \QQ(\Response = 1 \mid \Covariate) = \frac{\cprob_0}{\cprob_0 + \cprob_1 \big( \frac{1}{\PP(\Response = 1 \mid \Covariate)} - 1 \big)},
\end{align}
taking the convention that $1/\infty = 0$.
This suggests an alternative estimator for $\fstar$: we first use $\dataset$ to obtain an estimate of $\PP(\Response = 1 \mid \Covariate)$ and then form a plug-in estimator of $\fstar$ using~\Cref{eqn:transform}. Using this approach confers several advantages over the previous one. Most notably, in settings where $\Covariate$ is highly complex (e.g., text or image data), being able to generate $\SCovariate \sim \SynDist$ may be quite difficult without extensive resources and expertise. This plug-in estimator is a simple modification that circumvents many of the challenges associated with rebalancing approaches.

We consider the estimator that is implied by~\Cref{eqn:transform}. For convenience, we use $\gstar(\Covariate) \coloneqq \PP(\Response = 1 \mid \Covariate)$ and let $\ghat$ be the estimate of $\gstar$ that is obtained from running any classification algorithm on $\dataset = \{(\Covariate_i, \Response_i)\}_{i=1}^\numobs$. We can obtain the plug-in estimator of $\fstar$:
\begin{align}
\label{eqn:PLUG}
\fhat_{\text{PLUG}}(\Covariate) \coloneqq \frac{\cprob_0}{\cprob_0 + \cprob_1\left( \frac{1}{\ghat(\Covariate)} - 1 \right)}.
\end{align}
For convenience, we assume that $(\cprob_0, \cprob_1)$ is known, but the results can be extended to when they are estimated. The following bound then relates the error $\fhat_{\text{PLUG}}$ to the error in $\ghat$. 
\begin{proposition}
\label{prop:plug-in}
The estimator $\fhat_{\text{PLUG}}$ has estimation error bounded by
\begin{align}
\label{eqn:l2-plugin}
\big\|\fhat_{\text{PLUG}} - \fstar \big\|_{\QQ_\Covariate} \leq \left( \frac{\sqrt{\cprob_0/2}}{\cprob_1} \right) \cdot \left\| \ghat - \gstar \right\|_{\PP_\Covariate} \quad \text{a.s.}
\end{align}
\end{proposition}

See~\Cref{sec:proof-plugin} for a proof. This result implies that if we have a sufficiently good estimate $\ghat$ of $\gstar$, measured in the source distribution $\PP_\Covariate$, we should expect a good estimate $\fhat_{\text{PLUG}}$ of $\fstar$, measured in the target distribution $\QQ_\Covariate$. The plug-in approach has the benefit of being computationally very simple: just train a classifier on the original data and then use~\Cref{eqn:PLUG}. This is especially useful in areas where $\Covariate$ is highly complex such as text or images, so rebalancing methods may be infeasible to implement. However the dependence on $\cprob_1$ is unfavorable compared to rebalancing methods; in parametric settings we have that the above rate implies $\norm{\fhat_{\text{PLUG}} - \fstar}{\QQ_\Covariate} \lesssim \frac{1}{\sqrt{\cprob_1 \numobs_1}}$, as opposed to~\Cref{thm:fast-rates} which typically would result in a $\frac{1}{\sqrt{\numobs_1}}$ rate. This is the tradeoff between the simplicity of the plug-in method and the accuracy of the rebalancing approach.

\section{Numerical experiments}
\label{sec:sims}

\newcommand{\matI}{\mathbf{I}}

Here we present some simulations illustrating the behavior described by our guarantees. We focus on comparing SMOTE and bootstrapping. Let $\mathbf{0}_d, \mathbf{1}_d \in \RR^d$ denote the all-zeros and all-ones vectors, respectively, in $\RR^d$. We set
\begin{align*}
\CondDist{0} = \normal\left(\mathbf{0}_d, \matI_d\right) \qquad \text{and} \qquad \CondDist{1} = \normal\left(\tfrac{1}{\sqrt{d}} \mathbf{1}_d, \matI_d \right).
\end{align*}
The rescaling by $\frac{1}{\sqrt{d}}$ in $\CondDist{1}$ is to ensure that the signal-to-noise ratio remains constant even when varying dimension. We then choose $\cprob_0 = 0.9$ and generate $\dataset = \{(\Covariate_i, \Response_i)\}_{i=1}^\numobs$ according to~\Cref{eqn:observed}. Afterwards, we run both SMOTE and bootstrapping using linear logistic regression to obtain the estimate $\fhat_{\text{SMOTE}}$ and $\fhat_{\text{BOOT}}$ (i.e., they are implemented using logistic regression). This is repeated for several different choices of the pair $(\numobs, d)$. We finally report the ratio of the excess risks to compare the performance of the two approaches. 

\begin{figure}[t]
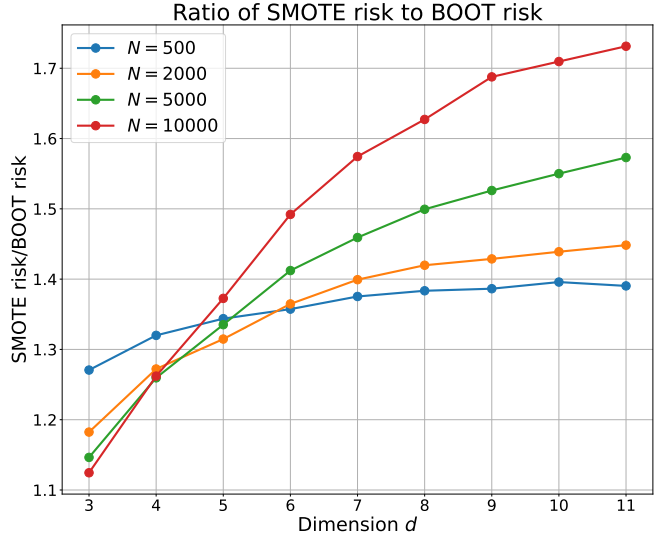

  \begin{center}
    \widgraph{0.65\textwidth}{results}
    \caption{Illustration of the ratio of the excess risk of SMOTE and bootstrapping. The $x$-axis shows the underlying dimension $d$, and the different colored lines represent different sample sizes. As we can see,
    for a fixed sample size, increasing $d$ worsens the relative performance of SMOTE compared to bootstrapping. For moderately large $d$, increasing $\numobs$ will also worsen the relative performance.}
\label{FigResult}    
  \end{center}
\end{figure}

Since we are performing logistic regression over a parametric class, we have $\RadComp_{\numobs} \lesssim \sqrt{\frac{d}{\numobs}}$. Thus, our theory suggests
\begin{align*}
\frac{\EE_\QQ\left[\loss(\Response, \fhat_{\text{SMOTE}}(\Covariate)) - \loss(\Response, \fstar(\Covariate)) \right]}{\EE_\QQ\left[\loss(\Response, \fhat_{\text{BOOT}}(\Covariate)) - \loss(\Response, \fstar(\Covariate)) \right]} \approx \frac{\big(\frac{1}{\numobs_1}\big)^{1/d}}{\big(\frac{1}{\numobs_1}\big)^{1/2}} = (\numobs_1)^{\frac{1}{2} - \frac{1}{d}}.
\end{align*}
Therefore, for a fixed but moderately large $d$, we should expect the relative excess risk between SMOTE and bootstrapping to grow as we increase $\numobs_1$, and similarly for a fixed $\numobs_1$ we should expect the relative risk to also grow as we increase $d$. The excess risk ratio is plotted in Figure~\ref{FigResult}, and the illustrated behavior is exactly as described. We also remark that in virtually all the simulations, $\fhat_{\text{SMOTE}}$ had larger excess risk than that of $\fhat_{\text{BOOT}}$, and that on average it often is much larger. Our simulations are consistent with the theoretical prediction that the excess risk for SMOTE grows more quickly with dimension than that of bootstrapping.

\section{Discussion}
\label{sec:discussion}

Here we have studied classification imbalance, casting it as a problem of transfer learning. This formulation allowed us to design a class of estimators that we refer to as plug-in estimators. Furthermore, we provided general guarantees on rebalancing methods, agnostic to the choice of synthetic data generation. Using our general theory, we obtain concrete guarantees for a variety of choices of synthetic data generation. When instantiated for SMOTE, we can establish guarantees that are stronger than existing results. Because of these results, we can provide theory-driven recommendations on which procedures to use. Unless the data is low-dimensional, our results suggest one should prefer bootstrapping to SMOTE. If the practitioner has additional structure to incorporate into a model for $\CondDist{1}$, they can look to density estimation, or if they have good estimates of the score functions, they can consider diffusion models. 

There are several interesting future directions to consider. Transfer learning is an active and growing field of statistics. One could incorporate more modern approaches for transfer learning in classification imbalance to see if there are any benefits. Another avenue that may be of interest is developing approaches to estimating the cost of transfer. Our guarantees make a strong case against using SMOTE but there may be hidden structure that our rates do not capture. Being able to adaptively identify which methods have better or worse cost of transfer would be especially beneficial to practitioners. 

\section{Acknowledgments}

Klusowski would like to thank Jacob Bien, Paromita Dubey, and Zijun Gao for insightful comments. Klusowski is grateful for support from the National Science Foundation through NSF CAREER DMS-2239448 and the Alfred P. Sloan Foundation through a Sloan Research Fellowship.

\section{Supplement}

The supplement~\cite{proofs} contains additional proofs, auxiliary lemmas, and extensions (including data-dependent choices of 
$J$, general target mixtures, and undersampling results), as well as further technical details omitted from the main text for readability.

\section{Proofs}
\label{sec:proofs}

\subsection{Proof of~\Cref{thm:slow-rates}}
\label{sec:proof-slow-rates}

Recall that $\fstar = \argmin_{f \in \fclass} \EE_\QQ[\loss(\Response, f(\Covariate))]$ and
\begin{align*}
&\fhat \in \argmin_{f \in \fclass}  \left\{\frac{1}{\numobs + J} \left( \sum_{i=1}^\numobs \loss(\Response_i, f(\Covariate_i)) + \sum_{j=1}^J \loss(1, f(\SCovariate_j)) \right) \right\}.
\end{align*}
By definition of $\fhat$ we have the basic inequality
\begin{align*}
\frac{1}{\numobs + J} \left( \sum_{i=1}^\numobs \loss(\Response_i, \fhat(\Covariate_i)) + \sum_{j=1}^J \loss(1, \fhat(\SCovariate_j)) \right) \leq \frac{1}{\numobs + J} \left( \sum_{i=1}^\numobs \loss(\Response_i, \fstar(\Covariate_i)) + \sum_{j=1}^J \loss(1, \fstar(\SCovariate_j)) \right).
\end{align*}
A standard empirical risk minimization argument gives
\begin{align*}
&\EE_\QQ[\loss(\Response, \fhat(\Covariate)) - \loss(\Response, \fstar(\Covariate))] \\
&\qquad \qquad \leq \EE_\QQ[\loss(\Response, \fhat(\Covariate))] - \frac{1}{\numobs + J} \left( \sum_{i=1}^\numobs \loss(\Response_i, \fhat(\Covariate_i)) + \sum_{j=1}^J \loss(1, \fhat(\SCovariate_j)) \right) \\
&\qquad \qquad \qquad + \frac{1}{\numobs + J} \left( \sum_{i=1}^\numobs \loss(\Response_i, \fstar(\Covariate_i)) + \sum_{j=1}^J \loss(1, \fstar(\SCovariate_j)) \right) - \EE_\QQ[\loss(\Response, \fstar(\Covariate))] \\
&\qquad  \qquad \leq 2\sup_{f \in \fclass} \left|\frac{1}{\numobs + J} \left( \sum_{i=1}^\numobs \loss(\Response_i, f(\Covariate_i)) + \sum_{j=1}^J \loss(1, f(\SCovariate_j)) \right) - \EE_\QQ[\loss(\Response, f(\Covariate))] \right|.
\end{align*}
Note that if $\SCovariate \sim \PP_{\Covariate \mid \Response = 1}$ and were independent of $\dataset = \{(\Covariate_i, \Response_i)\}_{i=1}^\numobs$, we could directly apply a concentration inequality (cf.~\Cref{lem:bounded-diff}) and arrive at the desired result. The remainder addresses this issue by theoretically drawing samples and controlling the differences.

To address the fact that $\SCovariate \sim \SynDist$, we use a coupling argument.
\begin{lemma}
\label{lem:coupling}
Consider two distributions $\PP_1$ and $\PP_2$ on some measurable space $(\Omega, \mathcal{S})$. Given a random draw $U_1 \sim \PP_1$, we can construct a random variable $U_2$ such that it has marginal $U_2 \sim \PP_2$ and joint distribution $(U_1, U_2) \sim \Pr$ such that
$
\Pr(U_1 \neq U_2) = \dTV{\PP_1}{\PP_2}.
$
\end{lemma}
\noindent See~\Cref{sec:proof-coupling} for a proof of this result. Using this lemma we can (theoretically) construct samples $\{\BCovariate_j\}_{j=1}^J$ from $\{\SCovariate_j\}_{j=1}^J$ with marginal distribution $\CondDist{1}$ and the pairs $(\SCovariate_j, \BCovariate_j) \stackrel{i.i.d.}{\sim} \Pr$ defined in~\Cref{lem:coupling}. Thus, we have
\begin{align*}
&\EE_\QQ[\loss(\Response, \fhat(\Covariate)) - \loss(\Response, \fstar(\Covariate))] \\
&\qquad \qquad \leq 2\sup_{f \in \fclass} \left|\frac{1}{\numobs + J} \left( \sum_{i=1}^\numobs \loss(\Response_i, f(\Covariate_i)) + \sum_{j=1}^J \loss(1, f(\BCovariate_j)) \right) - \EE_\QQ[\loss(\Response, f(\Covariate))] \right| \\
&\qquad \qquad \qquad \qquad + 2\sup_{f \in \fclass} \left|\frac{1}{\numobs + J} \sum_{j=1}^J \loss(1, f(\SCovariate_j)) - \frac{1}{\numobs + J} \sum_{j=1}^J \loss(1, f(\BCovariate_j) ) \right|.
\end{align*}
If the way we constructed $\SynDist$ is independent of $\dataset$, we could directly apply~\Cref{lem:bounded-diff} to the first term. However, since there is dependence between $\SynDist$ and $\dataset$, and dependence between $\BCovariate_j$ and $\SCovariate_j$, there will potentially be dependence between $\{\BCovariate_j\}_{j=1}^J$ and $\dataset$.  To address this, we will (theoretically) sample $\{\Covariate_j'\}_{j=1}^J \stackrel{i.i.d.}{\sim} \CondDist{1}$ in a manner that is independent of everything. Applying the triangle inequality again we get
\begin{align*}
\EE_\QQ[\loss(\Response, \fhat(\Covariate)) - \loss(\Response, \fstar(\Covariate))] \leq 2\, (\Term_1 + \Term_2 + \Term_3),
\end{align*}
where we have adopted the shorthands
\begin{align*}
\Term_1 &\coloneqq \sup_{f \in \fclass} \left|\frac{1}{\numobs + J} \left( \sum_{i=1}^\numobs \loss(\Response_i, f(\Covariate_i)) + \sum_{j=1}^J \loss(1, f(\Covariate'_j)) \right) - \EE_\QQ[\loss(\Response, f(\Covariate))] \right|, \\
\Term_2 &\coloneqq \sup_{f \in \fclass} \left|\frac{1}{\numobs + J} \sum_{j=1}^J \loss(1, f(\BCovariate_j)) - \frac{1}{\numobs + J} \sum_{j=1}^J \loss(1, f(\Covariate'_j)) \right|, \qquad \text{and} \\
\Term_3 &\coloneqq \sup_{f \in \fclass} \left| \frac{1}{\numobs + J} \sum_{j=1}^J \loss(1, f(\SCovariate_j)) - \frac{1}{\numobs + J} \sum_{j=1}^J \loss(1, f(\BCovariate_j)) \right|.
\end{align*}

The following lemmas control these terms individually; see~\Cref{sec:slow-rate-lemma} for a proof of these lemmas. The claim then follows from putting together the pieces. 

\begin{lemma}
\label{lem:app-bounded-diff}
We have, with probability exceeding $1 - \parprob$,
\begin{align*}
\Term_1 \leq 2\RadComp_{\numobs+J}(\QQ) + (2\sqrt{2\log(1/\parprob)} + 6) \cdot \frac{\bound}{\sqrt{\numobs + J}} + 6\bound \cdot \left| \frac{\numobs\cprob_0}{\numobs + J} - \frac{1}{2} \right|
\end{align*}
\end{lemma}
\begin{lemma}
\label{lem:double-sampling-control}
We have, with probability exceeding $1 - \parprob$,
\begin{align*} 
\Term_2 \leq 4\RadComp_{\numobs + J}(\QQ) + 4(\sqrt{2\log(2/\parprob)} + 3) \cdot \frac{\bound}{\sqrt{\numobs + J}} + 8\bound \cdot \left| \frac{\numobs\cprob_0}{\numobs + J} - \frac{1}{2} \right|
\end{align*}
\end{lemma}

\begin{lemma}
\label{lem:coupling-control}
We have, with probability exceeding $1 - \parprob$,
\begin{align*}
\Term_3 \leq 2\bound \cdot \dTV{\SynDist}{\CondDist{1}} + \sqrt{2\log(1/\parprob)} \cdot \frac{\bound}{\sqrt{\numobs + J}}.
\end{align*}
\end{lemma}

\subsection{Proof of auxiliary lemmas for~\Cref{thm:slow-rates}}
\label{sec:slow-rate-lemma}

\subsubsection{Proof of~\Cref{lem:app-bounded-diff}}
\label{sec:app-bounded-diff}

For convenience, we define the mixture distribution 
\begin{align*}
\QQ' \coloneqq \frac{\numobs}{\numobs + J} \cdot \PP_{\Covariate, \Response} + \frac{J}{\numobs + J} \cdot \CondDist{1} \otimes \Delta_{\Response = 1}.
\end{align*}
Note that $\dTV{\QQ}{\QQ'} \leq |\frac{\numobs\cprob_0}{\numobs + J} - \frac{1}{2}|$, which implies 
\begin{align}
\begin{split}
\label{eqn:QQ-relation}
\sup_{f \in \fclass} |\EE_{\QQ}\left[ \loss(\Response, f(\Covariate)\right] - \EE_{\QQ'} \left[\loss(\Response, f(\Covariate)) \right]| &\leq 2\bound \cdot \left| \frac{\numobs\cprob_0}{\numobs + J} - \frac{1}{2}\right|.
\end{split}
\end{align}
We also require the following lemma; see~\Cref{sec:proof-mixture-rad} for a proof. 
\begin{lemma}
\label{lem:annoying-stuff}
We have
\begin{align*}
\RadComp_{\numobs + J}(\QQ') \leq \RadComp_{\numobs + J}(\QQ)+ \frac{2\bound}{\sqrt{\numobs + J}} + 2\bound \cdot \left| \frac{\numobs \cprob_0}{\numobs + J} - \frac{1}{2} \right|.
\end{align*}
\end{lemma}
\noindent

We can define 
\begin{align*}
g(Z_1, \ldots, Z_{\numobs + J}) \coloneqq \sup_{f \in \fclass} \left| \frac{1}{\numobs + J} \left(\sum_{i=1}^\numobs \loss(\Response_i, f(\Covariate_i)) + \sum_{j=1}^J \loss(1, f(\Covariate_j')) \right)- \EE_\QQ\left[ \loss(\Response, f(\Covariate)) \right] \right|
\end{align*}
where
\begin{align*}
Z_i \coloneqq \begin{cases} (\Covariate_i, \Response_i) & \text{if} \quad i = 1, \ldots, \numobs, \\
(\Covariate_{i-\numobs}', 1) & \text{if} \quad i = \numobs+1, \ldots, \numobs + J. \end{cases}
\end{align*}

Then we can bound (using the fact that the loss $\bound$-uniformly bounded~\eqref{eqn:bound-loss})
\begin{align*}
\sup_{z_i, z_i'} |g(z_1, \ldots, z_{i}, \ldots, z_{\numobs + J}) - g(z_1, \ldots, z_i', \ldots, z_{\numobs+J})| \leq \frac{2\bound}{\numobs + J}.
\end{align*}
Applying~\Cref{lem:bounded-diff} yields, with probability exceeding $1 - \parprob$,
\begin{align*}
\Term_1 &\leq \EE\left[\sup_{f \in \fclass} \left| \frac{1}{\numobs + J} \big( \sum_{i=1}^\numobs \loss(\Response_i, f(\Covariate_i)) + \sum_{j=1}^J \loss(1, f(\Covariate_j')) \big) - \EE_\QQ[\loss(\Response, f(\Covariate))] \right| \right] + \sqrt{\frac{8\bound^2 \log(1/\parprob)}{\numobs + J}} \\
&\leq \EE\left[\sup_{f \in \fclass} \left| \frac{1}{\numobs + J} \big( \sum_{i=1}^\numobs \loss(\Response_i, f(\Covariate_i)) + \sum_{j=1}^J \loss(1, f(\Covariate_j')) \big) - \EE_{\QQ'}[\loss(\Response, f(\Covariate))] \right| \right] \\
&\qquad \qquad + 2\sqrt{2\log(1/\parprob)} \cdot \frac{\bound}{\sqrt{\numobs + J}} + 2\bound \cdot \left| \frac{\numobs\cprob_0}{\numobs + J} - \frac{1}{2}\right|,
\end{align*}
using~\Cref{eqn:QQ-relation} in the second step. 

Controlling the expectation term is considerably more challenging than standard empirical risk minimization analysis as we must relate the quantity to the Rademacher complexity over i.i.d.~draws from $\QQ$, the mixture distribution. First, we show that
\begin{align}
\begin{split}
\label{eqn:symmetrization}
&\EE\left[\sup_{f \in \fclass} \left| \frac{1}{\numobs + J} \left( \sum_{i=1}^\numobs \loss(\Response_i, f(\Covariate_i)) + \sum_{j=1}^J \loss(1, f(\Covariate_j')) \right) - \EE_{\QQ'}[\loss(\Response, f(\Covariate))] \right| \right] \\
&\qquad \qquad \leq 2 \EE\left[ \sup_{f\in\fclass} \left| \frac{1}{\numobs + J} \left( \sum_{i=1}^\numobs \rad_i \cdot \loss(\Response_i, f(\Covariate_i)) + \sum_{j=1}^J \rad_{\numobs + j} \cdot \loss(1, f(\Covariate_j'))\right) \right| \right].
\end{split}
\end{align}
See~\Cref{sec:additional-calcs} for its proof. The challenging step is to relate the above quantity to the Rademacher complexity involving $\QQ$. The following lemma allows us to do so.
\begin{lemma}
\label{lem:mixture-rad}
We have
\begin{align*}
&\EE\left[ \sup_{f\in\fclass} \left| \frac{1}{\numobs + J} \left( \sum_{i=1}^\numobs \rad_i \cdot \loss(\Response_i, f(\Covariate_i)) + \sum_{j=1}^J \rad_{\numobs + j} \cdot \loss(1, f(\Covariate_j'))\right) \right| \right] \\
&\qquad \qquad \leq \EE_{\rad_i \stackrel{i.i.d.}{\sim} \text{Rad},\; (\Covariate_i, \Response_i) \stackrel{i.i.d.}{\sim} \QQ'}\left[\sup_{f \in \fclass} \left| \frac{1}{\numobs + J} \sum_{i=1}^{\numobs + J} \rad_i \cdot \loss(\Response_i, f(\Covariate_i)) \right|\right] + \frac{\bound}{\sqrt{\numobs + J}}.
\end{align*}
\end{lemma}
\noindent See~\Cref{sec:proof-mixture-rad} for a proof of this result; it involves a combinatorial argument about permutations of mixtures distributions.

Using~\Cref{lem:annoying-stuff}, we conclude, with probability exceeding $1 - \parprob$,
\begin{align*}
\Term_1 \leq 2\RadComp_{\numobs+J}(\QQ) + (2\sqrt{2\log(1/\parprob)} + 6) \cdot \frac{\bound}{\sqrt{\numobs + J}} + 6\bound \cdot \left| \frac{\numobs\cprob_0}{\numobs + J} - \frac{1}{2} \right|.
\end{align*}

\subsubsection{Proof of~\Cref{lem:double-sampling-control}}
\label{sec:double-sampling-control}

Recall
\begin{align*}
\Term_2 = \sup_{f \in \fclass} \left|\frac{1}{\numobs + J} \sum_{j=1}^J \loss(1, f(\BCovariate_j)) - \frac{1}{\numobs + J} \sum_{j=1}^J \loss(1, f(\Covariate'_j)) \right|
\end{align*}
where $\{\BCovariate_j\}_{j=1}^J$ and $\{\Covariate_j'\}_{j=1}^J$ are i.i.d.~samples from $\CondDist{1}$. By triangle inequality, we have
\begin{align*}
\Term_2 &\leq \sup_{f \in \fclass} \left|\frac{1}{\numobs + J} \sum_{j=1}^J \left(\loss(1, f(\BCovariate_j)) - \EE_{\CondDist{1}}[\loss(1, f(\Covariate))] \right) \right| \\
&\qquad \qquad + \sup_{f \in \fclass} \left|\frac{1}{\numobs + J} \sum_{j=1}^J \left(\loss(1, f(\Covariate_j')) - \EE_{\CondDist{1}}[\loss(1, f(\Covariate))] \right) \right|.
\end{align*}
Note that it suffices to control one term as the two have the same distribution. By~\Cref{lem:bounded-diff}
\begin{align*}
&\sup_{f \in \fclass} \left|\frac{1}{\numobs + J} \sum_{j=1}^J \left(\loss(1, f(\BCovariate_j)) - \EE_{\CondDist{1}}[\loss(1, f(\Covariate))] \right) \right| \\
& \qquad \leq \EE_{\CondDist{1}}\left[\sup_{f \in \fclass} \left|\frac{1}{\numobs + J} \sum_{j=1}^J \left(\loss(1, f(\Covariate_j)) - \EE_{\CondDist{1}}[\loss(1, f(\Covariate))] \right) \right| \right]  + 2\sqrt{2\log(1/\parprob)} \cdot \frac{\bound}{\sqrt{\numobs + J}}.
\end{align*}
By symmetrization we have
\begin{align*}
&\EE_{\CondDist{1}}\left[\sup_{f \in \fclass} \left|\frac{1}{\numobs + J} \sum_{j=1}^J \left(\loss(1, f(\Covariate_j)) - \EE_{\CondDist{1}}[\loss(1, f(\Covariate))] \right) \right| \right] \\
&\qquad \qquad \leq 2\EE_{\rad_j \stackrel{i.i.d.}{\sim} \text{Rad},\;\Covariate_j \stackrel{i.i.d.}{\sim} \CondDist{1}}\left[\sup_{f \in \fclass} \left|\frac{1}{\numobs + J} \sum_{j=1}^J \rad_j \cdot \loss(1, f(\Covariate_j)) \right| \right]   = \frac{2J}{\numobs + J} \cdot \RadComp_{J}(\CondDist{1}).
\end{align*}
Thus, we have
\begin{align*}
\Term_2 \leq \frac{4J}{\numobs + J} \cdot \RadComp_J(\CondDist{1}) + 4 \sqrt{2\log(2/\parprob)} \cdot \frac{\bound}{\sqrt{\numobs + J}}
\end{align*}
with probability exceeding $1 - \parprob$. Our claim then follows from establishing 
\begin{align}
\label{eqn:awesome-jason-lemma}
\frac{J}{\numobs + J} \cdot \RadComp_{J}(\CondDist{1}) \leq \RadComp_{\numobs + J}(\QQ) + \frac{3\bound}{\sqrt{\numobs + J}} + 2\bound \cdot \left|\frac{\numobs\cprob_0}{\numobs + J} - \frac{1}{2} \right|.
\end{align}
See~\Cref{sec:additional-calcs} for its proof.

\subsubsection{Proof of~\Cref{lem:coupling-control}}
\label{sec:proof-coupling-control}

By boundedness, we have via the triangle inequality
\begin{align*}
\Term_3 &= \sup_{f \in \fclass} \left| \frac{1}{\numobs + J} \sum_{j=1}^J \left\{\loss(1, f(\SCovariate_j)) - \loss(1, f(\BCovariate_j)) \right\} \right| \leq 2\bound \cdot \left( \frac{1}{\numobs + J} \sum_{j=1}^{J} \mathbf{1}\{\SCovariate_j \neq \BCovariate_j\} \right).
\end{align*}
\Cref{lem:coupling} implies $\sum_{j=1}^{J} \mathbf{1}\{\SCovariate_j \neq \BCovariate_j\} \sim \text{Bin}\left(J, \dTV{\SynDist}{\CondDist{1}}\right)$, conditional on $\SynDist$, so we have
\begin{align*}
\sum_{j=1}^{J} \mathbf{1}\{\SCovariate_j \neq \BCovariate_j\} \leq J \cdot \dTV{\SynDist}{\CondDist{1}} + \sqrt{\frac{J\log(1/\parprob)}{2}}
\end{align*}
with probability exceeding $1 - \parprob$. Thus we conclude
\begin{align*}
\Term_3 &\leq 2\bound \cdot \dTV{\SynDist}{\CondDist{1}} + \sqrt{2\log(1/\parprob)} \cdot \frac{\bound}{\sqrt{\numobs + J}}.
\end{align*}

\subsection{Proof of~\Cref{thm:fast-rates}}
\label{sec:proof-fast-rates}

From the proof of~\Cref{thm:slow-rates}, we have
\begin{align*}
&\EE_\QQ[\loss(\Response, \fhat(\Covariate)) - \loss(\Response, \fstar(\Covariate))] \\
&\qquad \qquad \leq \EE_\QQ[\loss(\Response, \fhat(\Covariate))] - \frac{1}{\totobs} \left( \sum_{i=1}^\numobs \loss(\Response_i, \fhat(\Covariate_i)) + \sum_{j=1}^J \loss(1, \fhat(\SCovariate_j)) \right) \\
&\qquad \qquad \qquad \qquad  + \frac{1}{\totobs} \left( \sum_{i=1}^\numobs \loss(\Response_i, \fstar(\Covariate_i)) + \sum_{j=1}^J \loss(1, \fstar(\SCovariate_j)) \right) - \EE_\QQ[\loss(\Response, \fstar(\Covariate))].
\end{align*}
Our approach here differs from the previous one in that we condition on the responses $\{\Response_i\}_{i=1}^\numobs$ and do not invoke a coupling argument. For brevity, we introduce the shorthand
\begin{align*}
\ExcessLoss{f}(\Response, \Covariate) \coloneqq \loss(\Response, f(\Covariate)) - \loss(\Response, \fstar(\Covariate)).
\end{align*}

Draw samples $\{\Covariate_j'\}_{j=1}^J \stackrel{i.i.d.}{\sim} \CondDist{1}$; we can then write 
\begin{align*}
\frac{\gamma}{2} \|\fhat - \fstar \|_{\QQ_\Covariate}^2 \leq \EE_\QQ[\loss(\Response, \fhat(\Covariate)) - \loss(\Response, \fstar(\Covariate))] \leq \Term_1 + \Term_2 
\end{align*}
where we have used the shorthands
\begin{align*}
\Term_1 &\coloneqq -\frac{1}{\totobs} \left( \sum_{i=1}^\numobs  \ExcessLoss{\fhat}(\Response_i, \Covariate_i) + \sum_{j=1}^J \ExcessLoss{\fhat}(1, \Covariate_j')\right) + \EE_\QQ\left[ \ExcessLoss{\fhat}(\Response, \Covariate) \right], \\
&\qquad \text{and} \qquad \Term_2 \coloneqq \frac{1}{\totobs} \sum_{j=1}^J \left( \ExcessLoss{\fhat}(1, \Covariate_j') - \ExcessLoss{\fhat}(1, \SCovariate_j) \right).
\end{align*}
We can then write
$
\Term_1 \leq \STerm + \STerm_\EE
$
where
\begin{align*}
\STerm &\coloneqq \left| \frac{1}{\totobs} \left(\sum_{i \in I_0} \ExcessLoss{\fhat}(0, \Covariate_i) + \sum_{i \in I_1} \ExcessLoss{\fhat}(1, \Covariate_i) + \sum_{j=1}^J \ExcessLoss{\fhat}(1, \Covariate_j') \right)  \right. \\
&\qquad \qquad \qquad \left. - \frac{\numobs_0}{\totobs} \cdot \EE_{\CondDist{0}}\left[\ExcessLoss{\fhat}(0, \Covariate)\right] - \frac{\numobs_1 + J}{\totobs} \cdot \EE_{\CondDist{1}}\left[\ExcessLoss{\fhat}(1, \Covariate)\right]\right|, \\
\STerm_\EE &\coloneqq \left| \frac{\numobs_0}{\totobs} \cdot \EE_{\CondDist{0}}\left[\ExcessLoss{\fhat}(0, \Covariate)\right] + \frac{\numobs_1 + J}{\totobs} \cdot \EE_{\CondDist{1}}\left[\ExcessLoss{\fhat}(1, \Covariate) \right] - \EE_\QQ\left[ \ExcessLoss{\fhat}(\Response, \Covariate) \right]  \right|.
\end{align*}

The following lemma controls $\STerm$. See~\Cref{sec:proof-new-localization} for a proof of this result.
\begin{lemma}
\label{lem:new-localization}
On the event $\norm{\fhat - \fstar}{\QQ_\Covariate} \geq \ocrit_\totobs(\QQ_\Covariate)$ we have, with probability exceeding $1 - \parprob$,
\begin{align*}
\STerm \leq 4L \norm{\fhat - \fstar}{\QQ_\Covariate} \left(16\ocrit_\totobs(\QQ_\Covariate)+ \sqrt{\frac{\log(\logfun(\mcrit_\totobs)/\parprob)}{\totobs}} \right) + \frac{64L \log(\logfun(\mcrit_\totobs)/\parprob)}{\totobs}.
\end{align*}
\end{lemma}
\noindent  Note that if $\numobs_1 + J = \numobs_0$, then $\STerm_\EE = 0$. All that remains is to control $\Term_2$; for convenience, we introduce the shorthands
$
\nuhat \coloneqq \frac{d\SynDist}{d\QQ_\Covariate}$ and $\nustar \coloneqq \frac{d\CondDist{1}}{d\QQ_\Covariate}.
$
\begin{lemma}
\label{lem:new-transfer-localization}
On the event that $\norm{\fhat - \fstar}{\QQ_\Covariate} \geq \max\left\{\ocrit_\totobs(\QQ_\Covariate), \ocrit_\totobs(\SynDist), \norm{\nuhat - \nustar}{\QQ_\Covariate}\right\}$ we have, with probability exceeding $1 - \parprob$,
\begin{align*}
\Term_2 &\leq L\norm{\fhat - \fstar}{\QQ_\Covariate} \left(64 \ocrit_\totobs(\QQ_\Covariate) + 16\ocrit_\totobs(\SynDist) + \norm{\nuhat - \nustar}{\QQ_\Covariate} + 10 \sqrt{\frac{\log(2\logfun(\mcrit_\totobs)/\parprob)}{\totobs}}\right) \\
&\qquad \qquad \qquad \qquad + \frac{96L \log(2\logfun(\mcrit_\totobs)/\parprob)}{\totobs}.
\end{align*}
\end{lemma}
\noindent See~\Cref{sec:new-transfer-localization} for a proof of this result.

Putting the pieces together, on the event $\norm{\fhat - \fstar}{\QQ_\Covariate} \geq \max\{\ocrit_\totobs(\QQ_\Covariate), \ocrit_\totobs(\SynDist), \norm{\nuhat - \nustar}{\QQ_\Covariate}\}$ we have
\begin{align*}
\frac{\gamma}{2} \norm{\fhat - \fstar}{\QQ_\Covariate}^2 &\leq L\norm{\fhat - \fstar}{\QQ_\Covariate} \left(128\ocrit_\totobs(\QQ_\Covariate) + 16\ocrit_\totobs(\SynDist) +  \norm{\nuhat - \nustar}{\QQ_\Covariate} \right) \\
&\qquad \qquad  + 14L \norm{\fhat - \fstar}{\QQ_\Covariate} \cdot \sqrt{\frac{\log(3\logfun(\mcrit_\totobs)/\parprob)}{\totobs}} + \frac{160\log(3\logfun(\mcrit_\totobs)/\parprob)}{\totobs}
\end{align*}
with probability exceeding $1 - \parprob$. The claim then follows from dividing both sides by $\norm{\fhat - \fstar}{\QQ_\Covariate}$ and using the fact that $\norm{\fhat - \fstar}{\QQ_\Covariate} \geq \mcrit_\totobs$. We additionally add the expression $\ocrit_\totobs(\QQ_\Covariate) + \ocrit_\totobs(\SynDist) + \norm{\nuhat - \nustar}{\QQ_\Covariate}$ to account for the event that $\norm{\fhat - \fstar}{\QQ_\Covariate}$ is less than that max. We can then bound to get the $\chi^2$-divergence
\begin{align*}
\norm{\nuhat - \nustar}{\QQ_\Covariate}^2 &= \int \left(\tfrac{d\SynDist}{d\QQ_\Covariate} - \tfrac{d\CondDist{1}}{d\QQ_\Covariate}\right)^2 \, d\QQ_\Covariate\leq 2 \int \tfrac{\big(\frac{d\SynDist}{d\QQ_\Covariate} - \frac{d\CondDist{1}}{d\QQ_\Covariate}\big)^2}{\frac{d\CondDist{1}}{d\QQ_\Covariate}} \, d\QQ_\Covariate,
\end{align*}
where we use the inequality $\frac{d\CondDist{1}}{d\QQ_\Covariate} \leq 2$ a.s.

\subsection{Proof of auxiliary lemmas for~\Cref{thm:fast-rates}}

\subsubsection{Proof of~\Cref{lem:new-localization}}
\label{sec:proof-new-localization}

Consider the empirical process suprema
\begin{align*}
Z_{\totobs}(t) &\coloneqq \sup_{\substack{f \in \fclass \\ \norm{f - \fstar}{\QQ_\Covariate} \leq t}} \left| \frac{1}{\totobs} \left(\sum_{i =1}^{\numobs_0} \big(\ExcessLoss{f}(0, \Covariate_i) - \EE_{\CondDist{0}}\left[\ExcessLoss{f}(0, \Covariate)\right]\big)  \right. \right. \\
&\qquad \qquad \qquad \qquad \qquad  \left. \left. + \sum_{j=\numobs_0+1}^{\totobs} \big(\ExcessLoss{f}(1, \Covariate_j') - \EE_{\CondDist{1}}\left[\ExcessLoss{f}(1, \Covariate)\right]\big) \right) \right| 
\end{align*}
where $\{\Covariate_i\}_{i=1}^{\numobs_0} \stackrel{i.i.d.}{\sim} \CondDist{0}$ and $\{\Covariate_j'\}_{j=\numobs_0 + 1}^{\totobs} \stackrel{i.i.d.}{\sim} \CondDist{1}$. Note that, after reindexing, it suffices to control $Z_\totobs(\norm{\fhat - \fstar}{\QQ_\Covariate})$. Applying~\Cref{lem:emp-conc} to the above with $\tau = 1$ to
\begin{align*}
\gclass = \left\{g : g_i(\covariate) = \begin{cases} \ExcessLoss{f}(0, \covariate) - \EE_{\CondDist{0}}\left[\ExcessLoss{f}(0, \Covariate)\right] & \text{if} \quad i \leq \numobs_0, \\ \ExcessLoss{f}(1, \covariate) - \EE_{\CondDist{1}}\left[\ExcessLoss{f}(1, \Covariate)\right]  & \text{if} \quad \numobs_0 + 1 \leq i \leq \totobs, \end{cases} \quad \text{$\norm{f - \fstar}{\QQ_\Covariate} \leq t$}
 \right\},
\end{align*}
we get
\begin{align*}
Z_{\totobs}(t) \leq 2 \, \EE\left[Z_{\totobs}(t)\right] + \sqrt{\sigma^2(\gclass)} \sqrt{\frac{2s}{\totobs}} + \frac{16Ls}{\totobs}
\end{align*}
with probability exceeding $1 - e^{-s}$. We claim that, for $t \geq \ocrit_{\totobs}(\QQ)$, we have
\begin{align}
\label{eqn:emp-calculations}
\EE\left[Z_{\totobs}(t)\right] \leq 16Lt \cdot \ocrit_\totobs(\QQ_\Covariate) + \frac{128L}{\totobs \cdot 2^{\totobs}}, \qquad \text{and} \qquad \sigma^2(\gclass) \leq 2L^2t^2.
\end{align}
Taking this as given for now, we have
\begin{align*}
Z_\totobs(t) \geq Q(t, s) \coloneqq 32Lt \ocrit_\totobs(\QQ_\Covariate) + 2Lt \sqrt{\frac{s}{\totobs}} + \frac{32Ls}{\totobs} \qquad \text{with probability at most $e^{-s}$.}
\end{align*}
Recall the shorthand $\logfun(\ocrit_\totobs(\QQ_\Covariate)) = \log_2\big(\frac{4}{\ocrit_\totobs(\QQ_\Covariate)}\big)$. Applying~\Cref{lem:peeling} to the above with $U = \max\{\ocrit_\totobs(\QQ_\Covariate), \norm{\fhat - \fstar}{\QQ_\Covariate}\}$ yields
\begin{align*}
Z_\totobs\left(\norm{\fhat - \fstar}{\QQ_\Covariate}\right) &\leq 4L \cdot \max\left\{ \ocrit_\totobs(\QQ_\Covariate), \norm{\fhat - \fstar}{\QQ_\Covariate} \right\} \left( 16\ocrit_\totobs(\QQ_\Covariate) + \sqrt{\frac{\log(\logfun(\ocrit_\totobs(\QQ_\Covariate))/\parprob)}{\totobs}} \right) \\
&\qquad \qquad \qquad + \frac{64L\log(\logfun(\ocrit_\totobs(\QQ_\Covariate))/\parprob)}{\totobs},
\end{align*}
with probability exceeding $1 - \parprob$. The claim then follows from the fact that $\rho$ is a decreasing function and $\mcrit_{\totobs} \leq \ocrit_\totobs(\QQ_\Covariate)$.

\noindent \textbf{Proof of~\Cref{eqn:emp-calculations}:} A key relation used in these bounds is that for all functions $g$ we have
\begin{align*}
\norm{g}{\QQ_\Covariate} = \sqrt{\frac{1}{2} \|g\|_{\CondDist{0}}^2 + \frac{1}{2} \| g\|_{\CondDist{1}}^2} \geq \frac{1}{\sqrt{2}} \cdot \max\left\{ \norm{g}{\CondDist{0}}, \norm{g}{\CondDist{1}} \right\}.
\end{align*}
Thus we can then compute
\begin{align*}
\sigma^2(\gclass) &=  \sup_{\norm{f - \fstar}{\QQ_\Covariate}\leq t} \left( \frac{\numobs_0}{\totobs} \cdot \text{Var}_{\CondDist{0}}\big(\ExcessLoss{f}(0, \Covariate)\big) + \frac{\numobs_1 + J}{\totobs} \cdot \text{Var}_{\CondDist{1}}\big( \ExcessLoss{f}(1, \Covariate') \big) \right) \\
&\leq L^2 \sup_{\norm{f - \fstar}{\QQ_\Covariate}\leq t} \left( \frac{\numobs_0}{\totobs} \cdot \norm{f - \fstar}{\CondDist{0}}^2 + \frac{\numobs_1 + J}{\totobs} \cdot \norm{f-\fstar}{\CondDist{1}}^2\right) \leq 2L^2t^2,
\end{align*}
where we have used the Lipschitzness of the loss function and
\begin{align*}
\text{Var}_{\CondDist{0}}\big(\ExcessLoss{f}(0, \Covariate)\big) \leq \EE_{\CondDist{0}}\left[\left(\loss(0, f(\Covariate)) - \loss(0, \fstar(\Covariate)) \right)^2  \right] \leq L^2 \norm{f - \fstar}{\CondDist{0}}^2,
\end{align*}
and a similar calculation holds for $\text{Var}_{\CondDist{1}}\big(\ExcessLoss{f}(1, \Covariate)\big)$.

To control the other term, we have 
\begin{align*}
\EE[Z_{\totobs}(t)] &= \EE\left[\sup_{\substack{f \in \fclass \\ \norm{f - \fstar}{\QQ_\Covariate} \leq t}} \left| \frac{1}{\numobs+J} \left(\sum_{i =1}^{\numobs_0} \big(\ExcessLoss{f}(0, \Covariate_i) - \EE_{\CondDist{0}}\left[\ExcessLoss{f}(0, \Covariate)\right]\big)  \right. \right. \right. \\
&\qquad \qquad \qquad \qquad \qquad \left. \left. \left. + \sum_{j=\numobs_0+1}^{\numobs + J} \big(\ExcessLoss{f}(1, \Covariate_j') - \EE_{\CondDist{1}}\left[\ExcessLoss{f}(1, \Covariate)\right]\big) \right) \right| \right]\\
&\stackrel{(a)}{\leq} 2 \, \EE\left[ \sup_{\norm{f - \fstar}{\QQ_\Covariate} \leq t} \left| \frac{1}{\totobs} \left( \sum_{i=1}^{\numobs_0} \rad_i \cdot \ExcessLoss{f}(0, \Covariate_i) + \sum_{j=\numobs_0+1}^{\totobs} \rad_j \cdot \ExcessLoss{f}(1, \Covariate_j') \right) \right| \right] \\
&\stackrel{(\bound)}{\leq} 4L \, \EE\left[\sup_{\norm{f - \fstar}{\QQ_\Covariate} \leq t} \left| \frac{1}{\totobs} \left( \sum_{i=1}^{\numobs_0} \rad_i \cdot \big(f(\Covariate_i) - \fstar(\Covariate_i) \big) + \sum_{j=\numobs_0 + 1}^\totobs \rad_j \cdot \big( f(\Covariate_j') - \fstar(\Covariate_j') \big) \right) \right|\right]
\end{align*}
where step (a) follows from symmetrization and step (\bound) follows from applying the Ledoux-Talagrand inequality to
\begin{align*}
\phi_i(s) = \begin{cases} \loss(0, s + \fstar(\Covariate_i)) - \loss(0, \fstar(\Covariate_i)) & \text{if $i = 1, \ldots, \numobs_0$,} \\
\loss(1, s + \fstar(\Covariate_i')) - \loss(1, \fstar(\Covariate_i')) & \text{if $i = \numobs_0 + 1, \ldots, \totobs$.} \end{cases}
\end{align*}
The following lemma then relates the above quantity to $\RadComp_\totobs(t; \QQ_\Covariate)$.

\begin{lemma}
\label{lem:jason-blocking}
We have 
\begin{align*}
&\EE\left[\sup_{\norm{f - \fstar}{\QQ_\Covariate} \leq t} \left| \frac{1}{\totobs} \left( \sum_{i=1}^{\numobs_0} \rad_i \cdot \big(f(\Covariate_i) - \fstar(\Covariate_i) \big) + \sum_{j=\numobs_0 + 1}^\totobs \rad_{j} \cdot \big( f(\Covariate_j') - \fstar(\Covariate_j') \big) \right) \right|\right] \\
&\qquad \qquad \qquad \leq  4\RadComp_\totobs(t; \QQ_\Covariate) + \frac{32}{\totobs \cdot 2^{\totobs}}.
\end{align*}
\end{lemma}
\noindent See~\Cref{sec:proof-jason-blocking} for a proof of this result. Therefore we have
\begin{align*}
\EE[Z_\totobs(t)] \leq 16L \cdot \RadComp_\totobs(t; \QQ_\Covariate) + \frac{128L}{\totobs \cdot 2^{\totobs}} \leq 16Lt \cdot \ocrit_\totobs(\QQ_\Covariate) + \frac{128L}{\totobs \cdot 2^{\totobs}}.
\end{align*}
The last step follows since $t \mapsto \frac{\RadComp_\totobs(t; \QQ_\Covariate)}{t}$ is non-increasing (see Chapter 13 in~\citet{dinosaur2019}), for $t \geq \ocrit_\totobs(\QQ_\Covariate)$ we have 
\begin{align*}
\frac{\RadComp_\totobs(t; \QQ_\Covariate)}{t} \leq \frac{\RadComp_\totobs(\ocrit_\totobs(\QQ_\Covariate); \QQ_\Covariate)}{\ocrit_\totobs(\QQ_\Covariate)} \leq \ocrit_\totobs(\QQ_\Covariate).
\end{align*}

\subsubsection{Proof of~\Cref{lem:jason-blocking}}
\label{sec:proof-jason-blocking}

For convenience, define the function class
\begin{align*}
\fclass^*(t) \coloneqq \{f - \fstar : f \in \fclass,\; \norm{f - \fstar}{\QQ_\Covariate} \leq t\}.
\end{align*}
Let $\{U_i\} \stackrel{i.i.d.}{\sim} \CondDist{0}$, $\{V_j\} \stackrel{i.i.d.}{\sim} \CondDist{1}$, and $\{\rad_i\}\stackrel{i.i.d.}{\sim} \text{Rad}$, all generated independently of each other. For integers $n,m \geq 0$ define
\begin{align}
\label{eqn:Z-defn}
Z(n, m; t) \coloneqq \EE\left[ \sup_{g \in \fclass^*(t)} \left| \left( \sum_{i=1}^{n} \rad_i \cdot g(U_i) + \sum_{j=1}^{m} \rad_{n + j} \cdot g(V_j) \right) \right| \right].
\end{align}
We take as given that $Z(\cdot, \cdot; t)$ is monotonic in both entries, i.e., for $n_0' \geq n_0$ and $n_1' \geq n_1$
\begin{align}
\label{eqn:monotonic}
Z(n_0', n_1'; t) \geq Z(n_0, n_1; t).
\end{align}
See~\Cref{sec:additional-calcs-2} for its proof.

Let $B \sim \text{Bin}(\totobs, \frac{1}{2})$ drawn independently of the data; it suffices by~\Cref{eqn:combinatorial} to show
\begin{align*}
Z(\numobs_0, \numobs_1 + J; t) \leq  4\EE_{B}\left[Z(\totobs - B, B; t)\right] + \frac{32}{2^{\totobs}}.
\end{align*}
We do so by first establishing the following intermediate result: we have (defining $1/ 0 = +\infty$)
\begin{align}
\label{eqn:blocking-statement}
Z(\numobs_0, \numobs_1 + J; t) \leq \left\lceil \max\left\{ \frac{\numobs_0}{\totobs - K}, \frac{\numobs_1 + J}{K}  \right\}  \right\rceil \cdot Z(\totobs - K, K; t).
\end{align}
Letting $R(K) \coloneqq \left\lceil \max\left\{ \frac{\numobs_0}{\totobs - K}, \frac{\numobs_1 + J}{K}  \right\}  \right\rceil$ and $B' = \min\{\max\{B, 1\}, \totobs-1\}$ (i.e.,~$B'$ is $B$ but truncated so it does not take the value $0$ or $\totobs$), we have that $\frac{1}{R(B')} \cdot Z(\numobs_0, \numobs_1 + J; t) \leq Z(\totobs - B', B'; t)$.
Taking expectations and rearranging yields
\begin{align*}
Z(\numobs_0, \numobs_1 + J; t) \leq \frac{1}{\EE_B\left[1/R(B')\right]} \cdot \EE_B[Z(\totobs - B', B'; t)].
\end{align*}
A straightforward calculation will give 
\begin{align*}
\EE_B\left[Z(\totobs - B', B'; t)\right] \leq \EE_B\left[Z(\totobs - B, B; t)\right] + \frac{8}{2^{\totobs}}.
\end{align*}
The claim then immediately follows from the next lemma. See~\Cref{sec:proof-R-bound} for its proof.
\begin{lemma}
\label{lem:R-bound}
We have, under the condition that $\numobs_0 = \numobs_1 + J$,
$
\EE_B\left[ \frac{1}{R(B')} \right] \geq \frac{1}{4}.
$
\end{lemma}

To establish~\Cref{eqn:blocking-statement}, we proceed via partitioning. Note by definition of $R(K$) we have $R(K) \cdot (\totobs - K) \geq \numobs_0$ and $R(K) \cdot K \geq \numobs_1 + J$ so by monotonicity
\begin{align*}
Z(\numobs_0, \numobs_1 + J; t) \leq Z(R(K) \cdot (\totobs - K), R(K) \cdot K; t).
\end{align*}
The main idea is to take the samples in $Z(R(K) \cdot (\totobs - K), R(K) \cdot K; t)$,  split the samples of $U$'s into $R(K)$ groups of size $\totobs - K$, and split the samples of $V$'s into $R(K)$ groups of size $K$, and rearrange it to obtain $R(K)$ copies of $Z(\totobs - K, K; t)$. Split the $U$'s up to $R(K)$ groups of size $\totobs-K$ by defining 
\begin{align*}
I_r \coloneqq \big\{(r-1)(\totobs - K) + 1, \ldots, r(\totobs - K) \big\} \qquad \text{for $r = 1, \ldots, R(K)$,}
\end{align*}
and split the $V$'s up to $R(K)$ groups of size at $K$ by defining
\begin{align*}
J_r \coloneqq \big\{ (r-1)K + 1, \ldots, rK  \big\} \qquad \text{for $r =1, \ldots, R(K)$}
\end{align*}
and define
$
T_r(g) \coloneqq \sum_{i \in I_r} \rad_i \cdot g(U_i) + \sum_{j \in J_r} \rad_{R(K) \cdot (\totobs - K) + j} \cdot g(V_j).
$
Thus we have
\begin{align*}
&Z(R(K) \cdot (\totobs - K), R(K) \cdot K; t)  = \EE\left[ \sup_{g \in \fclass^*(t)} \left| \sum_{r=1}^{R(K)} T_r(g) \right| \right] \leq R(K) \cdot Z(\totobs - K, K; t).
\end{align*}

\AtNextBibliography{\small}
\printbibliography

\appendix

\section{Further results}
\label{sec:more-results}

\subsection{Choosing $J$}
\label{sec:random-J}

As mentioned previously, the guarantees in~\Cref{thm:slow-rates} as stated require $J$ to be fixed or independent of the data $\dataset$. Doing so would require knowing the class probabilities $(\cprob_0, \cprob_1)$ \emph{a priori}, which is often not the case. In practice we would use $\dataset$ to estimate the class probabilities to select the number of samples $J$ to generate. Recall the notation $\totobs = \numobs + J$. The following corollary is a simple extension that allows for $J$ to depend on $\dataset$. We denote $\Gamma(\numobs, J)$ to be the upper bound in~\Cref{thm:slow-rates}; that is, 
\begin{align*}
\Gamma(\numobs, J) &\coloneqq 12\RadComp_{\totobs}(\QQ) + t_{\totobs}(\parprob) +  4\bound \cdot  \dTV{\SynDist}{\CondDist{1}} + 28\bound \cdot \left| \frac{\numobs\cprob_0}{\numobs + J} - \frac{1}{2} \right|.
\end{align*}
\begin{corollary}
\label{cors:random-J-slow}
Under the conditions of~\Cref{thm:slow-rates} and $J$ chosen dependent on $\dataset$ and some arbitrary fixed $\Jstar \in \mathbb{N}$, we have
\begin{align*}
&\EE_{\QQ}\left[\loss(\Response, \fhat(\Covariate)) - \loss(\Response, \fstar(\Covariate))\right] \leq \frac{\numobs + \Jstar}{\numobs + J} \cdot \Gamma(\numobs, \Jstar) + 4\bound \cdot \frac{|J - \Jstar|}{\numobs + J}.
\end{align*}
with probability exceeding $1 - \parprob$.
\end{corollary} 
\noindent See~\Cref{sec:proof-random-J-slow} for a proof of this result. 

Concretely, consider a procedure where we estimate $\widehat{\cprob}_0 = \empsum{i} \mathbf{1}\{\Response_i = 0\}$, and then choose $J = (2\widehat{\cprob}_0 - 1) \numobs $, $\Jstar = \lceil (2\cprob_0 - 1) \numobs \rceil$. Then we have that $|J - \Jstar| = O_{\PP}(\sqrt{\numobs})$ and 
\begin{align*}
\EE_{\QQ}\left[\loss(\Response, \fhat(\Covariate)) - \loss(\Response, \fstar(\Covariate))\right]  \lesssim \RadComp_{\numobs + \Jstar}(\QQ) + \dTV{\SynDist}{\CondDist{1}} + \frac{1}{\sqrt{\numobs}},
\end{align*}
with high probability, achieving the same guarantee as~\Cref{thm:slow-rates} up to constant factors.

\subsection{Arbitrary target mixtures}
\label{sec:target-mixtures}

In the exposition, we argued that the real target should be obtaining a classifier that performs well according to the distribution $\QQ_{\Covariate, \Response}$
\begin{align*}
\Response \sim \text{Ber}(\tfrac{1}{2}), \qquad \Covariate \mid \Response \sim (1 - \Response) \cdot \CondDist{0} + \Response \cdot \CondDist{1},
\end{align*}
as this is the target distribution where the classifier will focus on minimizing type-I and type-II errors equally. However in some situations we may care more about a different weighting of the two types of error. This problem can be formulated as building a classifier that performs well according to the target distribution $\QQ_{\Covariate, \Response}^\star$
\begin{align*}
\Response \sim \text{Ber}(\cprob_1^\star), \qquad \Covariate \mid \Response \sim (1 - \Response) \cdot \CondDist{0} + \Response \cdot \CondDist{1}
\end{align*}
for some user-specified target $\cprob_1^\star = 1-\cprob_0^\star$. For example, if we cared much more about minimizing the type-II error we would set $\cprob_1^\star > \frac{1}{2}$. In this case, the target is
\begin{align*}
\fstar(\Covariate) = \QQ^\star(\Response = 1 \mid \Covariate) = \frac{\cprob_1^\star d\CondDist{1}}{\cprob_0^\star d \CondDist{0} + \cprob_1^\star d \CondDist{1}}.
\end{align*}

\begin{corollary}
\label{cors:target-mixtures}
Under the conditions of~\Cref{thm:slow-rates}, consider Algorithm~\ref{AlgRebal} with an arbitrary $\SynDist$. For a uniformly bounded loss, it returns an estimate $\fhat$ with excess risk at most
\begin{align*}
\EE_{\QQ^\star}\left[\loss(\Response, \fhat(\Covariate)) - \loss(\Response, \fstar(\Covariate))\right] & \leq 12\RadComp_{\numobs+J}(\QQ^\star)  +  4\bound \cdot  \dTV{\SynDist}{\CondDist{1}}  \\
&\qquad \qquad  + 28\bound \cdot \left| \tfrac{\numobs \cprob_0}{\numobs + J} - \cprob_0^\star \right| + t_{\numobs + J}(\parprob),
\end{align*}
with probability exceeding $1 - \parprob$.
\end{corollary}

\noindent See~\Cref{sec:proof-target-mixtures} for a proof of this result. Note that these results take the same structure as those of Theorems~\ref{thm:slow-rates}; the only difference is the mixture bias term which measures the deviation from the idealized sample distribution to the true target distribution $\QQ^\star$.

\subsection{Undersampling methods}
\label{sec:undersampling-methods}

Here we provide a guarantee for undersampling methods. The idea behind undersampling is to construct the training set by subsampling $H_0$ from the majority class to ensure balance. This provides an alternative to the rebalancing approaches that is less sample efficient (as it discards samples from $\dataset_0$) but is more computationally efficient, making it an attractive choice for very large datasets~\cite{he2014practical}.

\begin{algorithm}[t]
\caption{Undersampling method}
\begin{algorithmic}[1] 
\State \texttt{Inputs:} (i) Dataset $\dataset = \{(\Covariate_i, \Response_i)\}_{i=1}^\numobs$ and number of majority draws $ K\leq \numobs_0$. (ii) Loss function $\loss$ and function class $\fclass$ for estimating $\fstar$. 
\vspace{6pt}

\State Randomly subsample a set $H_0$ of size $K$ from $I_0$. 

\State Compute the estimate
\begin{align*}
\fhat \in \argmin_{f \in \fclass} \left( \frac{1}{K + \numobs_1} \left\{ \sum_{i \in H_0} \loss(0, f(\Covariate_i)) + \sum_{j\in I_1} \loss(1, f(\Covariate_j))  \right\} \right).
\end{align*}

\end{algorithmic}
\label{AlgUnder}
\end{algorithm}

\begin{proposition}
\label{prop:undersampling}
Given observations $\dataset = \{(\Covariate_i, \Response_i)\}_{i=1}^\numobs$, consider Algorithm~\ref{AlgUnder} implemented with a $b$-uniformly bounded loss. Then for any $\parprob \in (0, 1)$, the output $\fhat$ has, with probability exceeding $1 - \parprob$, excess risk at most
\begin{align*}
\EE_\QQ\left[ \loss(\Response, \fhat(\Covariate)) - \loss(\Response, \fstar(\Covariate)) \right] &\leq 4\RadComp_{K + \numobs_1}(\QQ) +12\bound \cdot \left| \frac{K}{K+\numobs_1} - \frac{1}{2} \right| \\
&\qquad + \left( 4\sqrt{2\log(1/\parprob)} + 12 \right) \cdot \frac{\bound}{\sqrt{K + \numobs_1}}.
\end{align*}
\end{proposition}
\noindent
See~\Cref{sec:proof-undersampling} for a proof of this result. We remark that we treat $K$ as a fixed quantity here, to allow for random $K$ we can apply the argument in~\Cref{cors:random-J-slow}. As before, the bound depends on the Rademacher complexity $\RadComp_{K + \numobs_1}(\QQ)$, which can be thought of as the oracle rate given $K+\numobs_1$ observations from $\QQ$. We remark that the proof does not follow from directly analyzing empirical risk minimization over $\dataset' = \{(\Covariate_i, 0) : i \in H_0\} \cup \{(\Covariate_j, 1) : j \in I_1\}$, as the data in $\dataset'$ is not drawn i.i.d.~from $\QQ$. 

We note that, although our result for undersampling in~\Cref{prop:undersampling} does not contain an explicit cost of transfer term such as the total-variation discrepancy $\dTV{\SynDist}{\CondDist{1}}$ appearing in the continuous-synthetic setting of~\Cref{thm:slow-rates}, or the minority-class Rademacher term $\RadComp_{\numobs_1}(\CondDist{1})$ appearing in the bootstrapping bound~\eqref{eqn:bootstrapping}, the leading-order oracle term $\RadComp_{K + \numobs_1}(\QQ)$ in ~\Cref{prop:undersampling} is larger than the corresponding $\RadComp_{\numobs + J}(\QQ)$ in our oversampling results, reflecting the fact that undersampling discards the majority-class data.

\section{Auxiliary lemmas}

Here we collect several lemmas that will be used throughout the proofs.

\subsection{McDiarmid's inequality}

We say $f: \stateset^n \to \RR$ satisfies the bounded differences property if there exist constants $c_1, c_2, \ldots, c_n$ such that for all $i \in [n]$ and $x_i \in \stateset$, we have
\begin{align}
\label{eqn:bound-diff}
\sup_{x_i' \in \stateset} \left| f(x_1, \ldots, x_{i-1}, x_i, x_{i+1}, \ldots, x_n) - f(x_1, \ldots, x_{i-1}, x_i', x_{i+1}, \ldots, x_n)  \right| \leq c_i.
\end{align}

\begin{lemma}[McDiarmid's inequality]
\label{lem:bounded-diff}
Let $\Covariate_i \in \stateset$ be independent random variables. Then for any $f$ satisfying~\Cref{eqn:bound-diff} and $\parprob \in (0, 1)$ we have
\begin{align*}
f(\Covariate_1, \ldots, \Covariate_n) \leq \EE[f(\Covariate_1, \ldots, \Covariate_n)] + \sqrt{2\log\left(\frac{1}{\parprob}\right) \cdot \left( \sum_{i=1}^n c_i^2 \right)},
\end{align*}
with probability exceeding $1 - \parprob$. 
\end{lemma}

\subsection{Suprema of empirical processes}

Here we let $\gclass$ consist of $n$-tuples of functions. Using the notation $\|\PP_n \|_{\gclass} \coloneqq \sup_{g \in \gclass} \left| \frac{1}{n}\sum_{i=1}^n g_i(\Covariate_i) \right|$ and $\sigma^2(\gclass) \coloneqq \sup_{g \in \gclass} \{\frac{1}{n}\sum_{i=1}^n \Var(g_i(\Covariate_i)) \}$, the following result gives an upper tail bound on $\| \PP_n \|_{\gclass}$. 

\begin{lemma}[\citep{klein2005empproc}]
\label{lem:emp-conc}
Consider a countable, $\bound$-uniformly bounded function class
$\gclass$ such that $\EE[g_i(\Covariate_i)] = 0$ for all $g \in \gclass$.  Then for
any $\tau > 0$ and $\parprob \in (0, 1)$ we have
\begin{align*}
\| \PP_n \|_\gclass \leq (1 + \tau) \EE[ \| \PP_n
\|_\gclass] + \sqrt{\sigma^2(\gclass) }
\sqrt{\tfrac{2\log(1/\parprob)}{n}} + \big(3 + \tfrac{1}{\tau}
\big) \tfrac{\bound \log(1/\parprob)}{n}
\end{align*}
with probability at least $1 - \parprob$.
\end{lemma}

\subsection{Abstract peeling}

The following lemma is taken from~\citet{xia2024prediction}. Consider a function class $\gclass$ equipped with a norm $\| \cdot \|$, and some empirical process $\{V_n(g): g \in \gclass\}$. Define the localized empirical process by $Z_{n}(r) \coloneqq \sup_{g \in \gclass: \|g\|\leq r} V_n(g)$, and suppose there exists scalar $s$ and function $Q: \RR^2 \to \RR$ such that we have
\begin{align}
\label{eqn:base-conc}
\PP\left(Z_n(r) \geq Q(r, t)\right) \leq e^{-t} \qquad \text{for all fixed $r \geq s$.}
\end{align}
The following lemma establishes that, given a bounded random variable $U$ (potentially correlated with $V_n$), we have $Z_n(U) \lesssim Q(U, t)$ with high probability. 

\begin{lemma}
\label{lem:peeling}
Suppose that~\Cref{eqn:base-conc} holds, the function $Q$ is increasing in its first argument, and $Q(2r, t) \leq 2Q(r ,t)$ for all $r \geq s$ and $t > 0$. Then for any random variable $U \in [s, b]$,
\begin{align*}
\PP\left(Z_n(U) \geq 2Q(U, t) \right) \leq \left \lceil \log_2\left(\tfrac{2b}{s}\right) \right \rceil \cdot e^{-t}.
\end{align*}
\end{lemma}

\subsection{Nearest neighbor control}

Here we present some intermediate results on nearest neighbors used in our proofs.

\subsubsection{Uniform distance control}

We require the following lemmas about controlling the maximal distance for $k$-nearest neighbors, taken from~\citet{pmlr-v84-xue18a}.

\begin{lemma}
\label{lem:nearest-neighbors}
Let $r_k(\covariate)$ be defined as the distance from $x \in \text{supp}(\Covariate) \subset \RR^d$ to its $k^{th}$ nearest neighbor in a sample $\{\Covariate_i\}_{i=1}^\numobs \stackrel{i.i.d.}{\sim} \PP$, where $\PP$ satisfies~\Cref{eqn:SMOTE-lower-bound}. Then we have
\begin{align*}
\sup_{x \in \text{supp}(\Covariate)} r_k(\covariate) \leq \left(\frac{3}{c_d}\right)^{1/d} \cdot \left(\frac{\max\{k, (d+1) \log(2\numobs) + \log(8/\parprob)\}}{\numobs}\right)^{1/d}
\end{align*}
with probability exceeding $1 - \parprob$.
\end{lemma}
\noindent Note that we will use the following simplification for presentation
\begin{align*}
\sup_{x \in \text{supp}(\Covariate)} r_k(\covariate) \leq \left(\frac{3k + 4d \cdot \log(16\numobs / \parprob)}{c_d\numobs}\right)^{1/d}
\end{align*}
with probability exceeding $1 - \parprob$.

The following lemma follows from a simple modification of the argument from~\citet{biau2015lectures}.
\begin{lemma}
\label{lem:nearest-distance}
Suppose $\text{supp}(\Covariate) \subset \BB_d(0, D)$ and let $\SCovariate_{(k)}(\Covariate)$ denote the $k^{th}$ nearest neighbor to $\Covariate$ in a given sample of size $\numobs$. Then we have
\begin{align*}
\EE\big[\|\SCovariate_{(k)}(\Covariate_1) - \Covariate_1 \|_2\big] \leq 6D \left( \frac{k}{\numobs} \right)^{1/d}.
\end{align*}
\end{lemma}

\subsubsection{Maximum degree}
\label{sec:maximum-degree}

This section is dedicated to answering the following question:
\begin{center}
\emph{Given a sample $\dataset$ of $\numobs$ points in $\RR^d$, what is the maximum number of times any given point $x \in \dataset$ can be one of the $k$-nearest neighbors for the other points in $\dataset$?}
\end{center}
For the case of $k =1$, the answer is given by the \emph{kissing number} $\tau_d$~\cite{eppstein1997nearest}. The work of~\citet{kabatiansky1978bounds} established the following upper bound 
\begin{align*}
\tau_d \leq (2^{0.41})^d.
\end{align*}
Here we prove that an upper bound for general $k$ that is weaker but sufficient for our purposes. Note there are connections between these results and spherical codes in information theory.

We begin by introducing some definitions. For $x \in \RR^d$, define the cone of angle $\alpha$ around some $x \neq 0$ as
\begin{align*}
C(x, \alpha) \coloneqq \left\{y \in \RR^d: y \neq 0 \quad \text{and} \quad \tfrac{\langle x, y \rangle}{\|x\|_2 \|y\|_2} \geq \cos(\alpha) 
\right\}.
\end{align*}
Note that $C(x, \alpha)$ consists of all points $y$ in $\RR^d$ such that the angle between $x$ and $y$ is less than $\alpha$.  An $\alpha$-conic covering of $\RR^d$ is a set of points $\{x_1, \ldots, x_l\}$ such that $\bigcup_{i=1}^l C(x_i, \alpha) = \RR^d$, and we use $N_d(\alpha)$ to denote the minimal number of points required to form such a covering. Similarly, we define $\{x_1, \ldots, x_m\}$ to be an $\alpha$-conic packing of $\RR^d$ if the angle between all the points is at least $\alpha$, i.e., $\frac{\langle x_i, x_j \rangle}{\|x_i\|_2 \|x_j\|_2} \leq \cos(\alpha)$, and we use $M_d(\alpha)$ to denote the size of the maximal such packing. It is straightforward to show that for $\alpha \in [0, \frac{\pi}{2}]$ we have
\begin{align}
\label{eqn:pack-cover-relation}
 M_d(2\alpha) \leq N_d(\alpha) \leq M_d(\alpha).
\end{align}
Indeed, a maximal packing must cover $\RR^d$; otherwise, we could add another point to it, a contradiction, establishing the upper bound. The lower bound follows from establishing the claim that for unit vectors $x, y, z \in \RR^d$ we have (assuming $\alpha \leq \frac{\pi}{2}$)
\begin{align}
\label{eqn:angle-sum}
\langle x, y \rangle \geq \cos(\alpha) \quad \text{and} \quad \langle y, z \rangle \geq \cos(\alpha) \implies \langle x, z \rangle \geq \cos(2\alpha).
\end{align}
Given this consider a $2\alpha$-packing $\{x_1, \ldots, x_K\}$ and an $\alpha$-covering $\{y_1, \ldots, y_L\}$.  If $K > L$ then by the pigeonhole principle we must have that two points $x_k$ and $x_j$ must belong to the same cone $C(y_l, \alpha)$, contradicting~\Cref{eqn:angle-sum}. 

The kissing number is defined as $\tau_d = M_d(\frac{\pi}{3})$. Given $\dataset \subset \RR^d$, we will show that the maximum number of points a given $x_1 \in \dataset$ can be a nearest neighbor for is upper bounded by $M_d(\frac{\pi}{3})$. To see why, suppose we had at least $M_d(\frac{\pi}{3}) + 1$ points for which $x_1$ is the nearest neighbor of. Then there must exist $x_j, x_l$ such that, if we define $v_1 = x_j - x_1$ and $v_2 = x_l - x_1$, the angle between $v_1$ and $v_2$ is (strictly) less than $\frac{\pi}{3}$. This follows by considering the maximal $\frac{\pi}{3}$ packing translated to be centered at $x_1$. Without loss of generality, let $\|v_2\|_2 \geq \|v_1\|_2$; we can compute (since $\cos(\angle(v_1, v_2)) > \frac{1}{2}$)
\begin{align*}
\| x_l - x_j\|_2^2 &= \|v_2 - v_1\|_2^2 \\
&= \|v_1 \|_2^2 + \|v_2\|_2^2 - 2 \cos(\angle(v_1, v_2)) \cdot \|v_1\|_2 \|v_2\|_2 \\
&< \|v_1\|_2^2 + \|v_2\|_2^2 - \|v_1\|_2 \cdot \|v_2\|_2 \\
&\leq \|v_2\|_2^2 = \|x_l - x_1\|^2_2.
\end{align*}
Thus $x_j$ is closer to $x_l$ than $x_1$, a contradiction.

We claim that an upper bound for general $k$ is given by $k \cdot N_d(\frac{\pi}{6})$. Consider the minimal covering $\{y_1, \ldots, y_N\}$ for $N = N_d(\frac{\pi}{6})$. If $x_1 \in \dataset$ had $kN + 1$ nearest neighbors, then by the pigeonhole principle at least $k+1$ of those points must fall into the cone $C(y_i - x_1, \frac{\pi}{6})$, which we denote by $\{z_1, \ldots, z_{k+1}\}$ and we define $v_j = z_j - x_1$ for $j = 1, \ldots, k+1$. By~\Cref{eqn:angle-sum} we have all the $v_j$ have angle at most $\frac{\pi}{3}$ between them; without loss of generality, let $v_{k+1}$ be the point with the largest norm amongst them. By the calculation above we have that the points $\{z_1, \ldots, z_k\}$ are closer to $z_{k+1}$ than $x_1$, a contradiction. We can compute an explicit upper bound for $N_d(\frac{\pi}{6})$ in the following manner: two unit vectors $u$ and $v$ with angular separation $\theta$ will have chord-distance given by
\begin{align*}
\|u - v \|_2 = 2\sin(\tfrac{\theta}{2}).
\end{align*}
Thus if we have an $\epsilon$-net of the hypersphere with $\epsilon = 2\sin(\frac{\theta}{2})$, then those points will also form an $\theta$-conic covering of the hypersphere. Therefore we conclude
\begin{align*}
N_d\left(\frac{\pi}{6}\right) \leq \left(1 + \frac{1}{\sin(\frac{\pi}{12})} \right)^d \leq 5^d.
\end{align*}
Note that this is quite a crude (but simple) bound and there is a small literature dedicated to proving sharp rates~\cite{cohn2003new, Cohn2012SPHEREPB}.

\noindent\textbf{Proof of~\Cref{eqn:angle-sum}:} It suffices to assume that $x, y, z$ are unit vectors, and let $\theta_{xy}$ be the angle between $x$ and $y$, and $\theta_{yz}$ be the angle between $y$ and $z$. Decompose $x$ and $z$ into components parallel to and orthogonal to $y$
\begin{align*}
x = \langle x, y \rangle \cdot y + x_\perp, \qquad \text{and} \qquad z = \langle z, y \rangle \cdot y + z_\perp.
\end{align*}
Observe that $\|x_\perp\|_2 = \sin(\theta_{xy})$ and $\|z_\perp\|_2 = \sin(\theta_{yz})$.  We can then compute 
\begin{align*}
\langle x, z \rangle &= \langle x, y \rangle \cdot \langle z, y \rangle + \langle x_\perp, z_\perp \rangle \\
&\geq \cos(\theta_{xy}) \cos(\theta_{yz}) - \sin(\theta_{xy}) \sin(\theta_{yz}) \\
& = \cos(\theta_{xy} + \theta_{yz}).
\end{align*}
The claim then follows from the fact that $\cos(\cdot)$ is decreasing on $[0, \pi]$.

\section{Additional proofs}

\subsection{Proof of~\Cref{lem:coupling}}
\label{sec:proof-coupling}

Consider some dominating measure $\nu$ of both $\PP_1$ and $\PP_2$, and define the measure $\mu$ via $\mu(A) = \int_A \min\{\frac{d\PP_1}{d\nu}, \frac{d\PP_2}{d\nu}\} \, d\nu$. Denoting $\eta \coloneqq \mu(\Omega)$, we have that
\begin{align*}
\dTV{\PP_1}{\PP_2} = \frac{1}{2} \int |d\PP_1 - d\PP_2| = \int (d\PP_1 - d\PP_2)_{+} = \int (d\PP_1 - \min\{d\PP_1, d\PP_2\}) = 1 - \eta.
\end{align*}
Furthermore let $c(\covariate) \coloneqq \frac{d\mu(\covariate)}{d\PP_1}$ be the density of $\mu$ with respect to~$\PP_1$ and observe that its range is $[0, 1]$. Now draw independent random variables $V \sim \text{Unif}[0, 1]$ and $W \sim \frac{\PP_2 - \mu}{1 - \eta}$ that are also both independent of $U_1$. We then construct $U_2$ as follows: 
\begin{align*}
U_2 = \begin{cases} U_1 & \text{if} \quad V \leq c(U_1), \\ W & \text{otherwise.} \end{cases}
\end{align*}
Note that this construction establishes the claim that $\Pr(U_1 \neq U_2) = \dTV{\PP_1}{\PP_2}$. Observe 
\begin{align*}
\Pr(U_1 = U_2) = \Pr(V \leq c(U_1)) + \Pr(V > c(U_1) \text{  and  } W = U_1).
\end{align*}
We have
\begin{align*}
\Pr(V \leq c(U_1)) = \int c(\covariate) \, d\PP_1 = \eta.
\end{align*}
Furthermore, note that
\begin{align*}
\Pr(V < c(U_1) \text{  and  } W = U_1) &= \EE\left[\mathbf{1}\{V > c(U_1)\} \cdot \mathbf{1}\{W = U_1\}\right] \\
&= \EE\left[(1 - c(U_1)) \cdot \mathbf{1}\{W = U_1\} \right].
\end{align*}
By definition we have $c(U_1) < 1$ if and only if $U_1 \in \{ d\PP_1 > d\PP_2\}$ and $W$ is supported on $\{d\PP_2 > d\PP_1\}$, implying that the above expectation is $0$.

To show that $U_2 \sim \PP_2$, we have
\begin{align*}
\Pr(U_2 \in B) &= \Pr(U_1 \in B \text{ and } V \leq c(U_1)) + \Pr(W \in B \text{ and } V > c(U_1)) \\
&= \EE\left[ \mathbf{1}\{U_1 \in B\} \cdot \mathbf{1}\{V \leq c(U_1)\} \right] + \EE\left[ \mathbf{1}\{W \in B\} \cdot \mathbf{1}\{V > c(U_1)\} \right] \\
&\stackrel{(a)}{=} \EE\left[\mathbf{1}\{U_1 \in B\} \cdot c(U_1) \right] + \frac{\PP_2(B) - \mu(B)}{1 - \eta} \cdot \EE[1 - c(U_1)] \\
&\stackrel{(b)}{=} \mu(B) + \frac{\PP_2(B) - \mu(B)}{1 - \eta} \cdot (1 - \eta) = \PP_2(B)
\end{align*}
where step (a) follows from conditioning on $U_1$ and using that $(U_1, W, V)$ are mutually independent and step (b) follows from the fact that $c$ is the Radon-Nikodym derivative.

\subsection{Proof of~\Cref{lem:mixture-rad} and~\Cref{lem:annoying-stuff}}
\label{sec:proof-mixture-rad}

Key to our proofs here is the following result: suppose we observe samples from the mixture $\WW = p \PP_1 + q \PP_2$, and let $B \sim \text{Bin}(n, q)$ independent of everything, and $\gclass$ is a $\bound$-uniformly bounded function class. Then we have
\begin{align}
\label{eqn:combinatorial}
\EE_{\substack{W_i \stackrel{i.i.d.}{\sim} \WW \\ \rad_i \stackrel{i.i.d.}{\sim} \text{Rad}}} \left[ \sup_{g \in \gclass} \left| \frac{1}{n} \sum_{i=1}^n \rad_i \cdot g(W_i)  \right| \right] = \EE_{\substack{U_i \stackrel{i.i.d.}{\sim} \PP_1 \\ V_j \stackrel{i.i.d.}{\sim} \PP_2 \\ \rad_i \stackrel{i.i.d.}{\sim} \text{Rad}}}\left[ \sup_{g \in \gclass} \left| \frac{1}{n} \left( \sum_{i=1}^{n - B} \rad_i \cdot g(U_i) + \sum_{j=1}^{B} \rad_{n + 1 - j} \cdot g(V_j) \right) \right| \right].
\end{align}
We defer the proof to the end of this section.

\noindent\textbf{Proof of~\Cref{lem:annoying-stuff}:} We require the following step: suppose $\gclass$ is a uniformly $\bound$-bounded function class and we have two mixture distributions, $\WW = p \PP_1 + q \PP_2$ and $\WW' = p' \PP_1 + q' \PP_2$. Then we have
\begin{align*}
\EE_{\substack{W_i \stackrel{i.i.d.}{\sim} \WW \\ \rad_i \stackrel{i.i.d.}{\sim} \text{Rad}}} \left[ \sup_{g \in \gclass} \left| \frac{1}{n} \sum_{i=1}^n \rad_i \cdot g(W_i)  \right| \right] \leq\EE_{\substack{W_i \stackrel{i.i.d.}{\sim} \WW' \\ \rad_i \stackrel{i.i.d.}{\sim} \text{Rad}}} \left[ \sup_{g \in \gclass} \left| \frac{1}{n} \sum_{i=1}^n \rad_i \cdot g(W_i)  \right| \right] + \frac{2\bound}{\sqrt{n}} + 2\bound \cdot |q - q'|.
\end{align*}
\Cref{lem:annoying-stuff} then follows from applying the above to
\begin{align*}
&U = V = (\Covariate, \Response), \qquad \gclass = \{ g_f: f \in \fclass \} \quad \text{where} \quad g_f(\covariate, \response) \coloneqq \loss(\response, f(\covariate)), \\
&\qquad \qquad \PP_1 = \CondDist{0} \otimes \Delta_{\Response = 0}, \qquad \PP_2 = \CondDist{1} \otimes \Delta_{\Response = 1}, \\
&\qquad \qquad (p, q) = \left(\frac{\numobs \cprob_0}{\numobs + J}, 1 - \frac{\numobs \cprob_0}{\numobs + J} \right), \quad \text{and} \quad (p', q') = \left(\frac{1}{2}, \frac{1}{2} \right).
\end{align*}

Let $B \sim \text{Bin}(n, q)$ and $K \sim \text{Bin}(n, q')$, independent of everything else and each other. We then have by~\Cref{eqn:combinatorial}
\begin{align*}
\EE_{\substack{W_i \stackrel{i.i.d.}{\sim} \WW \\ \rad_i \stackrel{i.i.d.}{\sim} \text{Rad}}} \left[ \sup_{g \in \gclass} \left| \frac{1}{n} \sum_{i=1}^n \rad_i \cdot g(W_i)  \right| \right] &= \EE_{\substack{U_i \stackrel{i.i.d.}{\sim} \PP_1,\\ V_j \stackrel{i.i.d.}{\sim} \PP_2 \\ \rad_i \stackrel{i.i.d.}{\sim} \text{Rad}}}\left[\sup_{g \in \gclass} \left| \frac{1}{n} \left( \sum_{i=1}^{n - B} \rad_i \cdot g(U_i) + \sum_{j=1}^{B} \rad_{n + 1 - j} \cdot g(V_j) \right) \right|\right], \\
\EE_{\substack{W_i \stackrel{i.i.d.}{\sim} \WW' \\ \rad_i \stackrel{i.i.d.}{\sim} \text{Rad}}} \left[ \sup_{g \in \gclass} \left| \frac{1}{n} \sum_{i=1}^n \rad_i \cdot g(W_i)  \right| \right] &= \EE_{\substack{U_i \stackrel{i.i.d.}{\sim} \PP_1,\\ V_j \stackrel{i.i.d.}{\sim} \PP_2 \\ \rad_i \stackrel{i.i.d.}{\sim} \text{Rad}}}\left[\sup_{g \in \gclass} \left| \frac{1}{n} \left( \sum_{i=1}^{n - K} \rad_i \cdot g(U_i) + \sum_{j=1}^{K} \rad_{n + 1 - j} \cdot g(V_j) \right) \right|\right].
\end{align*}
We also have
\begin{align*}
&\sup_{g \in \gclass} \left| \frac{1}{n} \left( \sum_{i=1}^{n - B} \rad_i \cdot g(U_i) + \sum_{j=1}^{B} \rad_{n + 1 - j} \cdot g(V_j) \right) \right| - \sup_{g \in \gclass} \left| \frac{1}{n} \left( \sum_{i=1}^{n - K} \rad_i \cdot g(U_i) + \sum_{j=1}^{K} \rad_{n + 1 - j} \cdot g(V_j) \right) \right|\\
&\qquad \qquad \qquad  \leq \frac{2\bound}{n} \cdot |B - K|
\end{align*}
by the fact that $\loss$ is $\bound$-uniformly bounded. We can then bound
\begin{align*}
\EE|B - K| &\leq \EE|B - n q| + \EE|K - n q'| + n \cdot |q - q'| \leq \sqrt{n} + n \cdot |q - q'|,
\end{align*}
and the claim follows from linearity of expectation. 

\noindent\textbf{Proof of~\Cref{lem:mixture-rad}:} We require the following intermediate result similar to above: suppose $\gclass$ is uniformly $b$-bounded function class and we have two distributions $(\PP_1, \PP_2)$, and define the distribution $\WW = \frac{\numobs}{\numobs + J} \cdot \PP_1 + \frac{J}{\numobs +J} \cdot \PP_2$. Then we have
\begin{align}
\begin{split}
\label{eqn:mixture-comp}
&\EE_{\substack{U_i \stackrel{i.i.d.}{\sim} \PP_1 \\ V_j \stackrel{i.i.d.}{\sim} \PP_2 \\ \rad_i \stackrel{i.i.d.}{\sim} \text{Rad}}}\left[\sup_{g \in \gclass} \left| \frac{1}{\numobs + J} \left(\sum_{i=1}^\numobs \rad_i \cdot g(U_i) + \sum_{j=1}^J \rad_{\numobs + j} \cdot g(V_j) \right) \right| \right] \\
&\qquad \qquad \leq \EE_{\substack{W_i \stackrel{i.i.d}{\sim} \WW \\ \rad_i \stackrel{i.i.d.}{\sim} \text{Rad}}} \left[\sup_{g \in \gclass} \left| \frac{1}{\numobs + J} \sum_{i=1}^{\numobs + J} \rad_i \cdot g(W_i) \right| \right] + \frac{\bound}{\sqrt{\numobs + J}}.
\end{split}
\end{align}
Note the lemma follows from applying~\Cref{eqn:mixture-comp} to
\begin{align*}
&U = V = (\Covariate, \Response), \qquad \gclass = \{ g_f: f \in \fclass \} \quad \text{where} \quad g_f(\covariate, \response) \coloneqq \loss(\response, f(\covariate)), \\
&\qquad \qquad \PP_2 = \CondDist{1} \otimes \Delta_{\Response = 1}, \qquad \text{and}  \qquad \PP_1 = \cprob_0 \cdot \big(\CondDist{0} \otimes \Delta_{\Response = 0}\big) + \cprob_1 \cdot \big(\CondDist{1} \otimes \Delta_{\Response = 1} \big),
\end{align*}
where $\Delta_{\Response = 1}$ denotes the Dirac delta distribution at $\Response = 1$.

Draw $B \sim \text{Bin}(\numobs + J, \frac{J}{\numobs +J})$ that is independent of everything else. Note by~\Cref{eqn:combinatorial} we have
\begin{align}
\begin{split}
&\EE_{\substack{W_i \stackrel{i.i.d}{\sim} \WW \\ \rad_i \stackrel{i.i.d.}{\sim} \text{Rad}}} \left[\sup_{g \in \gclass} \left| \frac{1}{\numobs + J} \sum_{i=1}^{\numobs + J} \rad_i \cdot g(W_i) \right| \right] \\
&\qquad \qquad = \EE_{\substack{U_i \stackrel{i.i.d.}{\sim} \PP_1 \\ V_j \stackrel{i.i.d.}{\sim} \PP_2 \\ \rad_i \stackrel{i.i.d.}{\sim} \text{Rad} }} \left[\sup_{g \in \gclass} \left|\frac{1}{\numobs + J} \left( \sum_{i=1}^{\numobs+J-B} \rad_i \cdot g(U_i) + \sum_{j=1}^{B} \rad_{\numobs + J +1 - j} \cdot g(V_j) \right) \right| \right].
\end{split}
\end{align}
Thus
\begin{align*}
&\sup_{g \in \gclass} \left| \frac{1}{\numobs + J} \left(\sum_{i=1}^\numobs \rad_i \cdot g(U_i) + \sum_{j=1}^J \rad_{\numobs + J + 1 - j} \cdot g(V_j) \right) \right| \\
&\qquad - \sup_{g \in \gclass} \left|\frac{1}{\numobs + J} \left( \sum_{i=1}^{\numobs+J-B} \rad_i \cdot g(U_i) + \sum_{j=1}^{B} \rad_{\numobs + J - B + j} \cdot g(V_j) \right) \right| \leq \frac{2b}{\numobs + J} \cdot |B - J|,
\end{align*}
using the fact that $\gclass$ is $\bound$-uniformly bounded. The claim then follows from linearity of expectation and bounding
\begin{align*}
\EE\left[\frac{2\bound}{\numobs + J} \cdot |B - J|\right] \leq \frac{2\bound}{\numobs + J} \cdot \sqrt{\text{Var}(B)} = \frac{2\bound}{\sqrt{\numobs + J}} \cdot \sqrt{\frac{\numobs}{\numobs + J} \cdot \frac{J}{\numobs + J}} \leq \frac{\bound}{\sqrt{\numobs + J}}.
\end{align*}

\noindent\textbf{Proof of~\Cref{eqn:combinatorial}:} Define the collection of random vectors 
\begin{align*}
\left((\rad_1', Z_1), \ldots, (\rad_n', Z_{n})\right) = \pi\left((\rad_1, U_1), \ldots, (\rad_{n-B}, U_{n-B}), (\rad_{n},V_1), \ldots, (\rad_{n- B + 1}, V_B)\right)
\end{align*}
where $\pi$ is a random permutation. Observe that
\begin{align*}
&\EE\left[\sup_{g \in \gclass} \left|\frac{1}{n} \left( \sum_{i=1}^{n-B} \rad_i \cdot g(U_i) + \sum_{j=1}^{B} \rad_{n + 1 - j} \cdot g(V_j) \right) \right| \right] = \EE\left[ \sup_{g \in \gclass} \left| \frac{1}{n}\sum_{i=1}^{n} \rad_i' \cdot g(Z_i) \right| \right].
\end{align*}
It suffices to prove that $Z_i$ constructed in this manner are i.i.d.~draws from $\WW$ since \mbox{$\{\rad_i\} \indep \{U_i\} \cup \{V_j\}$}. For convenience, if $S \in \{0, 1\}^{n}$, we write $|S|$ to denote the number of $1$'s. To establish this claim, we have
\begin{align*}
\Pr(Z_1 \in A_1, \ldots, Z_{n} \in A_{n}) &= \sum_{m=0}^{n} \Pr(B = m) \cdot \Pr(Z_1 \in A_1, \ldots, Z_{n} \in A_{n} \mid B = m).
\end{align*}
We can compute
\begin{align*}
&\Pr(Z_1 \in A_1, \ldots, Z_{n} \in A_{n} \mid B = m) \\
&\qquad \qquad  = \frac{1}{\binom{n}{m}} \sum_{|S| = m} \prod_{i=1}^{n} \big\{\mathbf{1}\{S_i = 0\} \PP_1(A_i) + \mathbf{1}\{S_i = 1\} \PP_2(A_i) \big\} \qquad \text{and} \\
&\Pr(B = m) = \binom{n}{m} \cdot q^m \cdot (1-q)^{n - m}.
\end{align*}
Therefore we have
\begin{align*}
\Pr(Z_1 \in A_1, \ldots, Z_{n} \in A_{n}) &= \sum_{m=0}^n \sum_{|S| = m} \prod_{i=1}^{n} \big\{ \mathbf{1}\{S_i = 0\} \cdot (1-q) \PP_1(A_i) + \mathbf{1} \{S_i = 1\} \cdot q \PP_2(A_i) \big\} \\
&= \sum_{S} \prod_{i=1}^{n} \big\{ \mathbf{1}\{S_i = 0\} \cdot (1-q) \PP_1(A_i) + \mathbf{1} \{S_i = 1\} \cdot q \PP_2(A_i) \big\} \\
&= \prod_{i=1}^{n} \big\{ (1-q) \PP_1(A_i) + q \PP_2(A_i) \big\} = \prod_{i=1}^{n} \WW(A_i),
\end{align*}
as desired.

\subsection{Additional calculations for~\Cref{thm:slow-rates}}
\label{sec:additional-calcs}

Recall that $\{\Covariate_j'\}_{j=1}^J \stackrel{i.i.d.}{\sim} \CondDist{1}$.

\vspace{3pt}
\noindent\textbf{Proof of~\Cref{eqn:symmetrization}:} We have
\begin{align*}
&\EE\left[\frac{1}{\numobs + J}  \left( \sum_{i=1}^\numobs \loss(\Response_i, f(\Covariate_i)) + \sum_{j=1}^J \loss(1, f(\Covariate_j')) \right)\right] = \EE_{\QQ'}[\loss(\Response, f(\Covariate))].
\end{align*}
Therefore we have
\begin{align*}
&\EE\left[\sup_{f \in \fclass} \left| \frac{1}{\numobs + J} \left( \sum_{i=1}^\numobs \loss(\Response_i, f(\Covariate_i)) + \sum_{j=1}^J \loss(1, f(\Covariate_j')) \right) - \EE_{\QQ'}[\loss(\Response, f(\Covariate))] \right| \right] \\
&\qquad = \EE\left[\sup_{f \in \fclass} \left| \frac{1}{\numobs + J} \left( \sum_{i=1}^\numobs \big(\loss(\Response_i, f(\Covariate_i)) - \EE_\PP[\loss(\Response, f(\Covariate))]\big) + \sum_{j=1}^J \big(\loss(1, f(\Covariate_j')) - \EE_{\CondDist{1}}[\loss(1, f(\Covariate))] \big) \right)\right| \right] \\
&\qquad \stackrel{(*)}{\leq} 2 \EE\left[ \sup_{f\in\fclass} \left| \frac{1}{\numobs + J} \left( \sum_{i=1}^\numobs \rad_i \cdot \loss(\Response_i, f(\Covariate_i)) + \sum_{j=1}^J \rad_{\numobs + j} \cdot \loss(1, f(\Covariate_j'))\right) \right| \right],
\end{align*}
where $\rad_i$ are i.i.d.~Rademacher random variables, and step $(*)$ follows from a standard symmetrization argument.

\vspace{3pt}
\noindent\textbf{Proof of~\Cref{eqn:awesome-jason-lemma}:} Adopting the shorthand $W_\numobs(f) \coloneqq \frac{1}{\numobs + J} \sum_{i=1}^\numobs \rad_i \cdot \loss(\Response_i, f(\Covariate_i))$,
\begin{align*}
&\frac{J}{\numobs + J} \cdot \EE\left[ \sup_{f \in \fclass} \left| \frac{1}{J} \sum_{j=1}^J \rad_{\numobs + j} \cdot \loss(1, f(\Covariate_j'))  \right| \right] \\
&\qquad \qquad =  \EE\left[ \sup_{f \in \fclass} \left| \EE\left[W_\numobs(f)\right] + \frac{1}{\numobs + J} \sum_{j=1}^J \rad_{\numobs + j} \cdot \loss(1, f(\Covariate_j'))  \right| \right] \\
&\qquad \qquad = \EE\left[ \sup_{f \in \fclass} \left| \EE\left[W_\numobs(f) + \frac{1}{\numobs + J} \sum_{j=1}^J \rad_{\numobs + j} \cdot \loss(1, f(\Covariate_j')) \mid \{(\rad_{\numobs + j}, \Covariate_j')\}_{j=1}^J \right] \right| \right] \\
&\qquad \qquad \leq \EE\left[ \sup_{f \in \fclass} \left| \frac{1}{\numobs + J} \sum_{i=1}^\numobs \rad_i \cdot \loss(\Response_i, f(\Covariate_i)) + \frac{1}{\numobs + J} \sum_{j=1}^J \rad_{\numobs + j} \cdot \loss(1, f(\Covariate_j')) \right| \right] \\
&\qquad \qquad \stackrel{(a)}{\leq} \RadComp_{\numobs + J}(\QQ') + \frac{\bound}{\sqrt{\numobs + J}}  \stackrel{(b)}{\leq} \RadComp_{\numobs + J}(\QQ) + 2\bound \cdot \left| \frac{\numobs\cprob_0}{\numobs + J} - \frac{1}{2} \right| + \frac{3\bound}{\sqrt{\numobs + J}},
\end{align*}
where step (a) follows from~\Cref{lem:mixture-rad}, and step (b) follows from~\Cref{lem:annoying-stuff}.

\subsection{Proof of~\Cref{lem:new-transfer-localization}}
\label{sec:new-transfer-localization}

Observe that we have
\begin{align*}
\Term_2 = H_1 + H_2 + H_3
\end{align*}
where (recall the shorthand $\totobs = \numobs + J$)
\begin{align*}
H_1 &\coloneqq \frac{1}{\totobs} \sum_{j=1}^J \left( \ExcessLoss{\fhat}(1, \Covariate_j') - \EE_{\CondDist{1}}\left[ \ExcessLoss{\fhat}(1, \Covariate) \right] \right), \\
H_2 &\coloneqq - \frac{1}{\totobs} \sum_{j=1}^J \left( \ExcessLoss{\fhat}(1, \SCovariate_j) - \EE_{\SynDist}\left[ \ExcessLoss{\fhat}(1, \Covariate) \right] \right), \qquad \text{and} \\
H_3 &\coloneqq \frac{J}{\totobs} \cdot \left( \EE_{\CondDist{1}}\left[ \ExcessLoss{\fhat}(1, \Covariate) \right] - \EE_{\SynDist}\left[\ExcessLoss{\fhat}(1, \Covariate) \right] \right)
\end{align*}

To control $H_1$, we appeal to the same localization argument as above,
\begin{lemma}
\label{lem:other-localization}
On the event that $\norm{\fhat - \fstar}{\QQ_\Covariate} \geq \ocrit_\totobs(\QQ_\Covariate)$, we have
\begin{align*}
H_1 \leq 4L \norm{\fhat - \fstar}{\QQ_\Covariate} \left( 16\ocrit_\totobs(\QQ_\Covariate) + \sqrt{\frac{\log(\logfun(\mcrit_\totobs)/\parprob)}{\totobs}} \right)  + \frac{64L \log(\logfun(\mcrit_\totobs)/\parprob)}{\totobs},
\end{align*}
with probability exceeding $1 - \parprob$.
\end{lemma}
\noindent See~\Cref{sec:proof-other-localization} for its proof. Controlling $H_2$ requires some more care, as it involves the synthetic distribution $\SynDist$. 

\begin{lemma}
\label{lem:syn-localization}
On the event that $\|\fhat - \fstar\|_{\QQ_\Covariate} \geq \max\{ \ocrit_\totobs(\SynDist), \|\nuhat - \nustar \|_{\QQ_\Covariate}\}$, we have
\begin{align*}
H_2 \leq 2L \|\fhat - \fstar \|_{\QQ_\Covariate} \left( 8\ocrit_\totobs(\SynDist) + 2\sqrt{\frac{2\log(\logfun(\mcrit_\totobs)/\parprob)}{\totobs}} \right) + \frac{32L \log(\logfun(\mcrit_\totobs)/\parprob)}{\totobs}
\end{align*}
with probability exceeding $1 - \parprob$. 
\end{lemma}
\noindent See~\Cref{sec:proof-syn-localization} for a proof of this result. Finally to control $H_3$, we can compute
\begin{align*}
\left|  \big(\EE_{\CondDist{1}} - \EE_{\SynDist}\big)\left[\ExcessLoss{\fhat}(1, \Covariate) \right] \right| &= \left| \EE_\QQ\left[ \big(\nustar(\Covariate) - \nuhat(\Covariate)\big) \cdot \ExcessLoss{\fhat}(1, \Covariate) \right]  \right| \\
&\leq L \cdot \EE\left[ \big|\nuhat(\Covariate) - \nustar(\Covariate) \big| \cdot \big| \fhat(\Covariate) - \fstar(\Covariate) \big|  \right] \\
&\leq L \cdot \|\nuhat - \nustar \|_{\QQ_\Covariate} \cdot \|\fhat - \fstar \|_{\QQ_\Covariate}.
\end{align*}
Putting together the pieces yields the claim.

\subsection{Proof of~\Cref{lem:R-bound}}
\label{sec:proof-R-bound}

Define the function $u(K) \coloneqq \min\left\{ \frac{\totobs - K}{\numobs_0}, \frac{K}{\numobs_1 + J} \right\} \in (0, 1]$. Note that $R(K) = \lceil 1/u(K) \rceil$ and for all $v \geq 1$ we have
\begin{align*}
\lceil v \rceil \leq v + 1 \leq 2v \implies \frac{1}{\lceil v \rceil} \geq \frac{1}{2v}.
\end{align*}
Thus this implies $\frac{1}{R(K)} \geq \frac{u(K)}{2}$. Thus we have
\begin{align*}
\EE\left[\frac{1}{R(B')}\right] \geq \frac{1}{2} \EE\left[\min\left\{  \frac{\totobs - B'}{\numobs_0}, \frac{B'}{\numobs_1 + J} \right\}\right].
\end{align*}
Note that, using the fact that $\min\{a, b\} = \frac{a + b}{2} - \frac{|a - b|}{2}$, we have
\begin{align*}
\EE\left[\min\left\{  \frac{\totobs - B'}{\numobs_0}, \frac{B'}{\numobs_1 + J} \right\}\right] &= \frac{1}{2} \left( \EE\left[\frac{\totobs - B'}{\numobs_0} + \frac{B'}{\numobs_1 + J}\right] - \EE\left[\left|\frac{\totobs - B'}{\numobs_0} - \frac{B'}{\numobs_1 + J}\right|\right] \right) \\
&= 1 - \frac{1}{\totobs} \cdot \EE\left[\left|2B' - \totobs|\right|\right] \geq 1 - \frac{1}{\totobs} \cdot \sqrt{\Var(2B')}
\end{align*}
where the final step follows from the fact that $\EE[B'] = \EE[B] = \frac{\totobs}{2}$ since $\numobs_0 = \numobs_1 + J$.

Also it is straightforward to verify that $\Var(B') \leq \Var(B) = \frac{\totobs}{4}$, so the desired follows.

\subsection{Proof of~\Cref{lem:other-localization}}
\label{sec:proof-other-localization}

Define the suprema of empirical process
\begin{align*}
Z_J(t) \coloneqq \sup_{\substack{f \in \fclass \\ \norm{f - \fstar}{\QQ_\Covariate} \leq t}} \left| \frac{1}{J} \left(\sum_{j=1}^J \ExcessLoss{f}(1, \Covariate_j') - \EE_{\CondDist{1}}\left[\ExcessLoss{f}(1, \Covariate)\right] \right) \right|.
\end{align*}
Applying~\Cref{lem:emp-conc} to the function class (where each function is the same for all $i$)
\begin{align*}
\gclass \coloneqq \left\{ \ExcessLoss{f}(1, x) - \EE_{\CondDist{1}}\left[ \ExcessLoss{f}(1, \Covariate)\right] : \quad f \in \fclass, \, \norm{f - \fstar}{\QQ_\Covariate} \leq t\right\},
\end{align*}
with $\tau = 1$ yields
\begin{align*}
Z_J(t) \geq 2\EE[Z_J(t)] + \sqrt{\sigma^2(\gclass)} \cdot \sqrt{\frac{2s}{J}} + \frac{16Ls}{J}
\end{align*}
with probability at most $e^{-s}$. We claim that, for $t \geq \ocrit_\totobs(\QQ_\Covariate)$, 
\begin{align}
\label{eqn:emp-calcs-II}
\EE[Z_J(t)] \leq \frac{16L\totobs}{J} \cdot t\, \ocrit_\totobs(\QQ_\Covariate) + \frac{128L}{J \cdot 2^\totobs}, \qquad \text{and} \qquad \sigma^2(\gclass) \leq 2L^2t^2.
\end{align}
Thus we have
\begin{align*}
\frac{J}{\totobs} \cdot Z_J(t) \geq Q(t, s) \coloneqq 32Lt \, \ocrit_\totobs(\QQ_\Covariate) + 2Lt \sqrt{\frac{s}{\totobs}} + \frac{32Ls}{\totobs} \quad \text{with probability at most $e^{-s}$.}
\end{align*}
Thus applying~\Cref{lem:peeling} to the above with $U = \max\left\{\ocrit_\totobs(\QQ_\Covariate), \norm{\fhat - \fstar}{\QQ_\Covariate} \right\}$ yields
\begin{align*}
\frac{J}{\totobs} \cdot Z_J\left(\norm{\fhat - \fstar}{\QQ_\Covariate}\right) &\leq 4L \cdot \max\left\{ \ocrit_\totobs(\QQ_\Covariate), \norm{\fhat - \fstar}{\QQ_\Covariate} \right\} \left( 16\ocrit_\totobs(\QQ_\Covariate) + \sqrt{\frac{\log(\logfun(\ocrit_\totobs(\QQ_\Covariate))/\parprob)}{\totobs}} \right) \\
& \qquad \qquad \qquad + \frac{64L \log(\logfun(\ocrit_\totobs(\QQ_\Covariate))/\parprob)}{\totobs}
\end{align*}
with probability exceeding $1 - \parprob$. The claim then follows from the fact that $\rho$ is a decreasing function and $\mcrit_{\totobs} \leq \ocrit_\totobs(\QQ_\Covariate)$.

\noindent \textbf{Proof of~\Cref{eqn:emp-calcs-II}:} To control the supremum of the variances, we have
\begin{align*}
\sigma^2(\gclass) &= \sup_{\|f - \fstar \|_{\QQ_\Covariate} \leq t} \Var_{\CondDist{1}}(\ExcessLoss{f}(1, \Covariate)) \\
&\leq \sup_{\|f - \fstar \|_{\QQ_\Covariate} \leq t}\EE_{\CondDist{1}}\left[\big(\loss(1, f(\Covariate)) - \loss(1, \fstar(\Covariate)) \big)^2\right] \\
&\leq L^2 \sup_{\|f - \fstar \|_{\QQ_\Covariate} \leq t} \|f - \fstar \|_{\CondDist{1}}^2 \leq 2 L^2 t^2.
\end{align*}
For the expectation, we have
\begin{align*}
\EE[Z_J(t)] &= \EE\left[\sup_{\norm{f - \fstar}{\QQ_\Covariate} \leq t} \left| \frac{1}{J} \left(\sum_{j=1}^J \ExcessLoss{f}(1, \Covariate_j') - \EE_{\CondDist{1}}\left[\ExcessLoss{f}(1, \Covariate)\right] \right) \right| \right] \\
&\leq 2 \, \EE\left[ \sup_{\norm{f - \fstar}{\QQ_\Covariate} \leq t} \left| \frac{1}{J} \sum_{j=1}^J \rad_j \cdot \ExcessLoss{f}(1, \Covariate_j') \right| \right] \\
&\leq 4L \cdot \EE\left[ \sup_{\norm{f - \fstar}{\QQ_\Covariate} \leq t} \left| \frac{1}{J} \sum_{j=1}^J \rad_j \cdot \big(f(\Covariate_j') - \fstar(\Covariate_j') \big) \right| \right] = \frac{4L}{J} \cdot Z(0, J; t),
\end{align*}
where we have used the notation from~\Cref{eqn:Z-defn}. Then, using monotonicity and~\Cref{lem:jason-blocking}, we have that
\begin{align*}
Z(0, J; t) \leq Z(\numobs_0, \numobs_1 + J; t) \leq 4\totobs \cdot \RadComp_\totobs(t; \QQ_\Covariate) + \frac{32}{2^{\totobs}}.
\end{align*}
Since $t \geq \ocrit_\totobs(\QQ_\Covariate)$, we have $\RadComp_\totobs(t; \QQ_\Covariate) \leq t \, \ocrit_\totobs(\QQ_\Covariate)$, yielding that
\begin{align*}
\EE[Z_J(t)] \leq \frac{16L\totobs}{J} \cdot t\, \ocrit_\totobs(\QQ_\Covariate) + \frac{128L}{J \cdot 2^\totobs},
\end{align*}
as desired.

\subsection{Proof of~\Cref{lem:syn-localization}}
\label{sec:proof-syn-localization}

We proceed as before, albeit with some differences. Consider the empirical process suprema
\begin{align*}
Z(t) \coloneqq \sup_{\substack{f \in \fclass \\ \|f - \fstar \|_{\SynDist} \leq t}} \left| \frac{1}{J} \sum_{j =1}^{J} \ExcessLoss{f}(1, \SCovariate_j) - \EE_{\SynDist}\left[ \ExcessLoss{f}(1, \Covariate) \right] \right|.
\end{align*}
Applying~\Cref{lem:emp-conc} yields, with probability at most $e^{-s}$,
\begin{align*}
Z(t) \geq 2\EE[Z(t)] + \sqrt{\sigma^2(\gclass)} \cdot \sqrt{\frac{2s}{J}} + \frac{16Ls}{J}.
\end{align*}
We claim that, for $t \geq \ocrit_\totobs(\SynDist)$,
\begin{align}
\label{eqn:emp-calcs-2}
\EE[Z(t)] \leq \frac{4LM}{J} \cdot t \,\ocrit_\totobs(\SynDist), \qquad \text{and} \qquad \sigma^2(\gclass) \leq L^2 t^2.
\end{align}
Thus applying~\Cref{lem:peeling} will yield 
\begin{align*}
H_2 &\leq L\max\left\{ \ocrit_\totobs(\SynDist), \|\fhat - \fstar \|_{\SynDist} \right\} \cdot \left(8\ocrit_\totobs(\SynDist) + 2\sqrt{\frac{2\log(\logfun(\ocrit_\totobs(\SynDist))/\parprob)}{\totobs}}\right) \\
&\qquad \qquad \qquad \qquad + \frac{32L\log(\logfun(\ocrit_\totobs(\SynDist))/\parprob)}{\totobs}
\end{align*}
with probability exceeding $1 - \parprob$. 
To complete the proof, observe that it suffices to establish 
\begin{align*}
\|\fhat - \fstar \|_{\SynDist} \leq 2\max\left\{ \|\fhat - \fstar\|_{\QQ_\Covariate}, \|\nuhat - \nustar \|_{\QQ_\Covariate} \right\},
\end{align*}
and the claim follows because we are on the event $\|\fhat - \fstar \|_{\QQ_\Covariate} \geq \max\{\ocrit_\totobs(\SynDist), \|\nuhat - \nustar \|_{\QQ_\Covariate}\}$. Indeed we have
\begin{align*}
\|\fhat - \fstar \|_{\SynDist}^2 &= \|\fhat - \fstar \|_{\CondDist{1}}^2 + \EE_\QQ\left[\big(\nuhat(\Covariate) - \nustar(\Covariate) \big) \cdot \big(\fhat(\Covariate) - \fstar(\Covariate))^2 \right] \\
&\leq 2 \left( \|\fhat - \fstar \|_{\QQ_\Covariate}^2 +  \EE_\QQ\left[\big|\nuhat(\Covariate) - \nustar(\Covariate) \big| \cdot \big|\fhat(\Covariate) - \fstar(\Covariate)| \right]\right) \\
&\leq 2 \left( \|\fhat - \fstar \|_{\QQ_\Covariate}^2 + \|\fhat - \fstar \|_{\QQ_\Covariate} \cdot \|\nuhat - \nustar \|_{\QQ_\Covariate} \right) \leq 4 \max\left\{ \|\fhat - \fstar\|_{\QQ_\Covariate}^2, \|\nuhat - \nustar \|_{\QQ_\Covariate}^2 \right\}.
\end{align*}
The claim then follows from the fact that $\rho$ is a decreasing function and $\mcrit_{\totobs} \leq \ocrit_\totobs(\QQ_\Covariate)$.

\noindent \textbf{Proof of~\Cref{eqn:emp-calcs-2}:} For the variance quantity, we have
\begin{align*}
\sigma^2(\gclass) &= \sup_{\|f - \fstar \|_{\SynDist} \leq t} \Var_{\SynDist}(\ExcessLoss{f}(1, \Covariate)) \leq L^2 \sup_{\|f - \fstar \|_{\SynDist} \leq t} \|f - \fstar \|_{\SynDist}^2 = L^2 t^2. 
\end{align*}
For the expectation quantity, we have
\begin{align*}
\EE[Z(t)] &= \EE\left[ \sup_{\|f - \fstar\|_{\SynDist} \leq t} \left| \frac{1}{J} \sum_{j=1}^J \ExcessLoss{f}(1, \SCovariate_j) - \EE_{\SynDist}\left[ \ExcessLoss{f}(1, \Covariate) \right] \right| \right] \\
&\leq 2\EE\left[ \sup_{\|f - \fstar\|_{\SynDist} \leq t} \left| \frac{1}{J} \sum_{j=1}^J \rad_j \cdot \ExcessLoss{f}(1, \SCovariate_j) \right| \right] \\
&\leq 4L \cdot \EE\left[ \sup_{\|f - \fstar\|_{\SynDist} \leq t} \left| \frac{1}{J} \sum_{j=1}^J \rad_j \cdot \big( f(\SCovariate_j) - \fstar(\SCovariate_j) \big)\right|  \right] \\
&= 4L \cdot \RadComp_J(t; \SynDist) \leq \frac{4L\totobs}{J} \cdot \RadComp_\totobs(t; \SynDist) \leq \frac{4L\totobs}{J} \cdot t \, \ocrit_\totobs(\SynDist),
\end{align*}
where we have used the monotonicity argument in the second to last step.

\subsection{Additional calculations for~\Cref{thm:fast-rates}}
\label{sec:additional-calcs-2}

\noindent \textbf{Proof of~\Cref{eqn:monotonic}:} By induction and symmetry, it suffices to show that $Z(n_0 + 1, n_1; t) \geq Z(n_0, n_1; t)$. Define the $\sigma$-algebra $\mathscr{S}_{n_0, n_1} \coloneqq \sigma(U_1, \ldots, U_{n_0}, V_1, \ldots, V_{n_1}, \rad_1, \ldots, \rad_{n_0 + n_1})$, and for each $g \in \fclass^*(t)$ define $S_g \coloneqq \sum_{i=1}^{n_0} \rad_i \cdot g(U_i) + \sum_{j=1}^{n_1} \rad_{n_0 + j} \cdot g(V_j)$. Note that $S_g$ is $\mathscr{S}_{n_0, n_1}$ measurable. Since $(\rad_{n_0 + n_1 + 1}, U_{n_0 + 1}) \indep \mathscr{S}_{n_0, n_1}$, an application of Jensen's inequality yields
\begin{align*}
\EE\left[ \sup_{g \in \fclass^*(t)} \big|S_g + \rad_{n_0 + n_1 + 1} \cdot g(U_{n_0 + 1}) \big| \mid \mathscr{S}_{n_0, n_1} \right] &\geq \sup_{g \in \fclass^*(t)} \left|S_g + \EE\big[\rad_{n_0 + n_1 + 1} \cdot g(U_{n_0 + 1})\big]  \right| \\
&= \sup_{g \in \fclass^*(t)} |S_g|.
\end{align*}
Thus taking expectation implies $Z(n_0 + 1, n_1; t) \geq Z(n_0, n_1; t)$, as desired.

\section{Auxiliary calculations}

\subsection{Proof of~\Cref{prop:plug-in}}
\label{sec:proof-plugin}

We define the function $h(x) \defn \frac{\cprob_0 x}{(\cprob_0 - \cprob_1) x + \cprob_1}$. Observe that for any $u, v \in [0, 1]$ we have
\begin{align*}
h(u) - h(v) = \frac{\cprob_0 \cprob_1 (u - v)}{\big( (\cprob_0 - \cprob_1) u + \cprob_1 \big) \big( (\cprob_0 - \cprob_1) v + \cprob_1 \big)}.
\end{align*}
Furthermore, since $\QQ_\Covariate = \frac{1}{2} \CondDist{0} + \frac{1}{2} \CondDist{1}$ and $\PP_\Covariate = \cprob_0 \CondDist{0} + \cprob_1 \CondDist{1}$ we can easily verify (where we recall $\gstar = \frac{\cprob_1 d\CondDist{1}}{\cprob_0 d\CondDist{0} + \cprob_1 d\CondDist{1}}$)
\begin{align*}
\frac{d\QQ_\Covariate}{d\PP_\Covariate} &= \frac{1}{2} \left( \frac{d\CondDist{0}}{\cprob_0 \CondDist{0} + \cprob_1 \CondDist{1}} + \frac{d\CondDist{1}}{\cprob_0 \CondDist{0} + \cprob_1 \CondDist{1}} \right) \\
&= \frac{1}{2} \left(\frac{1 - \gstar}{\cprob_0} + \frac{\gstar}{\cprob_1} \right) = \frac{\cprob_1 + (\cprob_0 - \cprob_1) \gstar}{2\cprob_0 \cprob_1}.
\end{align*}
Thus we have
\begin{align*}
\norm{\fhat_{\text{PLUG}} - \fstar}{\QQ_\Covariate}^2 &= \EE_{\Covariate \sim \QQ_\Covariate} \left[ \left( h(\ghat(\Covariate)) - h(\gstar(\Covariate))\right)^2\right] \\
&= \EE_{\Covariate \sim \PP_\Covariate}\left[ \left( h(\ghat(\Covariate)) - h(\gstar(\Covariate))\right)^2 \cdot \frac{d\QQ_\Covariate}{d\PP_\Covariate}(\Covariate)\right] \\
&= \EE_{\Covariate \sim \PP_\Covariate}\left[ \frac{\cprob_0 \cprob_1 \big(\ghat(\Covariate) - \gstar(\Covariate)\big)^2}{2\big( (\cprob_0 - \cprob_1) \ghat(\Covariate) + \cprob_1 \big)^2 \big( (\cprob_0 - \cprob_1) \gstar(\Covariate) + \cprob_1 \big)} \right] \\
&\leq \frac{\cprob_0}{2\cprob_1^2} \cdot \norm{\ghat - \gstar}{\PP_\Covariate}^2
\end{align*}
where the final inequality follows form the fact that $((\cprob_0 - \cprob_1) u + \cprob_1) \geq \cprob_1$ for all $u \in [0, 1]$.

\subsection{Calculations for~\Cref{sec:intuition}}
\label{app:intuition-calc}

\noindent\textbf{Proof of~\Cref{eqn:typeII-Qnorm}:} Note that we have the upper bound
\begin{align*}
\left| \PP(\qhat(\Covariate) = 0 \mid \Response = 1) - \PP(\qstar(\Covariate) = 0 \mid \Response = 1) \right| &\leq \EE_{\Covariate \sim \CondDist{1}}\left[\left|\mathbf{1}\{\qhat(\Covariate) = 0\} - \mathbf{1}\{\qstar(\Covariate) = 0\} \right|\right] \\
&= 2\EE_{\Covariate \sim \QQ_\Covariate}\left[\fstar(\Covariate) \cdot \left|\mathbf{1}\{\qhat(\Covariate) = 0\} - \mathbf{1}\{\qstar(\Covariate) = 0\} \right| \right] \\
&= 2\EE_{\Covariate \sim \QQ_\Covariate}\left[ \fstar(\Covariate) \cdot \mathbf{1}\{\qhat(\Covariate) \neq \qstar(\Covariate)\} \right].
\end{align*}
Now suppose $\fstar$ is bounded away from the margin, i.e., $|2\fstar(\Covariate) - 1| \geq \gamma$ a.s., for some constant $\gamma > 0$. We remark that this assumption can be relaxed by introducing a Tsybakov margin condition, although with a different rate. We can then bound
\begin{align*}
\QQ(\Response \neq \qhat(\Covariate)) - \QQ(\Response \neq \qstar(\Covariate)) = \EE_{\Covariate \sim \QQ_\Covariate}\left[ \left|2\fstar(\Covariate) - 1\right| \mathbf{1}\{\qhat(\Covariate) \neq \qstar(\Covariate)\} \right] \geq \gamma \QQ_\Covariate(\qhat(\Covariate) \neq \qstar(\Covariate)).
\end{align*}
Therefore we have
\begin{align*}
\EE_{\Covariate \sim \QQ_\Covariate}\left[ \fstar(\Covariate) \cdot \mathbf{1}\{\qhat(\Covariate) \neq \qstar(\Covariate)\} \right] &\leq \EE_{\Covariate \sim \QQ_\Covariate}\left[ |\fstar(\Covariate)-\tfrac{1}{2}| \cdot \mathbf{1}\{\qhat(\Covariate) \neq \qstar(\Covariate)\} \right] + \tfrac{1}{2} \QQ_\Covariate\left( \qhat(\Covariate) \neq \qstar(\Covariate) \right) \\
&\leq \left( \frac{1}{2} + \frac{1}{2\gamma} \right) \cdot \left( \QQ(\qhat(\Covariate) \neq \Response) - \QQ(\qstar(\Covariate) \neq \Response) \right).
\end{align*}

The first bound follows from~\citet{devroye2013probabilistic} where
\begin{align*}
\QQ(\Response \neq \qhat(\Covariate)) - \QQ(\Response \neq \qstar(\Covariate)) \leq 2\EE_{\QQ_\Covariate}\left[\left|\fhat(\Covariate) - \fstar(\Covariate)\right|\right] \leq 2 \|\fhat - \fstar\|_{\QQ_\Covariate}.
\end{align*}
The second bound follows that for most convex losses,~\citet{bartlett2006convexity} gives a construction for an increasing convex function $\psi_\loss$ with $\psi_\loss(0) = 0$ such that
\begin{align*}
\psi_\loss\left(\QQ(\Response \neq \qhat(\Covariate)) - \QQ(\Response \neq \qstar(\Covariate))  \right) \leq \EE_\QQ\left[\loss(\Response, \fhat(\Covariate)) - \loss(\Response, \fstar(\Covariate))  \right].
\end{align*}

\noindent\textbf{Computing the population minimizer:} We adopt the shorthands
\begin{align*}
c_0 \coloneqq \cprob_0 \cdot \frac{\numobs}{\numobs + J}, \qquad c_1 \coloneqq \cprob_1 \cdot \frac{\numobs}{\numobs + J}, \quad \text{and} \quad c_2 \coloneqq \frac{J}{\numobs + J}.
\end{align*}
Observe that we can write the true and estimated target distributions as 
\begin{align*}
\QQ &= c_0 \cdot \CondDist{0} \otimes \Delta_{\Response = 0} + (c_1 + c_2) \cdot \CondDist{1} \otimes \Delta_{\Response = 1}, \qquad \text{and} \\
\widehat{\QQ} &= c_0 \cdot \CondDist{0} \otimes \Delta_{\Response = 0} + c_1 \cdot \CondDist{1} \otimes \Delta_{\Response = 1} + c_2 \cdot \SynDist \otimes \Delta_{\Response = 1}.
\end{align*}
For the binary cross-entropy loss
\begin{align*}
\loss(\Response, f(\Covariate)) = - \Response \log f(\Covariate) - (1 - \Response) \log(1 - f(\Covariate)),
\end{align*}
consider solving the minimization problem over some distribution $R$
\begin{align*}
\min_{f \in \fclass} \EE_{(\Covariate, \Response) \sim R}\left[ \loss(\Response, f(\Covariate)) \right].
\end{align*}
It is well-known that the minimizer is given by the pointwise Bayes posterior:
\begin{align*}
\fstar_R(\covariate) = \argmin_{u \in (0, 1)} \EE_{\Response \sim R_{\Response \mid \Covariate = \covariate}} \left[ -\Response \log u - (1 - \Response) \log(1-u) \right] = R(\Response = 1 \mid \Covariate = \covariate).
\end{align*}
Furthermore observe that
\begin{align*}
\EE\left[\frac{1}{\numobs + J} \left\{ \sum_{i=1}^\numobs \loss(\Response_i, f(\Covariate_i)) + \sum_{j=1}^J \loss(1, f(\SCovariate_j)) \right\} \right] = \EE_{\widehat{\QQ}}\left[\loss(\Response, f(\Covariate)) \right].
\end{align*}
Thus we can easily compute 
\begin{align*}
\fstar(\Covariate) &= \frac{(c_1 + c_2) \cdot d\CondDist{1}}{c_0 \cdot d\CondDist{0} + (c_1 + c_2) \cdot d\CondDist{1}} \qquad \text{and} \\
\ftil(\Covariate) &= \frac{c_1 \cdot d\CondDist{1} + c_2 \cdot d\SynDist}{c_0 \cdot d\CondDist{0}  + c_1 \cdot d\CondDist{1} + c_2 \cdot d\SynDist},
\end{align*}
as desired. Here $J = (2\cprob_0 - 1) \numobs$ is chosen such that $c_1 + c_2 = \frac{1}{2}$.

\section{Proof of further results} 
\label{sec:proof-further}

\subsection{Proof of results in~\Cref{sec:discrete-synthetic}}
\label{sec:proof-instantiations}

We state a result whose proof is similar to that of~\Cref{thm:slow-rates}. See~\Cref{sec:proof-generic-slow} for a proof of this result.
\begin{lemma}
\label{lem:generic-slow}
Under the conditions of~\Cref{thm:slow-rates}, consider Algorithm~\ref{AlgRebal} with an arbitrary $\SynDist$. We have, with probability exceeding $1 - \parprob$,
\begin{align*}
\EE_\QQ\left[ \loss(\Response, \fhat(\Covariate)) - \loss(\Response, \fstar(\Covariate)) \right] &\leq 8\RadComp_{\numobs + J}(\QQ) + \frac{4J}{\numobs + J} \cdot \RadComp_{J}(\SynDist) \\
&\qquad  + (12\sqrt{2\log(1/\parprob)} + 24) \cdot \frac{\bound}{\sqrt{\numobs + J}} + 20\bound \cdot \left| \frac{\numobs \cprob_0}{\numobs + J} - \frac{1}{2} \right| \\
&\qquad  + \frac{2J}{\numobs + J} \cdot \sup_{f \in \fclass} \left| \left(\EE_{\SynDist} - \EE_{\CondDist{1}} \right)\left[ \loss(1, f(\Covariate)) \right] \right|.
\end{align*}
\end{lemma}

\vspace{3pt}
\noindent\textbf{Proof of~\Cref{eqn:bootstrapping}:} Key to our statement is the following lemma.
\begin{lemma}[Bootstrapping]
\label{lem:bootstrapping}
Under the conditions of~\Cref{lem:generic-slow}, we have if $\SynDist = \widehat{\PP}_{\dataset_1}$, then, each with probability exceeding $1 - \parprob$,
\begin{subequations}
\begin{align}
\begin{split}
\label{eqn:bootstrapping-a}
\sup_{f \in \fclass} \left| \left(\EE_{\SynDist} - \EE_{\CondDist{1}} \right)\left[ \loss(1, f(\Covariate)) \right] \right| &\leq  2\RadComp_{\numobs_1}(\CondDist{1}) + 2\sqrt{2\log(1/\parprob)} \cdot \frac{\bound}{\sqrt{\numobs_1}} ,
\end{split}
\end{align}
\begin{align}
\begin{split}
\label{eqn:bootstrapping-b}
\RadComp_{J}(\SynDist ) &\leq 4\big( \sqrt{\tfrac{\numobs_1}{J} \log(2\numobs_1)} + \tfrac{\numobs_1}{J} \log(2\numobs_1) \big)  \RadComp_{\numobs_1}(\CondDist{1})  + 2\sqrt{2\log(1/\parprob)} \cdot \frac{\bound}{\sqrt{\numobs_1}}.
\end{split}
\end{align}
\end{subequations}
\end{lemma}
\noindent See~\Cref{sec:proof-bootstrapping} for a proof of this lemma. Combining~\Cref{lem:generic-slow} and~\Cref{lem:bootstrapping}, we obtain
\begin{align*}
\EE_{\QQ}\left[\loss(\Response, \fhat(\Covariate)) - \loss(\Response, \fstar(\Covariate)) \right] &\lesssim \RadComp_{\numobs + J}(\QQ) + \bound \cdot \left| \tfrac{\numobs \cprob_0}{\numobs + J} - \tfrac{1}{2} \right| + t_{\numobs + J}(\parprob) + \bound \sqrt{\tfrac{\log(\tfrac{1}{\parprob})}{\numobs_1}} \\
&\qquad  + \left(1 + \sqrt{\tfrac{\numobs_1}{J} \log(2\numobs_1)}\right)^2 \cdot \RadComp_{\numobs_1}(\CondDist{1})
\end{align*}
with probability exceeding $1 - \parprob$. Using $\numobs_1 - \cprob_1 \numobs = O_\PP(\sqrt{\numobs})$, and
\begin{align*}
\RadComp_{\numobs_1}(\CondDist{1}) &= 2\, \EE_{\Covariate_i \stackrel{i.i.d.}{\sim} \CondDist{1}, \, \rad_i \stackrel{i.i.d.}{\sim} \text{Rad}}\left[\sup_{f \in \fclass} \left| \frac{1}{2\numobs_1} \sum_{i=1}^{\numobs_1} \rad_i \cdot \loss(1, f(\Covariate_i)) \right|\right] \\
&\leq 2\, \EE_{\substack{\Covariate_i \stackrel{i.i.d.}{\sim} \CondDist{1} \\ \Covariate'_i \stackrel{i.i.d.}{\sim} \CondDist{0}}, \, \rad_i \stackrel{i.i.d.}{\sim} \text{Rad}}\left[\sup_{f \in \fclass} \left|\frac{1}{2\numobs_1}  \sum_{i=1}^{\numobs_1}\left( \rad_i \cdot \loss(1, f(\Covariate_i)) + \rad_{\numobs_1 + i} \cdot \loss(0, f(\Covariate_i')) \right) \right| \right]
\end{align*}
which implies $\RadComp_{\numobs_1}(\CondDist{1}) \leq 2 \RadComp_{2\numobs_1}(\QQ) + \frac{2\bound}{\sqrt{2\numobs_1}},$ where this follows from~\Cref{eqn:mixture-comp}.

\vspace{3pt}
\noindent\textbf{Proof of~\Cref{eqn:SMOTE-bound}:} The following lemma establishes our claim. Define the terms
\begin{align*}
H_1 \coloneqq\left( \frac{3k + 4d \cdot \log(32\numobs_1/\parprob)}{c_d \numobs_1}  \right)^{1/d}, \quad \text{and} \quad H_2 \coloneqq 6D \left( \frac{k}{\numobs_1} \right)^{1/d} + 2D\sqrt{2\log(2/\parprob)} \cdot \frac{k5^d + 1}{\sqrt{\numobs_1}}.
\end{align*}

\begin{lemma}[SMOTE]
\label{lem:SMOTE}
Under the conditions of~\Cref{lem:generic-slow} and Equations~\eqref{eqn:SMOTE-lipschitz} and~\eqref{eqn:SMOTE-lower-bound}, we have if $\SynDist = \widehat{\PP}_\Covariate^{\text{SMOTE}}$, then
\begin{subequations}
\begin{align}
\sup_{f \in \fclass} \left|\big( \EE_{\SynDist} - \EE_{\CondDist{1}} \big)\left[\loss(1, f(\Covariate)) \right] \right| &\leq 2\RadComp_{\numobs_1}(\CondDist{1}) + L \min\{ H_1, H_2 \} \qquad \notag \\
\label{eqn:SMOTE-a}
&\quad + 2\sqrt{2\log(2/\parprob)} \,\frac{\bound}{\sqrt{\numobs_1}}, \qquad \text{and}
\\ \RadComp_J(\SynDist) &\leq 4\left(
\sqrt{\frac{\numobs_1}{J}\log(2\numobs_1)} + \frac{\numobs_1}{J}\log(2\numobs_1)\right)\RadComp_{\numobs_1}(\CondDist{1})\notag   \\
\label{eqn:SMOTE-b} 
&\quad + L\min\{H_1, H_2\} + 2\sqrt{2\log(1/\parprob)} \,\frac{\bound}{\sqrt{\numobs_1}}.
\end{align}
\end{subequations}
each with probability exceeding $1 - \parprob$. 
\end{lemma}
\noindent See~\Cref{sec:proof-SMOTE} for a proof of this result. Combining Lemmas~\ref{lem:generic-slow} and~\ref{lem:SMOTE}, we get
\begin{align*}
\EE_\QQ\left[ \loss(\Response, \fhat(\Covariate))  -  \loss(\Response, \fstar(\Covariate)) \right] &\lesssim \RadComp_{\numobs + J}(\QQ) + \bound \cdot \left| \tfrac{\numobs \cprob_0}{\numobs + J} - \tfrac{1}{2} \right| + t_{\numobs + J}(\parprob) + \bound \sqrt{\tfrac{\log(\tfrac{1}{\parprob})}{\numobs_1}} \\
&\qquad  + \left(1 + \sqrt{\tfrac{\numobs_1}{J} \log(2\numobs_1)}\right)^2 \cdot \RadComp_{\numobs_1}(\CondDist{1}) \\
&\qquad  + 6LD \left(\tfrac{k}{\numobs_1}\right)^{1/d} + L \cdot \min\{H_1, H_2\}
\end{align*}

\subsection{Proof of~\Cref{cors:random-J-slow}}
\label{sec:proof-random-J-slow}

By the argument given in~\Cref{sec:proof-slow-rates}, we have
\begin{align*}
&\EE_\QQ\left[\loss(\Response, \fhat(\Covariate)) - \loss(\Response, \fstar(\Covariate)) \right] \\
&\qquad  \qquad \leq 2\sup_{f \in \fclass} \left|\frac{1}{\numobs + J} \left( \sum_{i=1}^\numobs \loss(\Response_i, f(\Covariate_i)) + \sum_{j=1}^J \loss(1, f(\SCovariate_j)) \right) - \EE_\QQ[\loss(\Response, f(\Covariate))] \right| \\
&\qquad  \qquad \leq 2\sup_{f \in \fclass} \left|\frac{1}{\numobs + J} \left( \sum_{i=1}^\numobs \loss(\Response_i, f(\Covariate_i)) + \sum_{j=1}^{\Jstar} \loss(1, f(\SCovariate_j)) \right) - \EE_\QQ[\loss(\Response, f(\Covariate))] \right| \\
&\qquad \qquad \qquad \qquad + 2\sup_{f \in \fclass} \left| \frac{1}{\numobs + J} \sum_{j = \min\{J, \Jstar\} + 1}^{\max\{J, \Jstar\}} \loss(1, f(\SCovariate_j)) \right|.
\end{align*}
Also we have % p. 43
\begin{align*}
&\sup_{f \in \fclass} \left|\frac{1}{\numobs + J} \left( \sum_{i=1}^\numobs \loss(\Response_i, f(\Covariate_i)) + \sum_{j=1}^{\Jstar} \loss(1, f(\SCovariate_j)) \right) - \EE_\QQ[\loss(\Response, f(\Covariate))] \right| \\
&\qquad \qquad \leq \frac{\numobs + \Jstar}{\numobs + J} \cdot \sup_{f \in \fclass} \left|\frac{1}{\numobs + \Jstar} \left( \sum_{i=1}^\numobs \loss(\Response_i, f(\Covariate_i)) + \sum_{j=1}^{\Jstar} \loss(1, f(\SCovariate_j)) \right) - \EE_\QQ[\loss(\Response, f(\Covariate))] \right| \\
&\qquad \qquad \qquad \qquad + \frac{|J - \Jstar|}{\numobs + J} \cdot \sup_{f \in \fclass} \left| \EE_\QQ[\loss(\Response, f(\Covariate))] \right|.
\end{align*}
Since $\loss$ is $\bound$-uniformly bounded, we have
\begin{align*}
&\frac{|J - \Jstar|}{\numobs + J} \cdot \sup_{f \in \fclass} \left| \EE_\QQ[\loss(\Response, f(\Covariate))] \right| \leq \bound \cdot \frac{|J - \Jstar|}{\numobs + J}, \qquad \text{and} \\
&\sup_{f \in \fclass} \left| \frac{1}{\numobs + J} \sum_{j = \min\{J, \Jstar\} + 1}^{\max\{J, \Jstar\}} \loss(1, f(\SCovariate_j)) \right| \leq \bound \cdot \frac{|J - \Jstar|}{\numobs + J}.
\end{align*}
The proof of~\Cref{thm:slow-rates} also establishes 
\begin{align*}
2\sup_{f \in \fclass} \left|\frac{1}{\numobs + \Jstar} \left( \sum_{i=1}^\numobs \loss(\Response_i, f(\Covariate_i)) + \sum_{j=1}^{\Jstar} \loss(1, f(\SCovariate_j)) \right) - \EE_\QQ[\loss(\Response, f(\Covariate))] \right| \leq \Gamma(\numobs, \Jstar).
\end{align*}

\subsection{Proof of~\Cref{cors:target-mixtures}}
\label{sec:proof-target-mixtures}

Recall the definition of the mixture distribution $
\QQ' \coloneqq \frac{\numobs}{\numobs + J} \cdot \PP_{\Covariate, \Response} + \frac{J}{\numobs + J} \cdot \CondDist{1} \otimes \Delta_{\Response = 1}.
$
Upon inspection, it suffices to control $\dTV{\QQ^\star}{\QQ'}$ as this is a source of the dependence on $\QQ'$ in the proof. Indeed we have $\dTV{\QQ^\star}{\QQ'} \leq \left| \frac{\numobs \cprob_0}{\numobs + J} - \cprob_0^\star \right|$. Furthermore by the proof of~\Cref{lem:annoying-stuff} we have
\begin{align*}
\RadComp_{\numobs + J}(\QQ') \leq \RadComp_{\numobs + J}(\QQ^*) + \frac{2\bound}{\sqrt{N+J}} + 2\bound \cdot \left| \frac{\numobs\cprob_0}{\numobs + J} - \cprob_0^* \right|.
\end{align*}
and consequently we have
%R
\begin{align*}
\frac{J}{\numobs + J} \cdot \RadComp_J(\CondDist{1}) \leq \RadComp_{\numobs + J}(\QQ^*) + \frac{3\bound}{\sqrt{\numobs + J}} + 2\bound \cdot \left| \frac{\numobs \cprob_0}{\numobs + J} - \cprob_0^* \right|.
\end{align*}
With these substitutions in place, the guarantee of~\Cref{cors:target-mixtures} follows from the same calculations, with just $\left| \frac{\numobs \cprob_0}{\numobs + J} - \cprob_0^* \right|$ as the appropriate mixture bias term.

\subsection{Proof of~\Cref{prop:undersampling}}
\label{sec:proof-undersampling}

Via a standard empirical risk minimization approach, we have it suffices to control 
\begin{align*}
2 \sup_{f \in \fclass} \left| \frac{1}{K + \numobs_1} \left( \sum_{i \in H_0} \loss(0, f(\Covariate_i)) + \sum_{j \in I_1} \loss(1, f(\Covariate_j)) - \EE_\QQ[\loss(\Response, f(\Covariate))]\right) \right|.
\end{align*}
To control this, we proceed in the same manner as the proof of~\Cref{sec:app-bounded-diff}; define the mixture distribution
\begin{align*}
\tilde{\QQ} \coloneqq \frac{K}{K + \numobs_1} \cdot \CondDist{0} \otimes \Delta_{\Response = 0} + \frac{\numobs_1}{K + \numobs_1} \cdot \CondDist{1} \otimes \Delta_{\Response = 1}.
\end{align*}
We have that $\dTV{\QQ}{\tilde{\QQ}} \leq | \frac{K}{K + \numobs_1} - \frac{1}{2}|$, so
\begin{align*}
\sup_{f \in \fclass} \left| \EE_\QQ[\loss(\Response, f(\Covariate))] - \EE_{\tilde{\QQ}}[\loss(\Response, f(\Covariate))]\right| &\leq 2\bound \cdot \left| \frac{K}{K+ \numobs_1} - \frac{1}{2} \right|.
\end{align*}
Similar to above, we require the following result relating $\RadComp_{K + \numobs_1}(\tilde{\QQ})$ and $\RadComp_{K + \numobs_1}(\QQ)$:
\begin{align*}
\RadComp_{K + \numobs_1}(\tilde{\QQ}) \leq  \RadComp_{K + \numobs_1}(\QQ) + \frac{2\bound}{\sqrt{K + \numobs_1}} + 2\bound \cdot \left| \frac{K}{K + \numobs_1} - \frac{1}{2} \right|.
\end{align*}
It follows by the same argument as~\Cref{lem:annoying-stuff}. We also have the intermediate result
\begin{align*}
&\EE\left[ \sup_{f \in \fclass}\left| \frac{1}{K + \numobs_1} \left( \sum_{i \in H_0} \rad_i \cdot \loss(0, f(\Covariate_i)) + \sum_{j \in I_1} \rad_j \cdot \loss(1, f(\Covariate_j))  \right) \right| \right] \\
&\qquad \qquad \qquad \leq \RadComp_{K + \numobs_1}(\tilde{\QQ}) + \frac{\bound}{\sqrt{K + \numobs_1}}.
\end{align*}

By~\Cref{lem:bounded-diff} we have
\begin{align*}
&\sup_{f \in \fclass} \left| \frac{1}{K + \numobs_1} \left( \sum_{i \in H_0} \loss(0, f(\Covariate_i)) + \sum_{j \in I_1} \loss(1, f(\Covariate_j)) - \EE_\QQ[\loss(\Response, f(\Covariate))]\right) \right| \\
& \qquad \leq \EE\left[\sup_{f \in \fclass} \left| \frac{1}{K + \numobs_1} \left( \sum_{i \in H_0} \loss(0, f(\Covariate_i)) + \sum_{j \in I_1} \loss(1, f(\Covariate_j)) - \EE_\QQ[\loss(\Response, f(\Covariate))]\right) \right|\right] + \sqrt{\frac{8\bound^2\log(1/\parprob)}{K + \numobs_1}} \\
& \qquad \leq 2 \RadComp_{K + \numobs_1}(\tilde{\QQ}) + \left(\sqrt{8\log(1/\parprob)} + 2 \right) \cdot \frac{\bound}{\sqrt{K+ \numobs_1}} + 2\bound \cdot \left| \frac{K}{K + \numobs_1} - \frac{1}{2} \right| \\
& \qquad \leq 2\RadComp_{K + \numobs_1}(\QQ) + \left( \sqrt{8\log(1/\parprob)} + 6 \right) \cdot \frac{\bound}{\sqrt{K + \numobs_1}} + 6\bound \cdot \left| \frac{K}{K+\numobs_1} - \frac{1}{2} \right|.
\end{align*}

\subsection{Proof of~\Cref{lem:generic-slow}}
\label{sec:proof-generic-slow}

 We follow the same approach as the proof of~\Cref{thm:slow-rates}, although it is not necessary to use the coupling approach to construct $\BCovariate$. Let $\{\Covariate_j'\}_{j=1}^J$ be drawn i.i.d.~from $\CondDist{1}$. Then we have that
\begin{align*}
\EE_\QQ\left[ \loss(\Response, \fhat(\Covariate)) - \loss(\Response, \fstar(\Covariate)) \right] \leq 2\Term_1 + 2 \Term_2'
\end{align*}
where we define
\begin{align*}
\Term_2' &\coloneqq \sup_{f \in \fclass} \left| \frac{1}{\numobs + J} \left( \sum_{j=1}^J \big( \loss(1, f(\SCovariate_j)) - \loss(1, f(\Covariate_j')) \big) \right) \right|, \qquad \text{and} \\
\Term_1 &\coloneqq \sup_{f \in \fclass} \left| \frac{1}{\numobs + J} \left( \sum_{i=1}^\numobs \loss(\Response_i, f(\Covariate_i)) + \sum_{j=1}^J \loss(1, f(\Covariate_j')) \right) - \EE_\QQ\left[ \loss(\Response, f(\Covariate)) \right] \right|. 
\end{align*}
Observe that $\Term_1$ is the same as in the proof of~\Cref{thm:slow-rates} and we can use~\Cref{lem:app-bounded-diff} to control it. Therefore we focus on bounding $\Term_2'$. 

Note that, by the triangle inequality, we have
\begin{align*}
\Term_2' \leq G_1 + G_2 + G_3
\end{align*}
where
\begin{align*}
G_1 &\coloneqq \sup_{f \in \fclass} \left| \frac{1}{\numobs+J} \sum_{j=1}^J \big(\loss(1, f(\SCovariate_j)) - \EE_{\SynDist}\left[ \loss(1, f(\Covariate)) \right] \big) \right| \\
G_2 &\coloneqq \sup_{f \in \fclass} \left| \frac{1}{\numobs + J} \sum_{j=1}^J \big(\loss(1, f(\Covariate_j')) - \EE_{\CondDist{1}}\left[\loss(1, f(\Covariate)) \right] \big) \right|, \qquad \text{and} \\
G_3 &\coloneqq \frac{J}{\numobs + J}  \cdot \sup_{f \in \fclass} \left| \EE_{\SynDist}\left[ \loss(1, f(\Covariate)) \right] - \EE_{\CondDist{1}} \left[ \loss(1, f(\Covariate)) \right]  \right|.
\end{align*}
We now bound these terms individually.

\noindent\textbf{Controlling $G_2$:} By applying~\Cref{lem:bounded-diff} and symmetrization, we have
\begin{align*}
G_2 &\leq \frac{J}{\numobs + J} \cdot \left( 2\RadComp_J(\CondDist{1}) + \sqrt{8\log(1/\parprob)} \cdot \frac{\bound}{\sqrt{J}} \right),
\end{align*}
with probability exceeding $1 - \parprob$. Applying~\Cref{eqn:awesome-jason-lemma}, we have
\begin{align*}
G_2 \leq 2\RadComp_{\numobs + J}(\QQ) + \big( 2\sqrt{2\log(1/\parprob)} + 6 \big) \cdot \frac{\bound}{\sqrt{\numobs + J}} + 4\bound \cdot \left| \frac{\numobs \cprob_0}{\numobs + J} - \frac{1}{2} \right|.
\end{align*}

\noindent\textbf{Controlling $G_1$:} Similarly as above we have, with probability exceeding $1 - \parprob$,
\begin{align*}
G_1 &\leq \frac{J}{\numobs + J} \cdot \left( 2\RadComp_J(\SynDist) + \sqrt{8\log(1/\parprob)} \cdot \frac{\bound}{\sqrt{J}} \right),
\end{align*}
although this result is conditional on $\dataset_1$.

\subsection{Proof of~\Cref{lem:bootstrapping}}
\label{sec:proof-bootstrapping}

For this lemma, we work conditioning on the outcomes $\{\Response_i\}_{i=1}^\numobs$; recalling the definition $I_1 = \{i \in [\numobs]: \Response_i = 1\}$, we have that this implies $\numobs_1 = |I_1| = |\dataset_1|$ is fixed and $\dataset_1 = \{\Covariate_i: i \in I_1\} \stackrel{i.i.d.}{\sim} \CondDist{1}$.

\noindent\textbf{Proof of~\Cref{eqn:bootstrapping-a}:} We seek to control
\begin{align*}
\sup_{f \in \fclass} \left| \EE_{\SynDist}\left[\loss(1, f(\Covariate)) \right] - \EE_{\CondDist{1}}\left[ \loss(1, f(\Covariate)) \right] \right|.
\end{align*}
Note that since $\SynDist = \widehat{\PP}_{\dataset_1}$, we can write
\begin{align*}
\EE_{\SynDist}\left[ \loss(1, f(\Covariate)) \right] = \frac{1}{\numobs_1} \sum_{j \in I_1} \loss(1, f(\Covariate_j))
\end{align*}
where $I_1$ are the indices $j$ of the samples in $\dataset$ where $\Response_j = 1$. Thus we seek to control
\begin{align*}
\sup_{f \in \fclass}\left| \frac{1}{\numobs_1} \sum_{j \in I_1} \loss(1, f(\Covariate_j)) - \EE_{\CondDist{1}}\left[ \loss(1, f(\Covariate)) \right]\right|.
\end{align*}
By applying~\Cref{lem:bounded-diff} and symmetrization, we obtain
\begin{align*}
&\sup_{f \in \fclass}\left| \frac{1}{\numobs_1} \sum_{j \in I_1} \loss(1, f(\Covariate_j)) - \EE_{\CondDist{1}}\left[ \loss(1, f(\Covariate)) \right]\right| \\
&\qquad \qquad \leq 2\RadComp_{\numobs_1}(\CondDist{1}) + 2\sqrt{2\log(1/\parprob)} \cdot \frac{\bound}{\sqrt{\numobs_1}}
\end{align*} 
with probability exceeding $1 - \parprob$, as desired.

\noindent\textbf{Proof of~\Cref{eqn:bootstrapping-b}:} We write $\EE_{\dataset_1}$ to indicate that we take the expectation over a set of samples $\{\Covariate_j\}_{j=1}^{\numobs_1} \stackrel{i.i.d.}{\sim} \CondDist{1}$. We also write $\ind \in [\numobs_1]^{J}$ to indicate that it is a function mapping $[J] \to [\numobs_1]$ where we use $\ind_j \coloneqq \ind(j)$ for convenience. Observe that we can write 
\begin{align*}
h(\Covariate_1, \ldots, \Covariate_{\numobs_1}) &= \RadComp_{J}(\SynDist) 
= \frac{1}{\numobs_1^{J}} \sum_{\ind \in [\numobs_1]^J} \EE_{\rad}\left[ \sup_{f \in \fclass}\left| \frac{1}{J} \sum_{j=1}^J \rad_j \cdot \loss(1, f(\Covariate_{\ind_j})) \right| \right].
\end{align*}
Therefore we can bound
\begin{align*}
\left| h(\Covariate_1, \Covariate_2, \ldots, \Covariate_{\numobs_1}) - h(\Covariate_1', \Covariate_2, \ldots, \Covariate_{\numobs_1}) \right| 
&\leq \frac{1}{\numobs_1^{J}} \sum_{\ind \in [\numobs_1]^{J}} \big|\{j: \ind_j = 1\} \big| \cdot \frac{2\bound}{J}\\
&= \frac{2\bound}{J} \cdot \EE_{\ind' \sim \text{Unif}([\numobs_1]^{J})}\left[\big| \{j : \ind'_j = 1\} \big|\right] = \frac{2\bound}{\numobs_1}.
\end{align*}
The second step follows from the fact that the term  $\frac{1}{\numobs_1^J}\sum_{\ind \in [J]^{\numobs_1}} \big|\{j: \ind_j = 1\} \big|$ is the full expression for the expected number of times the first index will be drawn. Applying~\Cref{lem:bounded-diff} then yields, with probability exceeding $1 - \parprob$,
\begin{align*}
\RadComp_{J}(\SynDist) \leq \EE_{\dataset_1} \left[ \RadComp_{J}(\SynDist) \right] + 2\sqrt{2\log(1/\parprob)} \cdot \frac{\bound}{\sqrt{\numobs_1}}.
\end{align*}

Now it remains to control $\EE_{\dataset_1} \left[ \RadComp_{J}(\SynDist) \right]$. To do so, note that if we define $S_i \coloneqq \sum_{j=1}^J \rad_j \mathbf{1}\{\ind_j = i\}$, we can then write
\begin{align*}
\RadComp_J(\SynDist) = \EE_{S} \left[ \sup_{f \in \fclass} \left| \frac{1}{J} \sum_{i=1}^{\numobs_1} S_i \cdot \loss(1, f(\Covariate_i)) \right| \mid \dataset_1\right]
\end{align*}
where $\ind \sim \text{Unif}([\numobs_1]^{J})$. Note that since $\{\rad_j\}_{j=1}^J \indep \ind$, we have that $S_i$ is symmetric around $0$ and can correspondingly write
\begin{align*}
\EE_{S} \left[ \sup_{f \in \fclass} \left| \frac{1}{J} \sum_{i=1}^{\numobs_1} S_i \cdot \loss(1, f(\Covariate_i)) \right| \mid \dataset_1 \right] = \EE_{\xi, S}\left[ \sup_{f \in \fclass} \left| \frac{1}{J} \sum_{i=1}^{\numobs_1} \xi_i|S_i| \cdot \loss(1, f(\Covariate_i)) \right| \mid \dataset_1 \right] 
\end{align*}
where $\xi_i$ are i.i.d. Rademacher variables independent of $S$. Thus
\begin{align*}
\EE_{\xi, S}\left[ \sup_{f \in \fclass} \left| \frac{1}{J} \sum_{i=1}^{\numobs_1} \xi_i|S_i| \cdot \loss(1, f(\Covariate_i)) \right| \mid \dataset_1 \right]  &= \EE_{S}\left[ \EE_{\xi}\left[\sup_{f \in \fclass} \left| \frac{1}{J} \sum_{i=1}^{\numobs_1} \xi_i|S_i| \cdot \loss(1, f(\Covariate_i)) \right| \mid \dataset_1, S \right] \mid \dataset_1 \right] \\
&\leq 2\EE_{S}\left[\max_{i\in[\numobs_1]} |S_i| \right] \cdot \EE_{\xi}\left[\sup_{f \in \fclass} \left| \frac{1}{J} \sum_{i=1}^{\numobs_1} \xi_i \cdot \loss(1, f(\Covariate_i)) \right| \mid \dataset_1 \right]
\end{align*}
where the inequality follows from the Ledoux-Talagrand contraction inequality and independence. Therefore we conclude that
\begin{align*}
\EE_{\dataset_1}\left[ \RadComp_{J}(\SynDist) \right] \leq \frac{2\numobs_1}{J} \cdot  \EE\left[\max_{i \in [\numobs_1]} |S_i| \right] \cdot \RadComp_{\numobs_1}(\CondDist{1}).
\end{align*}
With the following lemma, we complete the proof.
\begin{lemma}
\label{lem:maximal-control}
We have
\begin{align*}
\EE\left[ \max_{i\in[\numobs_1]} |S_i| \right] \leq 2\left(\sqrt{\frac{J}{\numobs_1} \log(2\numobs_1)} + \log(2\numobs_1)  \right).
\end{align*}
\end{lemma}

\subsection{Proof of~\Cref{lem:maximal-control}}
\label{sec:proof-maximal-control}

By standard arguments we have
\begin{align*}
\EE\left[ \max_{i\in[\numobs_1]} |S_i| \right]  \leq \frac{1}{\lambda} \log\left(\EE\left[\max_{i\in[\numobs_1]} e^{\lambda|S_i|}\right]\right) \leq \frac{1}{\lambda} \log\left(2\numobs_1 \cdot \EE\left[ e^{\lambda S_i} \right] \right).
\end{align*}
We can bound (using the shorthand $p \coloneqq \frac{1}{\numobs_1}$)
\begin{align*}
\log\left( \EE\left[e^{\lambda S_i} \right] \right) &= J\log \left(1 - p + \frac{p}{2} \left(e^\lambda + e^{-\lambda} \right) \right) \\
&\leq J \log\left(1 - p + pe^{\lambda^2/2}  \right) \leq Jp\left( e^{\lambda^2/2} - 1 \right),
\end{align*}
where we used the inequalities $e^{z}+e^{-z} \leq 2e^{z^2/2}$ and $\log(1+z) \leq z$ for $z \geq 0$.
Therefore we have
\begin{align*}
\EE\left[ \max_{i\in[\numobs_1]} |S_i| \right] \leq \frac{\log(2\numobs_1)}{\lambda} + \frac{Jp}{\lambda} \left( e^{\lambda^2/2} - 1 \right).
\end{align*}
We now proceed by casework on the value of $Jp$.

\noindent\textbf{Case $Jp \leq \log(2\numobs_1)$:} In this case, we choose $\lambda = 1$ and we get
\begin{align*}
\EE\left[\max_{i\in[\numobs_1]} |S_i| \right] \leq e^{1/2} \cdot \log(2\numobs_1) \leq 2\log(2\numobs_1).
\end{align*}

\noindent\textbf{Case $Jp \geq \log(2\numobs_1)$:} Here we use the fact that $e^{x^2/2} -1 \leq x^2$ for all $x \in [0, 1]$ to get that choosing $\lambda = \sqrt{\frac{\log(2\numobs_1)}{Jp}}$ yields
\begin{align*}
\EE\left[\max_{i\in[\numobs_1]} |S_i| \right] \leq \frac{\log(2\numobs_1)}{\lambda} + Jp \cdot \lambda \leq 2\sqrt{\frac{J}{\numobs_1} \log(2\numobs_1)}.
\end{align*}

\subsection{Proof of~\Cref{eqn:rad-relations}}
\label{sec:proof-rad-relations}

By the same approach as the proof of~\Cref{eqn:awesome-jason-lemma}
\begin{align*}
\RadComp_{\numobs_1}(\CondDist{1}) &= \EE_{\substack{\Covariate'_j \stackrel{i.i.d.}{\sim} \CondDist{1} \\ \rad_j \stackrel{i.i.d.}{\sim} \text{Rad}}} \left[ \sup_{f \in \fclass} \left| \frac{1}{\numobs_1} \sum_{j=1}^{\numobs_1} \rad_j \loss(1, f(\Covariate'_j)) \right| \right] \\
&= \EE_{\substack{\Covariate'_j \stackrel{i.i.d.}{\sim} \CondDist{1} \\ \rad_j \stackrel{i.i.d.}{\sim} \text{Rad}}}\left[\sup_{f \in \fclass} \left|\EE_{\substack{\Covariate_j \stackrel{i.i.d.}{\sim} \CondDist{0} \\ \rad_{\numobs_1 + j} \stackrel{i.i.d.}{\sim} \text{Rad} }}\left[\frac{1}{\numobs_1} \sum_{j=1}^{\numobs_1} \left(\rad_j \loss(1, f(\Covariate'_j)) + \rad_{\numobs_1 + j} \loss(0, f(\Covariate_j))  \right)  \mid \{\rad_j, \Covariate'_j\}_{j=1}^{\numobs_1} \right] \right| \right] \\
&\leq \EE_{\substack{\Covariate_j \stackrel{i.i.d.}{\sim} \CondDist{0} \\ \Covariate'_j \stackrel{i.i.d.}{\sim} \CondDist{1} \\ \rad_j \stackrel{i.i.d.}{\sim} \text{Rad}}}\left[ \sup_{f \in \fclass} \left| \frac{1}{\numobs_1} \sum_{j=1}^{\numobs_1} \left( \rad_j \loss(1, f(\Covariate'_j)) + \rad_{\numobs_1 + j} \loss(0, f(\Covariate_j))  \right)  \right| \right] \\
&\leq 2\RadComp_{2\numobs_1}(\QQ) + \frac{\bound\sqrt{2}}{\sqrt{\numobs_1}},
\end{align*}
where the final step follows from applying~\Cref{eqn:mixture-comp}. The other statement follows from subadditivity of the Rademacher complexity.

\subsection{Proof of~\Cref{lem:SMOTE}}
\label{sec:proof-SMOTE}

We write $\SCovariate(\Covariate_j)$ to indicate the output of using SMOTE to generate a synthetic sample conditional on $\Covariate_j$ being chosen as the initial point, and $\SCovariate_{(k)}(\covariate)$ to denote the $k^{th}$ nearest neighbor in $\dataset_1$ of $x \in \text{supp}(\Covariate)$.

\noindent\textbf{Proof of~\Cref{eqn:SMOTE-a}:} We have
\begin{align*}
&\left| \left(\EE_{\SynDist} - \EE_{\CondDist{1}}\right) \left[ \loss(1, f(\Covariate)) \right] \right| \\
&\qquad \qquad = \left|\frac{1}{\numobs_1} \sum_{j\in I_1} \EE_{\SCovariate}\left[ \loss(1, f(\SCovariate(\Covariate_j))) \mid \dataset_1 \right] - \EE_{\CondDist{1}}\left[ \loss(1, f(\Covariate)) \right] \right| \\
&\qquad \qquad \leq \left|\frac{1}{\numobs_1} \sum_{j \in I_1} \left(\EE_{\SCovariate}\left[\loss(1, f(\SCovariate(\Covariate_j))) \mid \dataset_1 \right]- \loss(1, f(\Covariate_j)) \right) \right| \\
&\qquad \qquad \qquad \qquad  + \left| \frac{1}{\numobs_1} \sum_{j \in I_1} \loss(1, f(\Covariate_j)) - \EE_{\CondDist{1}} \left[ \loss(1, f(\Covariate)) \right]  \right|.
\end{align*}
By previous results we have
\begin{align*}
&\sup_{f \in \fclass} \left| \frac{1}{\numobs_1} \sum_{j \in I_1} \loss(1, f(\Covariate_j)) - \EE_{\CondDist{1}} \left[ \loss(1, f(\Covariate)) \right]  \right| \\
&\qquad \qquad \leq 2\RadComp_{\numobs_1}(\CondDist{1}) + 2\sqrt{2\log(1/\parprob)} \cdot \frac{\bound}{\sqrt{\numobs_1}}
\end{align*}
with probability exceeding $1 - \parprob$. 

To control the other term, we invoke a result on nearest neighbors estimation. To see why, we have by the Lipschitz assumption~\eqref{eqn:SMOTE-lipschitz}
\begin{align*}
&\left|\frac{1}{\numobs_1} \sum_{j \in I_1} \left(\EE_{\SCovariate}\left[\loss(1, f(\SCovariate(\Covariate_j))) \mid \dataset_1 \right]- \loss(1, f(\Covariate_j)) \right) \right| \\
&\qquad \qquad \leq  L \cdot \left|\frac{1}{\numobs_1} \sum_{j \in I_1} \EE_{\SCovariate}\left[ \norm{\SCovariate(\Covariate_j) - \Covariate_j}{2} \mid \dataset_1 \right] \right|   \leq \frac{L}{\numobs_1} \sum_{j\in I_1} \| \SCovariate_{(k)}(\Covariate_j) - \Covariate_j\|_2.
\end{align*}
We now proceed with two separate proofs for controlling the above quantity, leading to each result.

\noindent\textbf{Controlling the maximal distance:} We have
\begin{align*}
\frac{L}{\numobs_1} \sum_{j\in I_1} \| \SCovariate_{(k)}(\Covariate_j) - \Covariate_j\|_2 \leq \frac{L}{\numobs_1} \sum_{j \in I_1} r_k(\Covariate_j) \leq L \cdot \sup_{x \in \text{supp}(\Covariate)} r_k(\covariate)
\end{align*}
and applying~\Cref{lem:nearest-neighbors} yields
\begin{align*}
\frac{L}{\numobs_1} \sum_{j\in I_1} \| \SCovariate_{(k)}(\Covariate_j) - \Covariate_j\|_2 \leq L \left( \frac{3k + 4d \cdot \log(16\numobs_1/\parprob)}{c_d\numobs_1}\right)^{1/d}
\end{align*}
with probability exceeding $1 - \parprob$.

\noindent\textbf{Bounded differences:} In this approach we invoke bounded differences. Defining 
\begin{align*}
h(\{\Covariate_j: j \in I_1\}) \coloneqq \frac{L}{\numobs_1} \sum_{j \in I_1} \|\SCovariate_{(k)}(\Covariate_j) - \Covariate_j \|_2,
\end{align*}
we have by the results in~\Cref{sec:maximum-degree} the maximum number of points any given $\Covariate_i$ can be one of the $k$-nearest neighbor for is $k5^d$. Thus we have by applying~\Cref{lem:bounded-diff} that 
\begin{align*}
\frac{L}{\numobs_1} \sum_{j \in I_1} \|\SCovariate_{(k)}(\Covariate_j) - \Covariate_j \|_2 \leq \EE\left[\frac{L}{\numobs_1} \sum_{j \in I_1} \|\SCovariate_{(k)}(\Covariate_j) - \Covariate_j \|_2\right] + 2LD \sqrt{2\log(1/\parprob)} \cdot \frac{k5^d + 1}{\sqrt{\numobs_1}}
\end{align*}
with probability exceeding $1 - \parprob$. Then by~\Cref{lem:nearest-distance} we have
\begin{align*}
\EE\left[\frac{L}{\numobs_1} \sum_{j \in I_1} \|\SCovariate_{(k)}(\Covariate_j) - \Covariate_j \|_2\right] \leq 6LD \cdot \left( \frac{k}{\numobs_1} \right)^{1/d}.
\end{align*}

\noindent\textbf{Proof of~\Cref{eqn:SMOTE-b}:} Throughout this proof we condition on $\dataset_1$ where we reindex it to be $\dataset_1 = \{\Covariate_j\}_{j=1}^{\numobs_1}$. By definition we have
\begin{align*}
\RadComp_J(\SynDist) = \EE_{\SCovariate, \rad}\left[ \sup_{f \in \fclass} \left| \frac{1}{J} \sum_{j =1}^J \rad_j \cdot \loss(1, f(\SCovariate(\Covariate_{B_j})))  \right| \mid \dataset_1 \right],
\end{align*}
where, like previously, we let $B_j$ denote a random draw over $\text{Unif}[\numobs_1]$. By the triangle inequality we have 
\begin{align*}
&\EE_{\SCovariate, \rad}\left[ \sup_{f \in \fclass} \left| \frac{1}{J} \sum_{j =1}^J \rad_j \cdot \loss(1, f(\SCovariate(\Covariate_{B_j})))  \right| \mid \dataset_1 \right] \\
&\qquad \qquad \leq \EE_{\SCovariate, \rad}\left[ \sup_{f \in \fclass} \left| \frac{1}{J} \sum_{j = 1}^{J} \rad_j \cdot \left(\loss(1, f(\SCovariate(\Covariate_{B_j}))) - \loss(1, f(\Covariate_{B_j})) \right)  \right| \mid \dataset_1 \right] \\
&\qquad \qquad \qquad \qquad + \EE_{\rad}\left[\sup_{f \in \fclass} \left| \frac{1}{J} \sum_{j =1}^{J} \rad_j \cdot \loss(1, f(\Covariate_{B_j})) \right|\right] \\
&\qquad \qquad \leq L \cdot \EE\left[ \frac{1}{J} \sum_{j=1}^J \| \SCovariate(\Covariate_{B_j}) - \Covariate_{B_j} \|_2 \mid \dataset_1 \right] + \RadComp_{J}(\widehat{\PP}_{\dataset_1}).
\end{align*}
Here, $\widehat{\PP}_{\dataset_1}$ denotes drawing randomly with replacement from $\dataset_1$.

Note then we have
\begin{align*}
 \EE\left[ \frac{1}{J} \sum_{j=1}^J \| \SCovariate(\Covariate_{B_j}) - \Covariate_{B_j} \|_2 \mid \dataset_1 \right] &=  \EE\left[ \| \SCovariate(\Covariate_{B_1}) - \Covariate_{B_1} \|_2 \mid \dataset_1 \right] \\
 &\leq \frac{1}{\numobs_1} \sum_{j \in I_1} \norm{\SCovariate_{(k)}(\Covariate_j) - \Covariate_j}{2} \leq \min\{H_1, H_2\}
\end{align*}
with probability exceeding $1 - \parprob$, by the previous argument. By~\Cref{lem:bootstrapping} we have
\begin{align*}
\RadComp_{J}(\widehat{\PP}_{\dataset_1}) &\leq 4\left( \sqrt{\tfrac{\numobs_1}{J} \log(2\numobs_1)} + \tfrac{\numobs_1}{J} \log(2\numobs_1) \right) \cdot \RadComp_{\numobs_1}(\CondDist{1}) \\
&\qquad \qquad  + 2\sqrt{2\log(1/\parprob)} \cdot \frac{\bound}{\sqrt{\numobs_1}}.
\end{align*}
Putting together the pieces, we conclude
\begin{align*}
\RadComp_J(\SynDist) &\leq 4\left( \sqrt{\tfrac{\numobs_1}{J} \log(2\numobs_1)} + \tfrac{\numobs_1}{J} \log(2\numobs_1) \right) \cdot \RadComp_{\numobs_1}(\CondDist{1})  \\
&\qquad \qquad + L \min\{H_1, H_2\} + 2\sqrt{2\log(1/\parprob)} \cdot \frac{\bound}{\sqrt{\numobs_1}},
\end{align*}
with probability exceeding $1 - \parprob$.

\subsection{Proof of results in~\Cref{sec:localization-boot}}
\label{sec:proof-localization-boot}

Upon inspection of the proof of~\Cref{thm:fast-rates}, we have that
\begin{align*}
\frac{\gamma}{2} \norm{\fhat - \fstar}{\QQ_\Covariate}^2 \leq \STerm + \STerm_\EE + H_1 + H_2 + H_3,
\end{align*}
where the quantities are defined throughout its proof. The only difference is in how we control $H_2$ and $H_3$, which we recall
\begin{align*}
H_2 &\coloneqq - \frac{1}{\totobs} \sum_{j=1}^J \left( \ExcessLoss{\fhat}(1, \SCovariate_j) - \EE_{\SynDist}\left[ \ExcessLoss{\fhat}(1, \Covariate) \right] \right), \qquad \text{and} \\
H_3 &\coloneqq \frac{J}{\totobs} \cdot \left( \EE_{\CondDist{1}}\left[ \ExcessLoss{\fhat}(1, \Covariate) \right] - \EE_{\SynDist}\left[\ExcessLoss{\fhat}(1, \Covariate) \right] \right).
\end{align*}
The following lemmas control these terms.
\begin{lemma}
\label{lem:n1-localization}
On the event that $\norm{\fhat - \fstar}{\QQ_\Covariate} \geq \ocrit_{\numobs_1}(\CondDist{1})$, we have
\begin{align*}
H_3 \leq 4L \norm{\fhat - \fstar}{\QQ_\Covariate} \left( 6\ocrit_{\numobs_1}(\CondDist{1}) + \sqrt{\frac{\log(\logfun(\mcrit_\totobs)/\parprob)}{\numobs_1}}\right) + \frac{32L \log(\logfun(\mcrit_\totobs)/\parprob)}{\numobs_1}
\end{align*}
with probability exceeding $1 - \parprob$.
\end{lemma}
\noindent See~\Cref{sec:proof-n1-localization} for a proof of this result. The next lemma controls $H_2$; for convenience we use $\ocrit_{\numobs_1}(\dataset_1)$ to denote the empirical Rademacher complexity with respect to a fixed design given by $\dataset_1$ and the shorthand $\xi_{\numobs_1, J} \coloneqq  \sqrt{\tfrac{\numobs_1}{J} \log(2\numobs_1)} + \tfrac{\numobs_1}{J} \log(2\numobs_1)$.
\begin{lemma}
\label{lem:bootstrap-localization}
We have
\begin{align*}
H_2 &\leq 4L \max\left\{ \ocrit_{\numobs_1}(\dataset_1), \norm{\fhat - \fstar}{\SynDist} \right\} \left(16\xi_{\numobs_1, J}\cdot  \ocrit_{\numobs_1}(\dataset_1) + \sqrt{\frac{\log(\logfun(\ocrit_{\numobs_1}(\dataset_1))/\parprob)}{\totobs}} \right)  \\
&\qquad \qquad \qquad \qquad + \frac{32L\log(\logfun(\ocrit_{\numobs_1}(\dataset_1))/\parprob)}{\totobs}
\end{align*}
with probability exceeding $1 - \parprob$. 
\end{lemma}
\noindent See~\Cref{sec:proof-bootstrap-localization} for a proof of this result. The statement then follows from standard results on localized empirical processes. For example, Theorem 14.1 and Proposition 14.25 in~\citet{dinosaur2019} state that
\begin{align*}
\left|\norm{\fhat - \fstar}{\SynDist} - \norm{\fhat - \fstar}{\CondDist{1}} \right| \lesssim \ocrit_{\numobs_1}(\CondDist{1}) 
\end{align*}
and
\begin{align*}
\ocrit_{\numobs_1}(\dataset_1) \asymp \ocrit_{\numobs_1}(\CondDist{1}),
\end{align*}
respectively, each with high probability. Thus on the event $\norm{\fhat - \fstar}{\QQ_\Covariate} \geq \ocrit_{\numobs_1}(\CondDist{1})$ we have
\begin{align*}
H_2 \lesssim \norm{\fhat - \fstar}{\QQ_\Covariate} \cdot \xi_{\numobs_1, J} \, \ocrit_{\numobs_1}(\CondDist{1})
\end{align*}
which yields the claim.

\subsection{Proof of~\Cref{lem:n1-localization}}
\label{sec:proof-n1-localization}

Note that, by definition, we have
\begin{align*}
H_3 \coloneqq -\frac{J}{\totobs} \cdot \left( \frac{1}{\numobs_1} \sum_{j\in I_1} \ExcessLoss{\fhat}(1, \Covariate_j) - \EE_{\CondDist{1}}\left[ \ExcessLoss{\fhat}(1, \Covariate) \right] \right).
\end{align*}
Defining
\begin{align*}
Z_{\numobs_1}(t) \coloneqq \sup_{\norm{f - \fstar}{\QQ_\Covariate} \leq t} \left| \frac{1}{\numobs_1} \left( \sum_{j\in I_1} \ExcessLoss{f}(1, \Covariate_j) - \EE_{\CondDist{1}}\left[ \ExcessLoss{f}(1, \Covariate) \right] \right) \right|
\end{align*}
we have that applying~\Cref{lem:emp-conc} gives
\begin{align*}
Z_{\numobs_1}(t) \geq 2\EE[Z_{\numobs_1}(t)] + \sqrt{\sigma^2(\gclass)} \cdot \sqrt{\frac{2s}{\numobs_1}} + \frac{16Ls}{\numobs_1}
\end{align*}
with probability at most $e^{-s}$. We claim that, for $t \geq \ocrit_{\numobs_1}(\CondDist{1})$ we have
\begin{align}
\label{eqn:n1-emp-calcs}
\EE[Z_{\numobs_1}(t)] \leq 6Lt\,\ocrit_{\numobs_1}(\CondDist{1}), \qquad \text{and} \qquad \sigma^2(\gclass) \leq 2L^2t^2.
\end{align}
Therefore we have 
\begin{align*}
Z_{\numobs_1}(t) \geq Q(t, s) \coloneqq 12Lt \, \ocrit_{\numobs_1}(\CondDist{1}) + 2Lt \sqrt{\frac{s}{\numobs_1}} + \frac{16Ls}{\numobs_1} \quad \text{with probability at most $e^{-s}$.}
\end{align*}
Applying~\Cref{lem:peeling} yields 
\begin{align*}
Z_{\numobs_1}(\norm{\fhat - \fstar}{\QQ_\Covariate}) &\leq 4L \cdot \max\left\{\ocrit_{\numobs_1}(\CondDist{1}), \norm{\fhat - \fstar}{\QQ_\Covariate}\right\} \left(6\ocrit_{\numobs_1}(\CondDist{1}) + \sqrt{\frac{\log(\logfun(\ocrit_{\numobs_1}(\CondDist{1}))/\parprob)}{\numobs_1}} \right) \\
&\qquad \qquad \qquad + \frac{32L \log(\logfun(\ocrit_{\numobs_1}(\CondDist{1}))/\parprob)}{\numobs_1}.
\end{align*}
The claim then follows from the fact that $\logfun$ is a decreasing function and 
\begin{align*}
\ocrit_{\numobs_1}(\CondDist{1}) \geq \ocrit_\totobs(\CondDist{1}) \geq \mcrit_\totobs.
\end{align*}

\noindent \textbf{Proof of~\Cref{eqn:n1-emp-calcs}:} Note the bound on $\sigma^2(\gclass)$ follows from previous calculations. The other bound also follows similarly:
\begin{align*}
\EE[Z_{\numobs_1}(t)] &\leq \EE_{\CondDist{1}}\left[ \sup_{\norm{f - \fstar}{\QQ_\Covariate} \leq t} \left| \frac{1}{\numobs_1} \sum_{j \in I_1} \ExcessLoss{f}(1, \Covariate_j) - \EE_{\CondDist{1}}\left[ \ExcessLoss{f}(1, \Covariate) \right] \right| \right] \\
&\leq 4L \cdot \EE\left[\sup_{\norm{f - \fstar}{\QQ_\Covariate} \leq t} \left| \frac{1}{\numobs_1} \sum_{j \in I_1} \rad_j \cdot \big(f(\Covariate_j) - \fstar(\Covariate_j) \big)  \right|\right] \\
&\leq 4L \cdot \EE\left[\sup_{\norm{f - \fstar}{\CondDist{1}} \leq t\sqrt{2}} \left| \frac{1}{\numobs_1} \sum_{j \in I_1} \rad_j \cdot \big(f(\Covariate_j) - \fstar(\Covariate_j) \big)  \right| \right] \\
&= 4L \cdot \RadComp_{\numobs_1}(t\sqrt{2}; \CondDist{1}) \leq 4\sqrt{2} \cdot Lt\, \ocrit_{\numobs_1}(\CondDist{1}).
\end{align*}

\subsection{Proof of~\Cref{lem:bootstrap-localization}}
\label{sec:proof-bootstrap-localization}

Define (where $\{\SCovariate_j\}_{j=1}^J \stackrel{i.i.d.}{\sim} \SynDist$)
\begin{align*}
Z_J(t) \coloneqq \sup_{\norm{f -\fstar}{\SynDist} \leq t} \left| \frac{1}{J} \sum_{j=1}^J \ExcessLoss{f}(1, \SCovariate_j) - \EE_{\SynDist}\left[ \ExcessLoss{f}(1, \Covariate) \right] \right|.
\end{align*}
Note that, by previous arguments, we have
\begin{align*}
Z_J(t) \geq 2\EE[Z_J(t)] + \sqrt{\sigma^2(\gclass)} \cdot \sqrt{\frac{2s}{J}} + \frac{16Ls}{J}
\end{align*}
with probability at most $e^{-s}$. Furthermore we also have $\sigma^2(\gclass) \leq L^2 t^2$. To control the expectation term, we have
\begin{align*}
\EE\left[Z_J(t)\right] \leq 4L \cdot \EE\left[ \sup_{\norm{f - \fstar}{\SynDist} \leq t} \left| \frac{1}{J} \sum_{j=1}^J \rad_j \cdot \big(f(\SCovariate_j) - \fstar(\SCovariate_j) \big) \right| \right] = 4L \cdot \RadComp_J(t; \SynDist).
\end{align*}
By the arguments in~\Cref{sec:proof-bootstrapping} we have
\begin{align*}
\RadComp_J(t; \SynDist) \leq 4 \left(\sqrt{\tfrac{\numobs_1}{J} \log(2\numobs_1)} + \tfrac{\numobs_1}{J} \log(2\numobs_1) \right) \cdot \RadComp_{\numobs_1}(t; \dataset_1)
\end{align*}
where $\RadComp_{\numobs_1}(t; \dataset_1)$ is the empirical Rademacher complexity based on the fixed design for covariates given by $\dataset_1$. By definition of the critical radius, we have for $t \geq \ocrit_{\numobs_1}(\dataset_1)$ that $\RadComp_{\numobs_1}(t;\dataset_1) \leq t \, \ocrit_{\numobs_1}(\dataset_1)$. Thus putting together the pieces, we have
\begin{align*}
Z_J(t) \geq 32L \left(\sqrt{\tfrac{\numobs_1}{J} \log(2\numobs_1)} + \tfrac{\numobs_1}{J} \log(2\numobs_1) \right) \cdot t\ocrit_{\numobs_1}(\dataset_1) + Lt \cdot \sqrt{\frac{2s}{J}} + \frac{16Ls}{J}
\end{align*} 
with probability at most $e^{-s}$. Thus appealing to~\Cref{lem:peeling}  we have
\begin{align*}
Z_J(\norm{\fhat - \fstar}{\SynDist}) &\leq 4L \max\left\{ \ocrit_{\numobs_1}(\dataset_1), \norm{\fhat - \fstar}{\SynDist} \right\} \left(16\xi_{\numobs_1, J}\cdot  \ocrit_{\numobs_1}(\dataset_1) + \sqrt{\frac{\log(\logfun(\ocrit_{\numobs_1}(\dataset_1))/\parprob)}{J}} \right)  \\
&\qquad \qquad \qquad \qquad + \frac{32L\log(\logfun(\ocrit_{\numobs_1}(\dataset_1))/\parprob)}{J}
\end{align*}
with probability exceeding $1 - \parprob$. Thus we have
\begin{align*}
H_2 &\leq \frac{J}{M} \cdot Z_J(\norm{\fhat - \fstar}{\SynDist}) \\
&\leq 4L \max\left\{ \ocrit_{\numobs_1}(\dataset_1), \norm{\fhat - \fstar}{\SynDist} \right\} \left(16\xi_{\numobs_1, J}\cdot  \ocrit_{\numobs_1}(\dataset_1) + \sqrt{\frac{\log(\logfun(\ocrit_{\numobs_1}(\dataset_1))/\parprob)}{\totobs}} \right)  \\
&\qquad \qquad \qquad \qquad + \frac{32L\log(\logfun(\ocrit_{\numobs_1}(\dataset_1))/\parprob)}{\totobs}
\end{align*}
with probability exceeding $1 - \parprob$.

\end{document}